\documentclass[letterpaper]{article} % DO NOT CHANGE THIS
\usepackage{aaai2027}  % DO NOT CHANGE THIS
\usepackage[hyphens]{url}  % DO NOT CHANGE THIS
\usepackage{graphicx} % DO NOT CHANGE THIS
\usepackage{natbib}  % DO NOT CHANGE THIS AND DO NOT ADD ANY OPTIONS TO IT
\usepackage{caption} % DO NOT CHANGE THIS AND DO NOT ADD ANY OPTIONS TO IT
\usepackage{algorithm}
\usepackage{algorithmic}
\usepackage{enumitem}
\usepackage{amsfonts, amsthm}
\usepackage{newfloat}
\usepackage{listings}
\DeclareCaptionStyle{ruled}{labelfont=normalfont,labelsep=colon,strut=off} % DO NOT CHANGE THIS
\floatstyle{ruled}
\newfloat{listing}{tb}{lst}{}
\floatname{listing}{Listing}
\nocopyright
\usepackage{booktabs}

\newtheorem{assumption}{Assumption}
\newtheorem{definition}{Definition}
\usepackage{thmtools}
\usepackage{thm-restate}

\usepackage{amsmath}
\usepackage{amssymb}
\usepackage{subcaption}
\usepackage{makecell}
\usepackage{multirow}
\title{The Trajectory Alignment Coefficient in Two Acts: \\From Reward Tuning to Reward Learning}

\author{
    Calarina Muslimani\textsuperscript{\rm 1,\rm 2}\thanks{Work done while interning at Sony AI. Corresponding author: musliman@ualberta.ca.},
    Yunshu Du\textsuperscript{\rm 1},
    Kenta Kawamoto\textsuperscript{\rm 1},
    Kaushik Subramanian\textsuperscript{\rm 1},\\
    Peter Stone\textsuperscript{\rm 1,\rm 3},
    Peter Wurman\textsuperscript{\rm 1}
}
\affiliations{
    \textsuperscript{\rm 1}Sony AI\\
    \textsuperscript{\rm 2}University of Alberta\\
    \textsuperscript{\rm 3}The University of Texas at Austin
}

\begin{document}

\maketitle

\begin{abstract}
The success of reinforcement learning (RL) is fundamentally tied to having a reward function that accurately reflects the task objective. Yet, designing reward functions is notoriously time-consuming and prone to misspecification. 
To address this issue, our first goal is to understand how to support RL practitioners in specifying reward functions. 
We leverage the Trajectory Alignment Coefficient (TAC), 
a metric that evaluates how closely a reward function’s induced preferences match those of a domain expert. 
To test whether TAC provides meaningful support in practice, we evaluated its impact in a user study where practitioners designed reward functions for Lunar Lander. We found that receiving iterative TAC feedback allowed participants to design more performant reward functions while reporting lower cognitive workload compared to unguided reward design.
However, the study also underscored that manual reward design, even with TAC, remains labor-intensive. This limitation motivated our second goal: to learn a reward model that maximizes TAC directly. Specifically, we propose Soft-TAC, a differentiable approximation of TAC that can be used as a loss function to train reward models from human preferences. 
Our theoretical analysis guarantees that under Soft-TAC optimization: (1) the optimal reward function belongs exclusively to the set of human-aligned reward functions, and (2) this optimal reward remains invariant whether trained on clean or noisy preference data. 
Empirically, we first evaluate reward models trained with Soft-TAC across three open-source tasks (Lunar Lander, Reacher, and Hopper), demonstrating that they consistently match or outperform those trained with the Cross-Entropy loss. Next, we evaluate Soft-TAC in the high-fidelity simulator Gran Turismo 7, and find that Soft-TAC-trained reward models more successfully capture distinct, preference-specific driving styles.
Ultimately, this work demonstrates that TAC can serve as both a practical tool for guiding reward design and a reward learning objective in complex domains. 
\end{abstract}

\section{Introduction}
The reward function lies at the foundation of reinforcement learning (RL), as agents aim to maximize the expected cumulative reward \citep{sutton2018reinforcement}. In practice, however, solving a task with RL assumes access to a reward function that accurately captures the task objective. Without such a reward function, learning can fail entirely, leading agents to pursue undesired or unsafe behaviors \citep{amodei2016concreteproblemsaisafety,defining_reward_gaming}. 
This issue can be particularly pronounced in real-world domains such as autonomous driving \citep{reward_misdesign_AD}, where reward functions must capture preference-specific objectives (e.g., driving style or adherence to etiquette).
Most practitioners rely on a trial-and-error approach, iteratively adjusting reward functions based on the behavior of RL agents trained with those rewards. This process is costly, time-consuming, and often results in reward misspecification \citep{perils_reward_design}. Despite its importance, best practices for reward design have received relatively little attention, particularly regarding tools that help practitioners iteratively refine reward specifications.

The first goal of this work is to evaluate the utility of the Trajectory Alignment Coefficient (TAC) \citep{TAC}, a reward alignment metric, for supporting RL practitioners during manual reward design.
% While other reward evaluation metrics exist, we focus on TAC because alternative approaches are unsuitable for our goal. They either require access to a ground-truth reward (which we cannot assume) \citep{epic,dard, skalse2024starc}, or only offer a binary indication of alignment \citep{value_alignment,specifying_rl_objectives} Since our aim is to support iterative refinement, we require a continuous signal, which TAC provides.
While other reward evaluation metrics exist, we focus on TAC because alternative approaches either require access to a ground-truth reward \citep{epic,dard,skalse2024starc}, which we cannot assume, or fail to provide a continuous alignment signal \citep{specifying_rl_objectives}, which is needed for iterative reward refinement. 

We investigate whether TAC can guide RL practitioners in specifying reward functions. To do so, we conducted a human-subject study in Lunar Lander \citep{Brockman2016OpenAIGym}, where participants iteratively tuned the domain's eight reward weights either with or without TAC feedback.
In the TAC condition, participants received TAC scores for each proposed reward function, to help them refine their next set of weights.
This guidance led to statistically significant improvements: (1) the final tuned reward functions produced policies with higher task success rates, and (2) participants' perceived cognitive workload decreased by $30\%$ relative to unguided reward design without TAC.

While these results demonstrate that TAC can support effective reward design, the process still required substantial effort; with participants iterating over their designs 40 times.
This observation motivates the second goal of our work: to \emph{automatically learn a reward model that maximizes TAC}. We introduce a differentiable approximation of TAC, \emph{Soft-TAC}, which can be used as a loss function to train reward models directly from human preferences. 

First, we demonstrate theoretically that under Soft-TAC optimization: (1) the optimal reward function belongs exclusively to the set of human-aligned reward functions, and (2) this optimal reward remains invariant whether trained on clean or noisy preference data.  
Second, we empirically evaluate Soft-TAC across four distinct domains: Lunar Lander, Reacher, Hopper, and the high-fidelity driving simulator Gran Turismo 7 (GT7) \citep{wurman2022outracing}. In the first three environments \citep{Brockman2016OpenAIGym}, reward models trained with Soft-TAC consistently match or exceed the performance of the Cross-Entropy baseline \citep{christiano2017deep}.
Meanwhile, in GT7, Soft-TAC more successfully captured preference-specific driving styles, leading to qualitatively more distinct racing behaviors than models trained using Cross-Entropy. Together, our work highlights that TAC can serve as both a tool to support reward design and as an objective for learning human-aligned reward models.
% We evaluate Soft-TAC across two domains: Lunar Lander, to confirm its ability to learn effective reward functions where human practitioners struggled; and the high-fidelity driving simulator Gran Turismo 7 (GT7) \citep{wurman2022outracing}, to assess its effectiveness in producing qualitatively distinct behaviors given diverse human preferences. Our results show that reward models trained with Soft-TAC successfully captured human preferences in both domains. In Lunar Lander, these reward models enabled RL agents to achieve higher task success rates than both the standard Cross-Entropy loss \citep{christiano2017deep} and manually tuned rewards. In GT7, Soft-TAC trained reward models better captured preferences reflecting different driving objectives, leading to qualitatively more distinct racing behaviors than models trained using Cross-Entropy.
% Together, our work highlights that TAC can serve as both a tool to support 
%  reward design and as an objective for learning human-aligned reward models.

% To address this, metrics such as EPIC, DARD, and STARC have been proposed to quantify the divergence between a learned reward and a baseline~\citep{dard, epic, skalse2024starc}. However, these methods typically rely on the baseline being a ground-truth reward function, which is rarely available in real-world applications.

\section{Related Work}\label{sec:related_work}

Preference-based RL (PbRL) is a subfield of RL focused on learning reward functions from human preferences over trajectories. The standard formulation, introduced by \citet{christiano2017deep}, assumes that humans tend to prefer trajectories with higher total reward. This assumption supports a probabilistic preference model (e.g., \citet{bradley_terry_model}), enabling reward learning via the Cross-Entropy loss.
A majority of PbRL methods closely follow this framework, adding only techniques to reduce the number of human preferences or improve policy learning \citep{reward_learning_from_pref_demo,lee2021pebble,rune,surf,hwang2023sequential,shin2023offlinepbrl,hejna2023few,rime,hu2024querypolicy, hindsightpriors,choi2024lire, muslimani2025sdp, kang2025adversarialpbrl,pace2025preferenceelicitation}. Some approaches vary from this setup by modifying the preference model or loss function. However, these methods often come with limitations, such as: (1) requiring online, policy-coupled training \citep{biyik2018batch, liu2022metarewardnet, xie2024hypothesiscutting}; (2) relying on fixed sets of policies for value estimation \citep{knox2024modelspreferences}; (3) removing reward learning altogether \citep{hejna2024contrastive}; or (4) converting preferences into ``win'' and ``loss'' labels, ignoring that a trajectory's label may be dependent on what it is compared to \citep{sun2025rethinking}.
\section{Background}
This section first provides background on RL, then describes the Trajectory Alignment Coefficient and lastly concludes with a discussion of reward learning.
\subsection{Reinforcement Learning}
We consider a Markov decision process defined by the tuple 
$(\mathcal{S}, \mathcal{A}, r, p, \mu, \gamma)$. 
Here, $\mathcal{S}$ and $\mathcal{A}$ denote the state and action spaces, 
$r : \mathcal{S} \times \mathcal{A} \times \mathcal{S} \rightarrow \mathbb{R}$ 
specifies the reward per transition  and 
$p : \mathcal{S} \times \mathcal{A} \times \mathcal{S} \rightarrow [0,1]$ 
defines the transition dynamics. The initial-state distribution is given by $\mu$, 
and the discount factor $\gamma \in [0,1)$ controls the weighting of future rewards.
In this framework, an agent interacts with the environment in discrete time steps, where at time $t$ it observes state $s_t$, selects action $a_t$, transitions to $s_{t+1}$, and receives reward $r_{t+1}$. A trajectory $\tau$ consists of a sequence of $(s_t, a_t, s_{t+1})$ triplets that either ends in a terminal state (episodic tasks) or continues indefinitely (continuing tasks). A trajectory's discounted return with respect to reward function $r$ and discount factor $\gamma$ is:
$G_{r, \gamma}(\tau) = \sum_{t=0}^{T} \gamma^t r_{t+1}$
where $T = \vert{}\tau\vert{} - 1$ for episodic tasks, or $T \to \infty$ for continuing tasks.
The objective in RL is to learn a policy 
$\pi : \mathcal{S} \rightarrow \Delta(\mathcal{A})$ 
that maximizes the expected discounted return.

\subsection{Learning Reward Functions from Preferences}\label{sec:reward_learning_background}
% A common approach to reward learning is preference-based RL, where a reward model $\hat{r}_\theta$ is learned from human preferences \citep{christiano2017deep}.
Following the standard PbRL framework \citep{christiano2017deep}, we consider the problem of learning a parameterized reward model $\hat{r}_\theta$ from human preferences.

% \begin{definition}\label{def:human_pref_dataset}
% A human preference dataset $\mathcal{D}_{h}$ is defined as a finite collection of $N$ samples, $\mathcal{D}_h = \{(\mathbf{x}_k, y_k)\}_{k=1}^N$, where each $\mathbf{x}_k = (\tau_k^{(1)}, \tau_k^{(2)}) \in \mathcal{X}$ represents a trajectory comparison pair and $y_k \in \{-1, 1\}$ represents the corresponding human preference label.
% \end{definition}

\begin{definition}\label{def:human_pref_dataset}
A human preference dataset $\mathcal{D}_{h}$ is defined as $\{ (\tau_k^{(1)}, \tau_k^{(2)}, y_k) \}_{k=1}^N$, where $(\tau_k^{(1)}, \tau_k^{(2)})$ is the $k^{th}$ pair of trajectories and $y_k$ is the preference label.
\end{definition}
Following the Bradley–Terry model \citep{bradley_terry_model}, the probability $P_\theta^{(1 \succ 2)}$ of preferring $\tau^{(1)}$ over $\tau^{(2)}$ is modeled using the exponential of the predicted returns:
\begin{equation}\label{eq:bradley_terry}
P_{\theta}^{(1 \succ 2)} = \frac{\exp\sum_t \hat{r}_\theta(s^{(1)}_t, a^{(1)}_t)}{\sum_{i \in \{1, 2\}} \exp\sum_t \hat{r}_\theta(s^{(i)}_t, a^{(i)}_t)}.
\end{equation}
% Reward functions can be learned from $\mathcal{D}_h$ via supervised learning. For example, $\theta$ can be optimized to maximize the log-likelihood of the observed preferences by minimizing the Cross-Entropy loss:

The parameters $\theta$ are optimized by minimizing the Cross-Entropy loss over the observed preferences in $\mathcal{D}_h$:

\begin{equation}\label{eq:cross_entropy}
\mathcal{L}_{\mathrm{CE}} = -\mathbb{E}_{\mathcal{D}_h} \left[ y \log P_{\theta}^{(1 \succ 2)} + (1-y) \log P_{\theta}^{(2 \succ 1)} \right],
\end{equation}
where $y=1$ if $\tau^{(1)} \succ \tau^{(2)}$ and $y=0$ if $\tau^{(2)} \succ \tau^{(1)}$. 

\subsection{Trajectory Alignment Coefficient (TAC)}
TAC is a reward–alignment metric that assesses how well a reward function and discount factor pair denoted as \((r,\gamma)\) respects a set of human preferences. It is based on Kendall’s Tau--b \citep{kendall_tau_b_variant} which measures agreement between ranked data.
TAC compares finite datasets of binary preferences: human preferences $\mathcal{D}_h$ against $\mathcal{D}_{r, \gamma}$, a dataset constructed by re-evaluating each trajectory pair $(\tau^{(1)}, \tau^{(2)}) \in \mathcal{D}_h$ using expected returns under $(r, \gamma)$.
 For example, $\tau^{(1)}$ is preferred over $\tau^{(2)}$ $\iff G_{r, \gamma}(\tau^{(1)}) > G_{r, \gamma}(\tau^{(2)})$.
% From the human data, we extract the set of unordered trajectory pairs that were compared. For each such pair, we then construct the corresponding preference under the candidate reward,  by comparing their expected returns. Specifically, trajectory $\tau_A$ is preferred over $\tau_B$ if and only if $G_{r, \gamma}(\tau_\text{A}) >  G_{r, \gamma}(\tau_\text{B})$. 
Once we have both \( \mathcal{D}_h \) and \( \mathcal{D}_{r, \gamma} \), TAC measures their agreement using Kendall’s Tau-b:
\begin{equation}
   \sigma_{\textit{TAC}}(\mathcal{D}_h, \mathcal{D}_{r, \gamma}) \doteq \frac{P - Q}{\sqrt{(P + Q + X_0)(P + Q + Y_0)}}
    \label{eq:tac}
\end{equation}
where $P$ and $Q$ denote the number of concordant and discordant pairs between $\mathcal{D}_{r, \gamma}$ and $\mathcal{D}_h$, respectively. While, $X_0$ represents the number of pairs tied only in $\mathcal{D}_{r, \gamma}$ and $Y_0$ represents the number of pairs tied only in $\mathcal{D}_h$.
Equation \ref{eq:tac} outputs a score  $\in [-1,1]$, where $1$ indicates that $(r, \gamma)$ 
induces the same preference relations as the human over all trajectory pairs and $-1$ indicates complete disagreement.
% fully respects the human's preferences (i.e., all trajectory pairs are concordant with the preferences), 
% We selected TAC over existing reward-evaluation metrics such as EPIC, DARD, and STARC because it is the only metric that does not assume access to a ground-truth reward function [cite]. Since our objective is to \emph{design} a new reward function, no such ground truth reward is available.
% Note that given a subset of trajectory distributions, the Trajectory Alignment Coefficient can be applied in cases with either a full or partial ranking. A full ranking establishes a complete order over the elements in a subset, where all necessary pairwise comparisons are available. In contrast, a partial ranking occurs when some pairwise comparisons are missing (e.g., $D_{r, \gamma}, D_\textit{h}$, as the comparison between \( \eta_1 \) and \( \eta_2 \) is missing).
% This flexibility allows the Trajectory Alignment Coefficient to be used in settings where ranking information is limited or incomplete.

\section{Can TAC improve manual reward tuning?}
Our first goal is to evaluate whether TAC can aid RL practitioners in specifying reward weights. Prior work has shown that even in tabular domains, tuning as few as four reward weights is surprisingly difficult \citep{perils_reward_design}. While TAC has shown promise as a reward design tool \citep{TAC}, its practical utility in realistic design settings has not been established: previous studies restricted participants to simply choosing between two fully specified reward functions in a tabular domain. To address these limitations, we evaluate TAC in a high-dimensional environment where practitioners interactively tune weights from scratch.

\subsubsection{Test Domain}
We consider the continuous-action version of Lunar Lander, where the goal is to land a vehicle between two designated flags. This environment was chosen due to the complexity of its reward function. First, it comprises a linear combination of \emph{eight reward features}---significantly more than the MuJoCo OpenAI Gym domains, which typically have $2$ to $4$ features \citep{Brockman2016OpenAIGym}. 
Second, the default Lunar Lander reward function uses vastly different weight scales across features, making uniform weighting ineffective and manual selection more difficult. These properties make Lunar Lander a suitable choice for studying the challenge of tuning reward weights. 
% We describe the environment in more detail in Appendix \ref{sec:env_details_lunar_lander}.

\subsubsection{Study Protocol}\label{sec:study_protocol}
We conducted an ethics-approved user study with 11 RL practitioners. All participants apply RL in academic or professional settings and hold advanced degrees (MSc or PhD). While the sample size is compact, it reflects the specialized expertise of our target population. To prevent carryover effects, we employed a between-subjects design, randomly assigning each participant to either a Control condition or an Alignment condition (i.e., receiving TAC feedback). Moreover, baseline experience in reward design did not differ significantly between groups ($t = 0.88, p = 0.79$), ensuring comparability across conditions.

The study consisted of two primary components. First, participants read a description of Lunar Lander and were informed that they would be collaborating with a domain expert to tune a reward function for RL training. 
% The domain expert (one of the authors) manually reviewed 148 pairs of video clips and selected the
% preferred trajectory for each pair, ensuring the labels reflected
% human judgment rather than using synthetic preferences. 
The domain expert (one of the authors) manually reviewed 148~pairs of video clips and selected the preferred trajectory for each pair, ensuring the labels reflected human judgment instead of generating synthetic preferences.
To ensure behavioral diversity across the video clips, we followed the heuristic proposed by \citet{TAC}, generating rollouts from RL agents trained with different reward functions. We then randomly selected 15 pairs to present to participants, choosing a subset to avoid overwhelming them. Participants were instructed to review the clips to understand the domain expert’s preferences regarding landing behavior.

Second, participants were presented with descriptions and pseudocode for the eight reward features. Their goal was to tune the reward function weights\footnote{We assume the given feature set is sufficient to represent the target objective.} 
to align with domain expert preferences (specifically prioritized touching down on the target pad).
This process was iterative: participants could tune the weights, rewatch the trajectory video clips, and receive condition-specific feedback, adjusting the weights until satisfied. See the appendix for a visualization of the user study.
% This procedure enabled participants to observe how different weight choices affected cumulative trajectory returns and verify whether preferred trajectories yielded higher returns than non-preferred ones (visualized in the appendix).

%The study contained two conditions:
% \begin{itemize}
% \item \textbf{Control:} For each pair of trajectories (from their corresponding video clips), participants were provided with the computed returns under their chosen reward weights..
% \item \textbf{Alignment:} In addition, participants also received respective TAC score, computed from the domain expert's preferences and those induced by the participant’s selected reward function.
% \end{itemize}

\subsubsection{Experimental Conditions and Research Questions}
We evaluated two feedback conditions, where feedback was updated in real time after every weight modification. In the \textbf{Control condition} ($N=5$), participants received the individual returns for both trajectories across all pairs. This allowed them to check, within each pair, whether their selected weights assigned a higher return to the domain expert's preferred trajectory.
In the \textbf{Alignment condition} ($N=6$), participants received the same trajectory returns, augmented with the TAC score measuring alignment between the reward function's implied preferences and those of the domain expert.
By comparing outcomes and experiences between these groups, we investigate whether providing TAC feedback influences how RL practitioners tune reward weights. 
Specifically, we focus on two questions:

\begin{enumerate}[label=\textbf{RQ\arabic*:}, leftmargin=*]
    \item Does TAC help RL practitioners select reward weights that improve downstream RL training?
    \item Does TAC reduce the perceived cognitive workload incurred during reward tuning?
\end{enumerate}

We chose these conditions as most reward design tools including LLM based ones \cite{Ma2024Eureka} require time-prohibitive RL training, ruling out interactive user studies. 
\subsubsection{Evaluation}

To evaluate any reward function, we train an RL agent (e.g., SAC \cite{sac}) across $S=5$ random seeds. For each seed, we pool $K=5$ policy checkpoints near convergence and evaluate each deterministically over $E=5$ independent episodes, totaling $S \times K \times E = 125$ rollouts per reward function (detailed in the appendix). Performance is measured by the \emph{landing-pad success rate}: the fraction of rollouts where the lander touches down with at least one leg on the target pad.

To address RQ1, we evaluated each participant's final reward function using this protocol. This allows us to compare overall success rates between the Alignment and Control conditions.
Next, to evaluate RQ2, participants completed the NASA Task Load Index (NASA-TLX) \citep{hart1988development} immediately after the reward-tuning task to measure their perceived cognitive workload.

Unless otherwise stated, for all statistical comparisons conducted in this paper, we perform one-sided Welch's $t$-tests when normality assumptions hold and Mann–Whitney $U$-tests otherwise. For experiments with multiple comparisons, the $p$-values are adjusted using the Benjamini–Hochberg procedure to control the False Discovery Rate ($p_{\text{adj}}$).
% For all analyses, we used paired \(t\)--tests for continuous data when normality assumptions held or  Wilcoxon Signed-Rank tests otherwise. For categorical voting data, Fisher’s Exact Test was applied.
% The corresponding 
% $p$--values and test statistics are reported: \(t\) for the paired \(t\)--test and \(W\) for the Wilcoxon Signed-Rank test.
% To control for Type~I errors, the Bonferroni correction (\(\alpha=0.05\)) was performed.

\subsection{Results}
Regarding \textbf{RQ1}, we observed that RL agents trained with reward functions tuned in the Alignment condition achieved a substantially higher average landing-pad success rate than those in the Control condition (Alignment: mean $\mu=0.63$, standard error $\mathrm{SE}=0.18$; Control: $\mu=0.20$, $\mathrm{SE}=0.20$). This difference was statistically significant ($U=24.5$, $p=0.042$), suggesting that TAC feedback enabled users to design more effective reward functions.

\begin{figure}[htbp]
    \centering
    \includegraphics[width=\columnwidth]{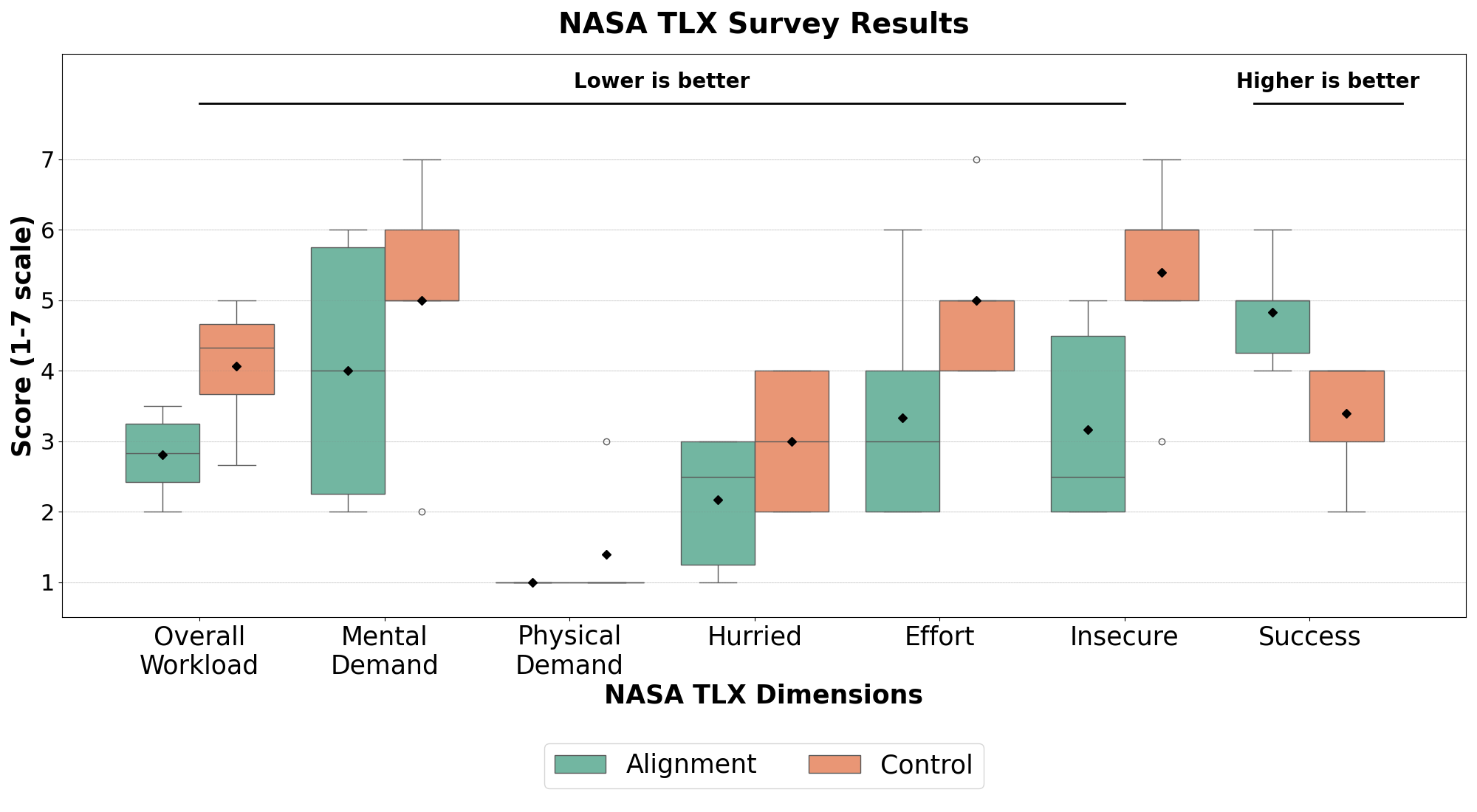}
    \caption{User study result: NASA-TLX survey across Alignment and Control conditions.}
    \label{fig:user_study_workload}
\end{figure}

For RQ2, Figure~\ref{fig:user_study_workload} summarizes the individual NASA-TLX subscales and the overall workload score, computed as the unweighted average.
Participants in the Alignment condition reported that reward tuning required less effort, caused less frustration, and was overall more successful.
% (all $p \le 0.04$)
Ultimately, these improvements led to a significantly lower overall workload score for the Alignment condition compared to Control ($t=-2.64$, $p=0.02$). This 
indicates that tuning reward functions with TAC made the process noticeably easier.

\section{Can TAC be used to learn reward weights?}
Our user study suggests that TAC can help RL practitioners manually tune reward weights. 
However, reward tuning remained a labor-intensive process: participants without TAC feedback adjusted weights an average of 50 times, while those with TAC did so 44 times.
This motivates \emph{learning} reward functions to maximize TAC instead of manual tuning. 

\subsection{Differentiable TAC}\label{sec:soft_tac}
To enable gradient-based optimization, we introduce \emph{Soft-TAC}, an approximation of TAC, denoted by $\tilde{\sigma}_{\text{TAC}}(\mathcal{D}_h, G_{r_,\gamma})$:
\begin{equation}\label{eq:soft_tac}
    \mathbb{E}_{\mathcal{D}_h}
    \Big[ y \cdot \tanh\Big(\alpha \Big( G_{r, \gamma}(\tau^{(1)}) - G_{r, \gamma}(\tau^{(2)})  \Big) \Big) \Big]
\end{equation}
where $y \in \{-1, 1\}$ is the preference label ($y = 1$ for $\tau^{(1)} \succ \tau^{(2)}$ and $-1$ for $\tau^{(2)} \succ \tau^{(1)}$),
and $\alpha$ controls the sensitivity of the objective to differences in returns. 

\begin{restatable}{proposition}{SoftTacConvergenceTac}
\label{soft_tac_convergence_tac}
As $\alpha \to \infty$, $\tilde{\sigma}_{\text{TAC}}(\mathcal{D}_h, G_{r, \gamma})$ recovers the Trajectory Alignment Coefficient.
\end{restatable}
See proof in the appendix. To apply Soft-TAC to reward learning, we optimize a parameterized reward model $\hat{r}_\theta$ by minimizing its negative Soft-TAC score, encouraging the model to assign higher returns to preferred trajectories:
\begin{equation}\label{eq:soft_tac_loss}
\mathcal{L}_{\mathrm{Soft-TAC}} = 1 - \tilde{\sigma}_{\mathrm{TAC}}(\mathcal{D}_h, G_{r_{\theta},\gamma})
\end{equation}
Equation \ref{eq:soft_tac_loss} provides a generic loss formulation suitable for both online and offline reward learning methodologies.

% Our choice of a $\tanh$-based relaxation of Kendall’s Tau is motivated by \citet{zheng2023diffkendall}, where a differentiable Kendall’s $\tau$ was used as a similarity metric to improve image classification. In contrast, Soft-TAC uses Kendall’s Tau directly as a loss function to align reward models with human preferences. We provide a detailed analysis of these structural differences in the appendix.

Our choice of a $\tanh$-based relaxation of Kendall’s Tau is motivated by \citet{zheng2023diffkendall}, who applied a similar differentiable metric to image classification. 
In contrast, Soft-TAC is the first Kendall’s Tau–derived loss function, which we apply to align reward models with human preferences. A detailed analysis of the formulation and optimization differences is provided in the appendix.

% While both methods share a $\tanh$ relaxation for differentiability, they diverge in formulation and optimization role. \citet{zheng2023diffkendall} uses Kendall’s $\tau$ as a similarity metric to improve image classification. In contrast, Soft-TAC uses  Kendall’s $\tau$ directly as a loss function to align reward models with human preferences. We provide a detailed analysis of these structural differences in the appendix.

% While both methods share a $\tanh$ relaxation for differentiability, they diverge in formulation and optimization role. \citet{zheng2023diffkendall} uses Kendall’s $\tau$ as a similarity metric to improve image classification. In contrast, Soft-TAC uses  Kendall’s $\tau$ directly as a loss function to align reward models with human preferences. We provide a detailed analysis of these structural differences in the appendix.
Next, we discuss two theoretical properties of Soft-TAC.
% \subsection{Theoretical Results of Soft-TAC}
% \begin{definition}\label{def:r_human}
% Let $\mathcal{R}_{\mathrm{human}}$ be the set of human-aligned reward functions $r$ such that for every comparison $(\tau^i, \tau^j, y) \in \mathcal{D}_h$, the condition $
% y \cdot \big(G_r(\tau^i) - G_r(\tau^j)\big) > 0 $
% is satisfied.
% \end{definition}

% \begin{definition}\label{def:r_human}
% Let $\mathcal{R}_{\mathrm{human}}$ be the set of human-aligned reward functions $r$ such that for every comparison $(\tau_k^{(1)}, \tau_k^{(2)}, y_k) \in \mathcal{D}_h$, the condition $
% y_k \cdot \big(G_r(\tau_k^{(1)}) - G_r(\tau_k^{(2)})\big) > 0 $
% is satisfied.
% \end{definition}

% \begin{definition}\label{def:r_human}
% Let $\mathcal{R}_{\mathrm{human}}$ be the set of human-aligned reward functions $r$ such that
% \begin{equation}
%     y_k \cdot \big(G_{r, \gamma}(\tau_k^{(1)}) - G_{r, \gamma}(\tau_k^{(2)})\big) > 0,  \forall (\tau_k^{(1)}, \tau_k^{(2)}, y_k) \in \mathcal{D}_h.
% \end{equation}
% \end{definition}

\begin{definition}\label{def:r_human}
Let $\mathcal{R}_{\mathrm{human}}$ be the set of human-aligned reward functions $r$ such that for all $(\tau_k^{(1)}, \tau_k^{(2)}, y_k) \in \mathcal{D}_h$, 
\begin{equation}
    y_k \cdot \Big(G_{r, \gamma}\big(\tau_k^{(1)}\big) - G_{r, \gamma}\big(\tau_k^{(2)}\big)\Big) > 0.
\end{equation}
\end{definition}

% \begin{assumption}\label{ass:pref_structure}$\mathcal{D}_h$ is noise-free.
% \end{assumption}

% \begin{assumption}\label{ass:realizability}
% The hypothesis class $\{r_\theta\}_{\theta \in \Theta}$ is sufficiently expressive such that there exists at least one parameter vector $\theta^* \in \Theta$ such that the reward function $r_{\theta^*} \in \mathcal{R}_{\mathrm{human}}$.
% \end{assumption}
\begin{assumption}\label{ass:realizability}
The hypothesis class $\{r_\theta\}_{\theta \in \Theta}$ contains at least one parameter vector $\theta^* \in \Theta$ such that $r_{\theta^*} \in \mathcal{R}_{\mathrm{human}}$.
\end{assumption}
Note that Assumption \ref{ass:realizability} is common for theoretical analysis in preference learning \cite{PrincipledRLHF}.

\begin{restatable}{theorem}{globalmin}\label{theorem_global_min}
Let $\mathcal{D}_h$ be noise-free. Under Assumption \ref{ass:realizability} and as $\alpha \to \infty$, a reward function $r_{\theta^*}$ is a global minimizer of the Soft-TAC loss if and only if $r_{\theta^*} \in \mathcal{R}_{\mathrm{human}}$.
\end{restatable}
Theorem \ref{theorem_global_min} establishes that under noise-free preferences, the solution set of Soft-TAC is complete (it contains all human-aligned reward functions) and minimal (it cannot be expanded without treating misaligned reward functions as optimal).
While Theorem \ref{theorem_global_min} assumes noise-free preferences, we now relax this assumption to consider \textbf{simple non-uniform label noise}~\cite{ghosh2017robust}. This accounts for realistic settings where the label noise is dependent on the specific trajectories being compared. For example, an evaluator may easily distinguish starkly different trajectories (low label noise) but struggle to differentiate highly similar ones (high label noise).
Theorem \ref{noise_tolerance} demonstrates that Soft-TAC is provably noise-tolerant, \emph{meaning its optimal solution is unaffected by the label noise}.
\begin{restatable}{theorem}{noisetolerance}\label{noise_tolerance}
We consider label noise where the noise rate $\eta_{\mathbf{x}}$ depends on the input $\mathbf{x}$ and satisfies $\eta_{\mathbf{x}} < \frac{n-1}{n}$ for all $\mathbf{x} \in \mathcal{X}$, with $n$ denoting the number of classes.  Under this condition and as $\alpha \to \infty$, $\mathcal{L}_{\text{Soft-TAC}}$ is noise-tolerant: the minimizer of the expected risk under noisy labels is identical to that under clean labels ($\theta^*_{\text{noisy}} = \theta^*_{\text{clean}}$).
\end{restatable}
\begin{definition}\label{def:symmetric_loss}
A loss function $L$ is symmetric if for any prediction $\hat{y}$, $\sum_{y=1}^{n} L(\hat{y}, y) = C$ for some constant $C$, where $n$ is the number of classes.
\end{definition}
Theorem 2 leverages the work of \citet{ghosh2017robust}, which proved that symmetric loss functions (Definition \ref{def:symmetric_loss}) can be noise-tolerant under certain conditions. Because Soft-TAC satisfies this symmetry property, it meets one of the core requirements ensuring that Theorem \ref{noise_tolerance} holds. Cross-Entropy, however, does not possess this property; consequently its noise tolerance remains unguaranteed. See the appendix for the proofs of both theorems.

Beyond theoretical guarantees, Soft-TAC's symmetry property provides a practical mechanism for robustness against errors. 
In practice, when the reward model predicts a large difference in the wrong direction (i.e., $G_{r_{\theta}, \gamma}(\tau^i) - G_{r_{\theta},\gamma}(\tau^j) \gg 0$ but $\tau^j$ is preferred over $\tau^i$), the Soft-TAC loss saturates and its gradient approaches zero. While this behavior might seem undesirable, it prevents wasted optimization on highly noisy preferences or those impossible for the model to fit—such as those resulting from human error or hidden context not captured by the reward features.

In contrast, Cross-Entropy is unbounded; therefore, its loss never saturates when a mismatched prediction occurs. Instead, its gradient remains large and non-zero. This can lead to conflicting gradient updates that cause the optimization to oscillate and converge to suboptimal solutions. We explicitly observed this phenomenon with the Cross-Entropy loss in the GT7 experiments and in a toy example in the appendix.

\subsection{Experiments and Results -- Open-Source Domains}\label{sec:experimental_design}
We first evaluate whether optimizing a linear reward model with Soft-TAC can recover effective weights in Lunar Lander. We then test its viability when training non-linear, black-box reward models for OpenAI Gym Hopper and Reacher.
\begin{enumerate}[label=\textbf{RQ\arabic*:}, leftmargin=*]
    \setcounter{enumi}{2}
    \item Can we learn effective linear and black-box reward models from human preferences using Soft-TAC?
\end{enumerate}

\subsubsection{Shared Training and Evaluation Pipeline}
To isolate the impact of the \emph{loss function} on PbRL, we employ the Cross-Entropy loss (Equation \ref{eq:cross_entropy}) combined with the BT model (Equation \ref{eq:bradley_terry}) as our primary baseline. This configuration is widely adopted by most PbRL algorithms, as discussed in the Related Work section. By stripping away confounding mechanisms from other algorithms, we ensure a direct comparison of the loss functions themselves. Additionally, we evaluate the environment's default reward function only as an upper bound for performance.

In all domains, we train our reward models \emph{exclusively on human preferences} using the Adam optimizer \citep{kingma2015adam}, in contrast to standard PbRL works that rely on synthetic preferences \citep{muslimani2025sdp}. Hyperparameters are selected via grid search over learning rates and batch sizes, applying early stopping to prevent overfitting. To evaluate each loss function, we adopt the evaluation protocol as described in the User Study Evaluation section. We perform reward learning across $S=10$ random seeds (each yielding a distinct reward model), followed by RL training with SAC. For each seed, we pool $K$ policy checkpoints near convergence and evaluate each deterministically over $E$ episodes. Specifically, we set $K=E=5$ for Lunar Lander, and $K=E=10$ for both Hopper and Reacher. We report the mean performance $\pm$ standard error across the 10 seeds. 
See the appendix for training hyperparameters.
% --- ENVIRONMENT 1: LUNAR LANDER ---
\subsubsection{Lunar Lander} Following the setup of our user study, the reward model is parameterized as a linear function of the environment's state features. We train this model on the preference dataset collected for the user study, which contains 148 pairwise preferences annotated by a single human. Performance in this environment is measured by the landing-pad success rate.

% \subsubsection{Lunar Lander Results}
 We found that Soft-TAC performed comparably to Cross-Entropy regarding landing-pad success rate, with a mean and standard error of $0.70 \pm 0.12$ versus $0.50 \pm 0.17$, respectively ($U = 57.5$, $p_{\text{adj}} = 0.29$). 
  For reference, the default environment reward achieved an landing-pad success rate of $1.00 \pm 0.00$. In a secondary analysis, we observed a clear trend toward safer landings outside the target pad for Soft-TAC compared to Cross-Entropy ($0.14 \pm 0.07$ vs. $0.00 \pm 0.00$; $U = 65.0, p_{\text{adj}} = 0.08$).
Finally, Soft-TAC achieved performance on par with participant-tuned reward weights that used TAC ($U = 32.5, p_{\text{adj}} = 0.41$), while significantly outperforming weights tuned without TAC ($U = 42.0, p_{\text{adj}} = 0.04$). This demonstrates that reward learning offers a viable alternative to manual tuning.
 
% --- ENVIRONMENT 2 & 3: MUJOCO ---
\subsubsection{Hopper and Reacher} % Specific Setup}
In these environments, we train a \emph{black-box reward model} parameterized by a deep neural network that maps raw state-action vectors directly to a scalar reward. This removes the need for manual feature engineering or assuming a known linear relationship. 

Furthermore, rather than relying on a single annotator, we leverage an open-source rating dataset collected from 6–7 human annotators \cite{rmse}. We convert the scalar ratings into pairwise preferences, yielding an average of $9,900$ and $5,700$ pairs per annotator for Reacher and Hopper, respectively. 
To evaluate Soft-TAC's performance across a diverse set of human preference distributions, we train a separate reward model per annotator and report performance averaged across all participants. Performance is measured by the final environment return.

% \begin{table}[t]
% \centering
% \caption{Shows the mean landing-pad success rates ($\pm$ SE) in Lunar Lander after RL training with each reward function (higher is better). }\label{tab:lunar_lander}

% \begin{tabular}{l c}
% \toprule
% \textbf{Method} & \textbf{Landing-pad Success Rate} \\
% \midrule
% \textsc{Soft-TAC} & 0.72 $\pm$ 0.16 \\
% \textsc{Cross-Entropy} & 0.6 $\pm$ 0.24 \\
% \textsc{Default Reward} & 1.00 $\pm$ 0.0\\

% \bottomrule
% \end{tabular}
% \end{table}

% \subsubsection{Reacher and Hopper Results}
In Reacher, Soft-TAC trained reward models achieved a mean final return of $-23.26 \pm 4.41$ (SE) compared to $-30.58 \pm 5.92$ for Cross-Entropy, demonstrating a strong positive trend toward higher final performance (Wilcoxon signed-rank $W = 24.0$, $p = 0.055$). In Hopper, Soft-TAC performed comparably to Cross-Entropy, achieving a mean final return of $1010.92 \pm 41.01$ (SE) versus $991.20 \pm 56.40$ (paired $t$-test $t = 0.26$, $p = 0.401$). For comparison, policies trained directly on the true environment rewards achieved mean final returns of $-3.46 \pm 0.03$ in Reacher and $2540.21 \pm 213.76$ in Hopper.

\subsection{Experiments and Results -- Gran Turismo 7}\label{sec:GT_experiments_results}
Next, we consider GT7, a high-fidelity racing simulator characterized by diverse car models and realistic car–track dynamics. Developed by Polyphony Digital, Inc, this domain represents a significant increase in complexity over the open-source domains.
In particular, we consider two types of racing tasks: (1) \emph{time-trial}, where the agent races alone with the objective of completing the track as quickly as possible, and (2) \emph{versus}, where the agent races against 19 opponents.

We selected this domain because driving preferences are inherently subjective, making it an ideal testbed for evaluating whether our method can capture \emph{distinct driving styles from human preferences}. We formalize our investigation through the following research questions:

% Can we learn preference-specific reward models with Soft-TAC that produce distinct driving behaviors?

\begin{enumerate}[label=\textbf{RQ\arabic*:}, leftmargin=*]
    \setcounter{enumi}{3}
    \item Can Soft-TAC learned reward models induce a \emph{fast} driving style in the time-trial task?
        \item Can they also induce \emph{aggressive} and \emph{timid} driving behaviors in the versus setting?
\end{enumerate}

\begin{figure*}[t]
    \centering
    \begin{subfigure}[b]{0.45\textwidth}
        \centering
        \includegraphics[width=8cm, height=4.2cm]{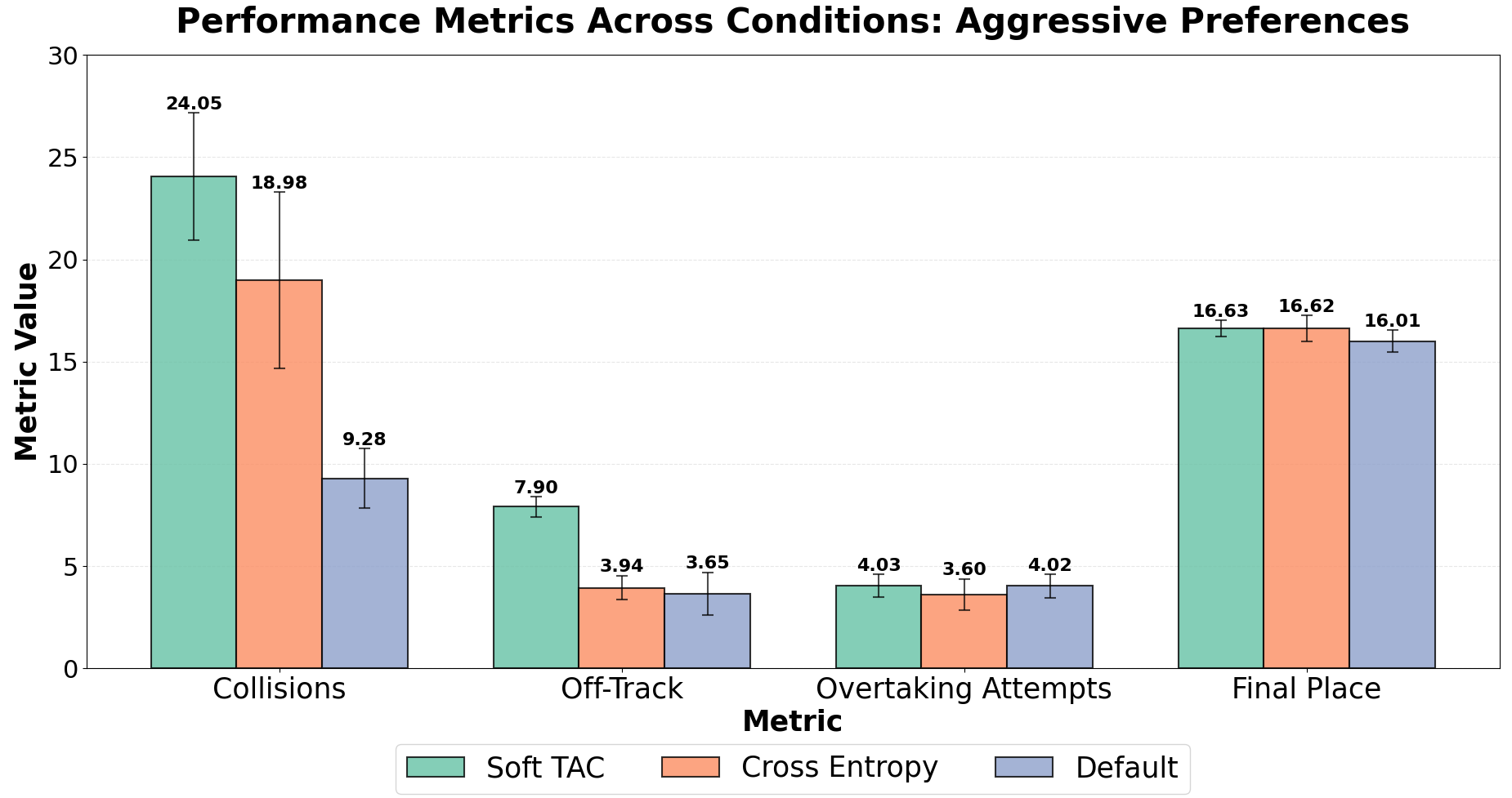}
        \caption{}
        \label{fig:versus_task_aggressive}
    \end{subfigure}
    \hfill
    \begin{subfigure}[b]{0.45\textwidth}
        \centering
        \includegraphics[width=8cm, height=4.2cm]{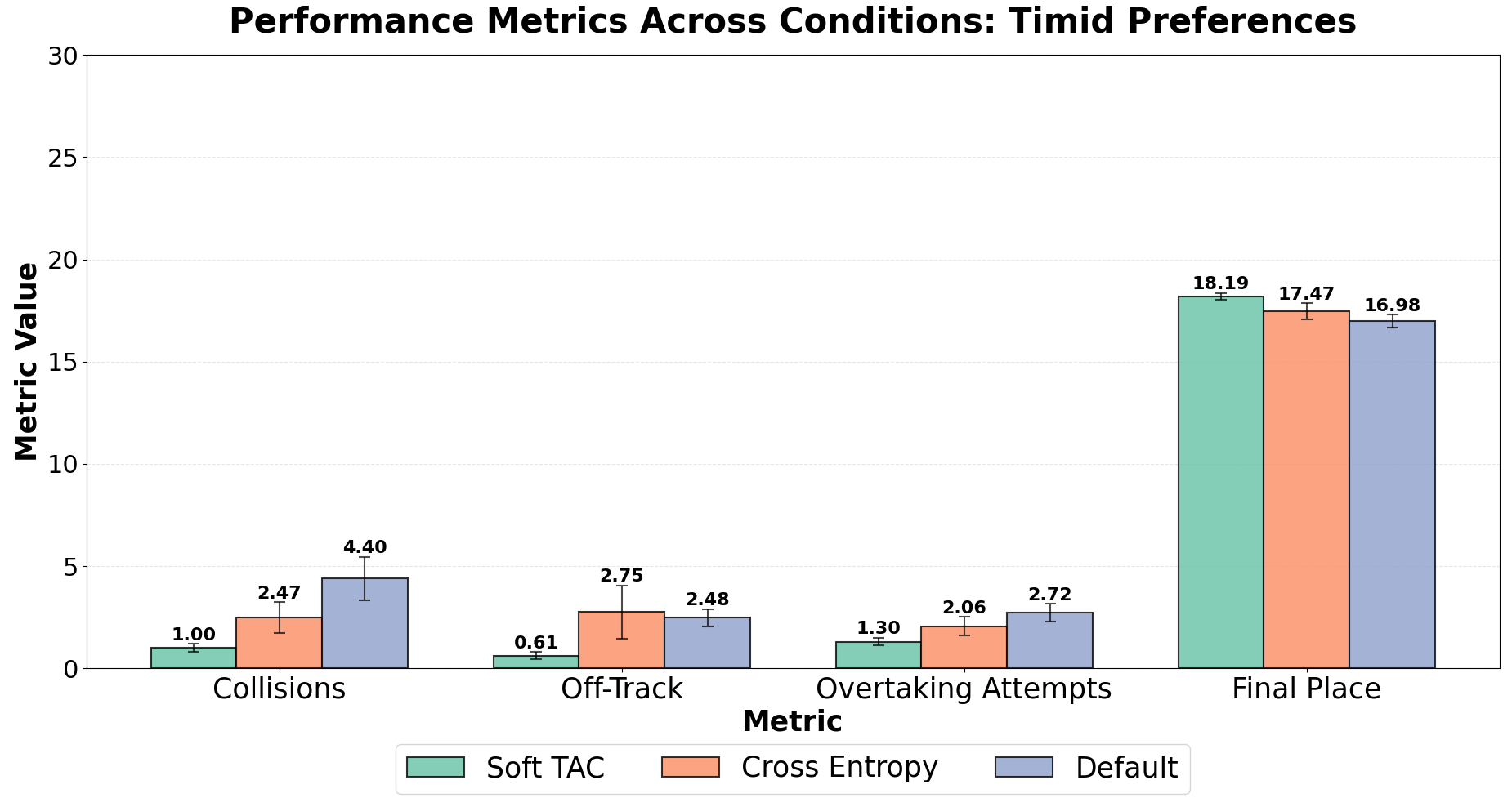} % 
        \caption{}
        \label{fig:versus_task_timid}
    \end{subfigure}

    \caption{GT7 results for the versus tasks, illustrating how aggressive (a) or timid (b) each agent is (per the mean $\pm$ SE of the four metrics) when trained with different reward functions. For aggressive, higher metrics are better, and for timid, lower metrics are better (except for final place).}
    \label{fig:versus_task_results}
\end{figure*}

\subsubsection{Preference Data Collection}

For \textbf{RQ4}, we aimed to learn a reward function that explicitly prioritizes speed.  To gather preference data, we first collected 56 time-trial trajectories generated by the built-in AI (BIAI) agents, which are pre-programmed opponent drivers in the game.
We then randomly sampled pairs of trajectories from this pool and labeled them according to their lap times. A shorter lap time is preferred, indicating faster driving. This process resulted in a total of 1429 preferences. 

Next, for \textbf{RQ5}, our goal was to learn reward models for \emph{aggressive} and \emph{timid} driving styles in the versus task. 
Aggressive driving is characterized by more collisions and off-track instances, which may lead to more frequent overtaking attempts. In contrast, timid driving reflects a tendency to keep a distance from other cars, resulting in fewer such instances. 
% To form the preference dataset, we first trained RL agents using slightly modified reward functions from GT Sophy \citep{wurman2022outracing}, such that they exhibited different driving behaviors than GT Sophy. Then, we collected 79 versus-task trajectories from the trained agents and organized them into groups by car.  
To form the preference dataset, we trained a style-conditioned UVFA RL agent to induce different driving behaviors \citep{capobianco2026coachableagentsinteractivegameplay}. We then collected 79 versus-task trajectories and organized them by car.
%Then, to elicit preferences, we organized trajectories into groups by car model. The authors then 
For each car group, we expressed preferences—ranging from aggressive to timid—based on the behavioral indicators described above. This process resulted in 118 pairwise preferences. 
% Further details on the collection process and the metric definitions are provided in Appendix \ref{sec:preference_collection_details}.
See the appendix for details on metrics and the preference collection process.

All collected trajectories were driven on Circuit de la Sarthe: a 13.6 km high-speed circuit that requires slipstreaming on long straights and managing car downforce. To ensure our datasets captured diverse driving dynamics, the agents generating these trajectories drove a subset of 19 cars drawn from the 552 available in GT7. These cars were specifically chosen to span a wide range of Performance Points (PP)---an in-game metric measuring car competitiveness, where a higher PP indicates a more powerful and faster car.

\subsubsection{Training}
We followed the same reward learning protocol and baseline comparisons as in the previous experiments. Here, the reward model is structured as a linear combination of features. To focus on the specific behaviors reflected in the preference data, we restrict learning to a relevant subset of these reward features.
For \textbf{fast driving (RQ4)}, we consider the weight for the \emph{progress} feature. This measures distance advanced along the track centerline; the primary driver of speed. For \textbf{aggressive/timid styles (RQ5)}, we consider the weights associated with the \emph{overtaking and collision} features, as these are tied to interactive driving behaviors.
All other reward weights remained fixed.

We performed the reward learning procedure across six seeds. For each resulting reward model, we trained a new QR-SAC agent \citep{wurman2022outracing} on the Sarthe track. Note that although the preference data was collected from a subset of 19 cars, we train the policy to control all 552 cars in the game.
For RQ4, training was done in the time-trial task, where a single car attempts to complete laps as quickly as possible. For RQ5, we trained with various versus tasks as described in \citet{wurman2022outracing} where the agent races against a range of opponents.

% To RL training, 
% To address our generalization research question (\textbf{RQ5}), we designed the training setup to differ significantly from the data collection phase. Crucially, all agents were trained in a multi-car setting featuring over 500 distinct vehicles, whereas the preference datasets contained only 19.

% Additionally, for the lap time agents, we introduced track-level generalization. While aggressive and timid agents were trained on the Sarthe track, lap time agents were trained on Autodrome Lago Maggiore. We chose the Maggiore track, as its technical layout—requiring complex cornering sequences and elevation management—contrasts with the high-speed, longitudinal nature of Sarthe. This setup allows us to evaluate generalization across unseen cars for all tasks, and specifically for the lap time experiment, generalization to a track different from the one used to collect preferences.
\paragraph{Evaluation}
To select policies for evaluation, we periodically monitored performance during training and identified those policies that best optimized the metrics associated with the intended driving style (e.g., maximizing collisions for aggressive agents). To ensure a fair comparison, this selection criterion was applied uniformly to all baselines.
Selected policies were then evaluated across all 552 cars on Sarthe. 

Given that driving styles are inherently multi-faceted, relying on a single success metric is insufficient. Instead, we evaluate policy performance across a set of metrics specifically tailored to each driving objective. For the fast driving experiments, evaluation was conducted in a time-trial setting where the agent completes a solo three-lap race (retaining only the fastest lap). Performance is assessed across three metrics, where lower scores indicate greater proficiency: (1) BIAI ratio, defined as the agent's lap time divided by the BIAI's lap time for the same car (values below 1.0 indicates outperforming the BIAI), (2) lap time in seconds, and (3) number of incomplete laps (i.e.,  lap time $\ge 1,800$ seconds).

For the aggressive/timid experiments, we evaluate performance in a specific versus task, where the agent starts in last place and chases 19 BIAI opponents over a three-lap race. We assess these behaviors through four metrics: collisions, off-track instances, overtaking attempts, and final position. Higher values in the first three metrics typically indicate an aggressive, high-risk driving style, whereas lower values reflect a cautious, risk-averse approach. Final position measures overall racing performance.

All metrics are averaged over the 552 cars per seed, followed by an average across seeds.
% We evaluate statistical significance using Student's $t$-tests when normality assumptions are met and Mann–Whitney $U$-tests otherwise. 
For further details on training and evaluation, see the appendix.

% \begin{table}[t]
% \centering
% \caption{Comparison of methods on three metrics.}
% \label{tab:method_three_metrics}
% \begin{tabular}{l ccc}
% \toprule
% \textbf{Method} & \textbf{BIAI Ratio} & \textbf{lap time (sec.)} & \textbf{Fastest lap time Rate} \\
% \midrule
% \textsc{Soft-TAC} & 0.00 & 0.00 & 0.00 \\
% \textsc{Cross Entropy} & 0.00 & 0.00 & 0.00 \\
% \textsc{Default Reward} & 0.00 & 0.00 & 0.00 \\
% \bottomrule
% \end{tabular}
% \end{table}

\begin{table}[t]
\centering
\caption{GT fast driving results in the time-trial task. Table shows mean $\pm$ SE of relevant metrics. Lower is better.}
\label{tab:GT_control_lap_time}
\resizebox{\linewidth}{!}{%
\begin{tabular}{l ccc}
\toprule
\textbf{Method} & \textbf{BIAI} & \textbf{Lap} & \textbf{Incomplete} \\
 & \textbf{Ratio} & \textbf{Time} & \textbf{Laps} \\
\midrule
\textsc{Soft-} & 0.971 $\pm$ 0.003 & 270.72 $\pm$ 0.73 & 1.17 $\pm$ 0.40 \\
\textsc{TAC} & & & \\
\textsc{Cross-} & 0.996 $\pm$ 0.001 & 277.71 $\pm$ 0.31 & 11.00 $\pm$ 1.95 \\
\textsc{Entropy} & & & \\
\textsc{Default} & 0.991 $\pm$ 0.001 & 276.03 $\pm$ 0.21 & 5.00 $\pm$ 1.44 \\
\bottomrule
\end{tabular}%
}
\end{table}
\subsubsection{Results: Controlling Fast Driving Styles}

As shown in Table~\ref{tab:GT_control_lap_time}, Soft-TAC’s learned reward model substantially improved performance over the Cross-Entopry and default reward baselines. Specifically, it significantly reduced lap times ($U=0.0$, $p_{\text{adj}}=0.003$ vs. CE; $U=0.0$, $p_{\text{adj}}=0.003$ vs. default) and BIAI ratios ($U=0.0$, $p_{\text{adj}}=0.003$ vs. CE; $U=0.0$, $p_{\text{adj}}=0.003$ vs. default), while also averaging fewer incomplete laps ($U=0.0$, $p_{\text{adj}}=0.003$ vs. CE; $U=4.5$, $p_{\text{adj}}=0.02$ vs. default).
This performance gap directly reflects differences in the learned progress reward weights: while default uses $1.0$, Cross-Entropy and Soft-TAC converged to $0.5$ and $7.0$, respectively. 
This difference stems from how the losses handled a subset of misclassified preferences. Under Cross-Entropy, large gradients from these errors offset the gradients from the correctly classified points, resulting in a net gradient near zero that led the weight to plateau. Conversely, Soft-TAC suppressed the gradients of these errors, leaving only the signal from the correct data to increase the weight.

\subsubsection{Results: Controlling Aggressive and Timid Behavior}
Policies trained with Soft-TAC reward models exhibited noticeably heightened aggression (Figure~\ref{fig:versus_task_aggressive}). Soft-TAC generated significantly more off-track instances than both Cross-Entropy ($t=5.20$, $p_{\text{adj}}=0.002$) and the default reward ($U=33.0$, $p_{\text{adj}}=0.02$), as well as significantly more collisions than the default reward ($t=4.30$, $p_{\text{adj}}=0.007$). However, this aggression did not translate to a significant increase in overtaking attempts or an improved final placement. This illustrates the double-edged nature of aggressive racing: while crowding opponents and cutting corners reflect an assertive strategy, heavy contact can trigger spin-outs, and running off-track onto low-traction surfaces can cause speed loss that ultimately offsets any tactical gains.
These behavioral differences trace back to the learned reward weights. Compared to the default reward, which weights overtaking and collisions equally at 0.5, both Soft-TAC and Cross-Entropy learned near-zero collision weights, effectively removing contact penalties. However, Soft-TAC learned a substantially larger overtaking weight ($2.7$) than Cross-Entropy ($1.4$), driving the policy toward higher-risk maneuvers.

% A similar pattern holds in the timid condition (Figure~\ref{fig:versus_task_timid}). In this setting, both Soft-TAC and Cross-Entropy learned a zero weight for overtaking, eliminating any incentive to pass. However, Soft-TAC assigned a much higher collision penalty ($2.4$) than Cross-Entropy ($0.8$) and the default reward ($0.5$), resulting in a markedly conservative driving style. Relative to the default reward, Soft-TAC agents produced significantly fewer overtaking attempts ($U=4.0$, $p=0.01$), collisions ($U=2.0$, $p=0.004$), and off-track instances ($U=1.0$, $p=0.002$). As a direct consequence of this cautious driving, Soft-TAC agents finished in a significantly lower final place ($U=34.0$, $p=0.004$). Furthermore, while comparisons against Cross-Entropy showed marginal trends toward fewer overtaking attempts ($U=9.0$, $p=0.09$) and collisions ($U=10.0$, $p=0.12$), Soft-TAC produced significantly fewer off-track instances ($U=6.0$, $p=0.03$) and finished in a significantly lower final place ($U=29.0$, $p=0.05$).

A similar pattern holds in the timid condition (Figure~\ref{fig:versus_task_timid}). In this setting, both Soft-TAC and Cross-Entropy learned a zero weight for overtaking, eliminating any incentive to pass. However, Soft-TAC assigned a much higher collision penalty ($2.4$) than Cross-Entropy ($0.8$) and the default reward ($0.5$), resulting in a markedly conservative driving style. Relative to the default reward, Soft-TAC agents produced significantly fewer overtaking attempts ($t=-3.04$, $p_{\text{adj}}=0.02$), collisions ($t=-3.16$, $p_{\text{adj}}=0.02$), and off-track instances ($t=-4.06$, $p_{\text{adj}}=0.02$). As a direct consequence of this cautious driving, Soft-TAC agents finished in a significantly worse final place ($t=3.47$, $p_{\text{adj}}=0.02$). Furthermore, compared to Cross-Entropy, Soft-TAC showed a strong trend toward a worse final place alongside fewer off-track instances, overtaking attempts, and collisions ($0.052 \le p_{\text{adj}} \le 0.086$).

Lastly, to evaluate whether Soft-TAC successfully translates different preference profiles into distinct policies, we directly compared agents trained under aggressive versus timid reward functions. As expected, agents trained under the aggressive Soft-TAC reward function exhibited a significantly more competitive driving profile than their timid Soft-TAC counterparts, attempting far more overtaking maneuvers ($t = 4.67$, $p_{\text{adj}} = 0.002$) and achieving a significantly better final race placement ($t = -3.67$, $p_{\text{adj}} = 0.004$). However, this heightened competitiveness entailed substantially higher risk-taking: aggressive agents incurred significantly more collisions ($t = 7.40$, $p_{\text{adj}} < 0.001$) and spent far more time off-track ($t = 13.87$, $p_{\text{adj}} < 0.001$).
Comprehensive pairwise statistical comparisons for all GT7 experiments are provided in the appendix.
Together, these results demonstrate that Soft-TAC successfully learns reward functions capable of inducing significantly different driving behaviors.

\section{Conclusion}
Designing effective reward functions is challenging, motivating our investigation into tools that better support this process. In a human-subject study, we found that TAC meaningfully assists RL practitioners in specifying reward weights: participants using TAC produced more performant reward functions while experiencing lower cognitive workload. However, the study also highlighted that manual tuning, even with TAC, remains labor-intensive, underscoring the need for automated approaches.
To this end, we introduce Soft-TAC, a differentiable approximation used to train reward models so they agree as closely as possible with human preferences. Across evaluations spanning four environments, Soft-TAC-trained reward models performed comparably to or outperformed competing baselines. Furthermore, our theoretical analysis guarantees that the optimization objective is complete, minimal, and robust to noisy human preferences.
Overall, our findings show that TAC serves as both an interactive tool for reward tuning and a robust objective for reward learning. Future work can extend this framework in two key directions: first, expanding Soft-TAC to support pluralistic alignment to reconcile diverse preferences across multiple domain experts; second, adapting TAC to analyze reward alignment at the feature level to pinpoint precisely which components of a reward function are misaligned.
% \section*{Acknowledgments}
% AAAI is especially grateful to Peter Patel Schneider for his work in implementing the original aaai.sty file, liberally using the ideas of other style hackers, including Barbara Beeton. We also acknowledge with thanks the work of George Ferguson for his guide to using the style and BibTeX files --- which has been incorporated into this document --- and Hans Guesgen, who provided several timely modifications, as well as the many others who have, from time to time, sent in suggestions on improvements to the AAAI style. We are especially grateful to Francisco Cruz, Marc Pujol-Gonzalez, and Mico Loretan for the improvements to the Bib\TeX{} and \LaTeX{} files made in 2020.

% The preparation of the \LaTeX{} and Bib\TeX{} files that implement these instructions was supported by Schlumberger Palo Alto Research, AT\&T Bell Laboratories, Morgan Kaufmann Publishers, The Live Oak Press, LLC, and AAAI Press. Bibliography style changes were added by Sunil Issar. \verb+\+pubnote was added by J. Scott Penberthy. George Ferguson added support for printing the AAAI copyright slug. Additional changes to aaai2027.sty and aaai2027.bst have been made by Francisco Cruz, Marc Pujol-Gonzalez, and Mico Loretan.

% \bigskip
% \noindent Thank you for reading these instructions carefully. We look forward to receiving your electronic files!

\bibliography{aaai2027}

% \appendix
% \input{appendix}

\clearpage
\onecolumn
\appendix

% \documentclass{article}
% \usepackage{AAAI_AuthorKit27/aaai2027}
% \usepackage{graphicx}
% % 2. Force single-column layout
% \onecolumn
% % Essential packages for mathematical formatting, layout, and tables
% \usepackage{amsmath, amsfonts, amssymb, amsthm}
% \usepackage{booktabs}

% % Theorem and environment definitions
% \newtheorem{theorem}{Theorem}
% \newtheorem{proposition}{Proposition}

% \begin{document}

\section{Theoretical Appendix}

\section{Preliminaries: Data Distributions, Reward Sets, and Expected Risk}

We first establish the foundational concepts and notation useful for understanding the theory.

\paragraph{Preference Concordance}
A reward function (with discounting) has the same preference as a human on a trajectory pair if the sign of the return difference matches the human label, satisfying $\operatorname{sgn}\left(\Delta G_{r, \gamma}(\tau^{(1)}, \tau^{(2)})\right) = y$, which can be also written as the strict inequality $y \cdot \Delta G_{r, \gamma}(\tau^{(1)}, \tau^{(2)}) > 0$. If this is true, the trajectory pair is considered \emph{concordant}. Conversely, if the return difference contradicts the human label, satisfying $\operatorname{sgn}\left(\Delta G_{r, \gamma}(\tau^{(1)}, \tau^{(2)})\right) \neq y$ (or equivalently $y \cdot \Delta G_{r, \gamma}(\tau^{(1)}, \tau^{(2)}) < 0$), the pair is considered \emph{discordant}.

\paragraph{Data Distributions and the $i.i.d.$ Assumption:} 
We consider our collected human preference dataset, $\mathcal{D}_h = \{(\tau_k^{(1)}, \tau_k^{(2)}, y_k)\}_{k=1}^N$, to be a finite sample of $N$ observations drawn \emph{i.i.d.} from an underlying, true, infinite population distribution, denoted by $\mathcal{D}_h^{\mathrm{pop}}$. Mathematically, $\mathcal{D}_h^{\mathrm{pop}}$ represents the joint probability distribution over the data space $\mathcal{X} \times \mathcal{Y}$ that generates these samples, where $\mathcal{X}$ encompasses the domain of trajectory pairs $(\tau^{(1)}, \tau^{(2)})$ and $\mathcal{Y} = \{-1, 1\}$ represents the binary label space. 

\paragraph{Empirical vs. Population Aligned Reward Sets:}
We define $\mathcal{R}_{\mathrm{human}}$ as the empirical set of reward functions whose induced preferences perfectly match the human preferences captured within our finite sample $\mathcal{D}_h$:
\[
\mathcal{R}_{\mathrm{human}} = \left\{ r \;\middle|\; \operatorname{sgn}\left(\Delta G_{r, \gamma}(\tau_k^{(1)}, \tau_k^{(2)})\right) = y_k, \; \forall (\tau_k^{(1)}, \tau_k^{(2)}, y_k) \in \mathcal{D}_h \right\}.
\]
Extending this concept to the population level, we define $\mathcal{R}_{\mathrm{human}}^{\mathrm{pop}}$ as the set of all reward functions $r$ that perfectly align with the true human preference distribution. That is, a randomly drawn trajectory pair and label from $\mathcal{D}_h^{\mathrm{pop}}$ will match the reward function's induced preference with a probability ($\mathbb{P}$) of exactly $1$:
\[
\mathcal{R}_{\mathrm{human}}^{\mathrm{pop}} = \left\{ r \;\middle|\; \mathbb{P}_{(\tau_k^{(1)}, \tau_k^{(2)}, y_k) \sim \mathcal{D}_h^{\mathrm{pop}}} \Big[ y_k \cdot \Delta G_{r, \gamma}(\tau_k^{(1)}, \tau_k^{(2)}) > 0 \Big] = 1 \right\}.
\]

\paragraph{Expected Risk Framework:}
While empirical risk minimization optimizes parameters over the finite dataset $\mathcal{D}_h$, the noise-tolerance framework of Ghosh et al. (2017) operates on the expected risk. For any arbitrary predictor $f$, this risk is defined as the expectation of the loss function taken over the true joint population distribution $\mathcal{D}_h^{\mathrm{pop}}$:
\[
R_L(f) = \mathbb{E}_{(\mathbf{x}, y) \sim \mathcal{D}_h^{\mathrm{pop}}}[L(f(\mathbf{x}), y)].
\]
Let $f^*$ denote the global minimizer of this expected risk, such that $f^* = \arg\min_f R_L(f)$.
\paragraph{Characterizing Uniform versus Non Uniform Label Noise}

When a human preference dataset contains label noise, the learning algorithm does not have access to the clean joint population distribution $\mathcal{D}_h^{\mathrm{pop}}$ over data pairs $(\mathbf{x}, y)$, where the input $\mathbf{x} = (\tau^{(1)}, \tau^{(2)}) \in \mathcal{X}$ represents a trajectory comparison pair and $y \in \mathcal{Y} = \{-1, 1\}$ represents the true, uncorrupted preference label. Instead, observations are drawn from a noise-corrupted distribution, denoted by $\mathcal{D}_{\eta}$, over pairs $(\mathbf{x}, \hat{y})$, where $\hat{y}$ denotes the observed, noisy preference label.

The behavior of this label noise is dictated by how the corruption depends on the input context $\mathbf{x}$ and the true label $y$. Specifically, the total probability that an input $\mathbf{x}$ receives a corrupted label is given by the overall flip rate, $\eta_{\mathbf{x}} = \mathbb{P}(\hat{y} \neq y \mid \mathbf{x})$. When a flip does occur, the probability that the observed noisy label matches a specific incorrect class $j$ is given by $\bar{\eta}_{\mathbf{x}j} = \mathbb{P}(\hat{y} = j \mid \mathbf{x}, y)$.

In the most restrictive setting, the noise is termed \emph{uniform} (or symmetric) if the corruption is completely independent of both the input and the label, meaning the overall noise rate is a constant $\eta_{\mathbf{x}} = \eta$ across the entire domain $\mathcal{X}$, and satisfies $\bar{\eta}_{\mathbf{x}j} = \frac{\eta}{n-1}$ for all $j \neq y$. A much more natural and relaxed assumption is \emph{simple non-uniform noise}, where the overall noise rate $\eta_{\mathbf{x}}$ is allowed to vary arbitrarily as a function of the input across the domain $\mathcal{X}$, but the corrupted label continues to split its remaining probability uniformly across all incorrect classes, satisfying $\bar{\eta}_{\mathbf{x}j} = \frac{\eta_{\mathbf{x}}}{n-1}$ for all $j \neq y$. 

To ground these concepts, we now walk through an example in our binary preference learning setting ($n=2$). Consider a trajectory pair $\mathbf{x} = (\tau^{(1)}, \tau^{(2)}) \in \mathcal{X}$ where the true clean label is $y = 1$ (indicating a preference for $\tau^{(1)}$). If the label is corrupted with an input-dependent noise rate of $\eta_{\mathbf{x}} = 0.30$, the probability of observing the incorrect label $-1$ is $\bar{\eta}_{\mathbf{x},-1} = \frac{0.30}{2-1} = 0.30$. (Note that $\eta_{\mathbf{x}}$ and $\bar{\eta}_{\mathbf{x},-1}$ are identical here because there is only a single incorrect class available to flip to). This means there is a 30\% chance the observed label flips to $\hat{y} = -1$, and a 70\% chance it remains correctly identified as $\hat{y} = 1$.
Crucially, because the noise is non-uniform, these rates are not static; they vary across the domain $\mathcal{X}$ depending on the specific input $\mathbf{x}$ encountered.

\section{Proofs of Theoretical Results}
\SoftTacConvergenceTac*
\begin{proof}
Let $(\mathbf{x}, y) \in \mathcal{D}_h$ be an arbitrary triplet from the human preference dataset, where the input $\mathbf{x} = (\tau^{(1)}, \tau^{(2)})$ represents a pair of trajectories and $y \in \{-1, 1\}$ is its associated preference label ($1$ indicates $\tau^{(1)} \succ \tau^{(2)}$, and $-1$ otherwise). The difference in returns under the reward function $r$ with discount factor $\gamma$ is given by $\Delta G_{r, \gamma}(\mathbf{x}) \doteq G_{r, \gamma}(\tau^{(1)}) - G_{r, \gamma}(\tau^{(2)})$.

We break the analysis into two cases based on whether the reward differences have the same sign as the human preference labels:

\paragraph{Case 1: Concordant pair} 
This occurs when the difference in returns agrees with the human preference label. More specifically, $\operatorname{sign}\left(\Delta G_{r, \gamma}(\mathbf{x})\right) = y$). We evaluate the two possible subcases explicitly:
\begin{itemize}
    \item If $y = 1$ and $\Delta G_{r, \gamma}(\mathbf{x}) > 0$, then:
    \begin{align*}
    \lim_{\alpha \to \infty} y \cdot \tanh\left(\alpha \Delta G_{r, \gamma}(\mathbf{x})\right) 
    &= y \cdot \left[ \lim_{\alpha \to \infty} \tanh\left(\alpha \Delta G_{r, \gamma}(\mathbf{x})\right) \right] \quad \text{\small (by scalar linearity of limits)} \\
    &= (1) \cdot (1) \quad \text{\small (since $\lim_{z \to \infty} \tanh(z) = 1$ for positive arguments)} \\
    &= 1.
    \end{align*}
    \item If $y = -1$ and $\Delta G_{r, \gamma}(\mathbf{x}) < 0$, then:
    \begin{align*}
    \lim_{\alpha \to \infty} y \cdot \tanh\left(\alpha \Delta G_{r, \gamma}(\mathbf{x})\right) 
    &= y \cdot \left[ \lim_{\alpha \to \infty} \tanh\left(\alpha \Delta G_{r, \gamma}(\mathbf{x})\right) \right] \quad \text{\small (by scalar linearity of limits)} \\
    &= (-1) \cdot (-1) \quad \text{\small (since $\lim_{z \to -\infty} \tanh(z) = -1$ for negative arguments)} \\
    &= 1.
    \end{align*}
\end{itemize}
Thus, for any concordant pair, the product evaluates to $\lim_{\alpha \to \infty} y \cdot \tanh\left(\alpha \Delta G_{r, \gamma}(\mathbf{x})\right) = 1$.

\paragraph{Case 2: Discordant pair} 
This occurs when the difference in returns disagrees with the human preference label (i.e., $\operatorname{sign}\left(\Delta G_{r, \gamma}(\mathbf{x})\right) \neq y$). We evaluate the two possible subcases explicitly:
\begin{itemize}
    \item If $y = 1$ and $\Delta G_{r, \gamma}(\mathbf{x}) < 0$, then:
    \begin{align*}
    \lim_{\alpha \to \infty} y \cdot \tanh\left(\alpha \Delta G_{r, \gamma}(\mathbf{x})\right) 
    &= y \cdot \left[ \lim_{\alpha \to \infty} \tanh\left(\alpha \Delta G_{r, \gamma}(\mathbf{x})\right) \right] \quad \text{\small (by scalar linearity of limits)} \\
    &= (1) \cdot (-1) \quad \text{\small (since $\lim_{z \to -\infty} \tanh(z) = -1$ for negative arguments)} \\
    &= -1.
    \end{align*}
    \item If $y = -1$ and $\Delta G_{r, \gamma}(\mathbf{x}) > 0$, then:
    \begin{align*}
    \lim_{\alpha \to \infty} y \cdot \tanh\left(\alpha \Delta G_{r, \gamma}(\mathbf{x})\right) 
    &= y \cdot \left[ \lim_{\alpha \to \infty} \tanh\left(\alpha \Delta G_{r, \gamma}(\mathbf{x})\right) \right] \quad \text{\small (by scalar linearity of limits)} \\
    &= (-1) \cdot (1) \quad \text{\small (since $\lim_{z \to \infty} \tanh(z) = 1$ for positive arguments)} \\
    &= -1.
    \end{align*}
\end{itemize}
Hence, for any discordant pair, the product evaluates to $\lim_{\alpha \to \infty} y \cdot \tanh\left(\alpha \Delta G_{r, \gamma}(\mathbf{x})\right) = -1$.

\paragraph{Combining all pairs:} 
Let $N$ be the total number of trajectory pairs in $\mathcal{D}_h$. We aggregate our individual pointwise limits over the dataset via the empirical expectation:

\begin{align*}
\lim_{\alpha \to \infty} \tilde{\sigma}_{\text{TAC},\alpha}(\mathcal{D}_h, G_{r, \gamma}) 
&= \lim_{\alpha \to \infty} \mathbb{E}_{(\mathbf{x}, y) \sim \mathcal{D}_h} \left[ y \cdot \tanh\left(\alpha \Delta G_{r, \gamma}(\mathbf{x})\right) \right] \\
&= \lim_{\alpha \to \infty} \frac{1}{N} \sum_{k=1}^N y_k \cdot \tanh\left(\alpha \Delta G_{r, \gamma}(\mathbf{x}_k)\right) \\
&= \frac{1}{N} \sum_{k=1}^N \left[ \lim_{\alpha \to \infty} y_k \cdot \tanh\left(\alpha \Delta G_{r, \gamma}(\mathbf{x}_k)\right) \right] \quad \text{\small (by linearity of limits for a finite sum)}.
\end{align*}
We partition the dataset indices $\{1, \dots, N\}$ into two mutually exclusive sets: the set of concordant pairs $\mathcal{C}$ (where the individual limit evaluates to $1$) and the set of discordant pairs $\mathcal{D}$ (where the individual limit evaluates to $-1$). Letting $P = |\mathcal{C}|$ and $Q = |\mathcal{D}|$ such that $N = P + Q$, splitting the summation yields:
\begin{align*}
\frac{1}{N} \sum_{k=1}^N \left[ \lim_{\alpha \to \infty} y_k \cdot \tanh\left(\alpha \Delta G_{r, \gamma}(\mathbf{x}_k)\right) \right]
&= \frac{1}{N} \left( \sum_{c \in \mathcal{C}} (1) + \sum_{d \in \mathcal{D}} (-1) \right) \\
&= \frac{1}{N} \Big( P \cdot (1) + Q \cdot (-1) \Big) = \frac{P - Q}{N} = \sigma_{\text{TAC}}.
\end{align*}
\end{proof}

\globalmin*

\begin{proof}
Adopting the return difference notation $\Delta G_{r, \gamma}(\tau_k^{(1)}, \tau_k^{(2)})$ established in the proof of Proposition~\ref{soft_tac_convergence_tac}, we define the limiting pointwise individual loss incurred by the $k^{\text{th}}$ pair as:
\[
\ell_k \doteq \lim_{\alpha \to \infty} \left[ 1 - y_k \cdot \tanh\left(\alpha \Delta G_{r_{\theta^*}, \gamma}(\tau_k^{(1)}, \tau_k^{(2)})\right) \right].
\]

\noindent $(\impliedby)$ Let $r_{\theta^*} \in \mathcal{R}_{\mathrm{human}}$ as provided by Assumption~1. 
By Definition 2, for every triplet in the dataset $\mathcal{D}_h$, the reward's induced preference matches that of the human. More specifically, $\operatorname{sign}(\Delta G) = y_k$, which implies:
\[
y_k \cdot \Delta G_{r_{\theta^*}, \gamma}(\tau_k^{(1)}, \tau_k^{(2)}) > 0.
\]

Since $y_k \in \{-1, 1\}$ and $\tanh$ is an odd function, we have $y_k \tanh(z) = \tanh(y_k z)$, which lets us fold the label inside the $\tanh$:
\[
\ell_k = 1 - \lim_{\alpha \to \infty} y_k \tanh\!\left(\alpha \Delta G_{r_{\theta^*}, \gamma}\right)
       = 1 - \lim_{\alpha \to \infty} \tanh\!\left(\alpha \, y_k \, \Delta G_{r_{\theta^*}, \gamma}\right).
\]
Because $r_{\theta^*} \in \mathcal{R}_{\mathrm{human}}$ ensures $y_k \, \Delta G_{r_{\theta^*}, \gamma} > 0$, the argument satisfies $\alpha \, y_k \, \Delta G_{r_{\theta^*}, \gamma} \to +\infty$ as $\alpha \to \infty$. Therefore,
\[
\ell_k = 1 - 1 = 0 \quad \text{\small (since $\lim_{z \to +\infty} \tanh(z) = 1$)}.
\]
Next, we consider the empirical Soft-TAC loss over the dataset $\mathcal{D}_h$, given by:
\begin{align*}
\mathcal{L}_{\text{Soft-TAC}}(\theta^*; \mathcal{D}_{h}) 
&= \lim_{\alpha \to \infty} \mathbb{E}_{(\tau_k^{(1)}, \tau_k^{(2)}, y_k) \sim \mathcal{D}_h} \left[1 - y_k \cdot \tanh\left(\alpha \Delta G_{r_{\theta^*}, \gamma}(\tau_k^{(1)}, \tau_k^{(2)})\right) \right] \\
&= \lim_{\alpha \to \infty} \frac{1}{N} \sum_{k=1}^N \left[ 1 - y_k \cdot \tanh\left(\alpha\Delta G_{r_{\theta^*}, \gamma}(\tau_k^{(1)}, \tau_k^{(2)})\right) \right] \quad \text{\small (by expansion of the empirical expectation)} \\
&= \frac{1}{N} \sum_{k=1}^N \left[\lim_{\alpha \to \infty} \left( 1 - y_k \cdot \tanh\left(\alpha\Delta G_{r_{\theta^*}, \gamma}(\tau_k^{(1)}, \tau_k^{(2)})\right) \right) \right] \quad \text{\small (by linearity of limits for a finite sum)} \\
&= \frac{1}{N} \sum_{k=1}^N \ell_k \quad \text{\small (by substitution of the defined pointwise loss)} \\
&= \frac{1}{N} \sum_{k=1}^N (0) = 0.
\end{align*}
Because the loss function is globally bounded below by $0$, achieving an empirical loss value of $0$ establishes that $r_{\theta^*}$ is a global minimizer.
\medskip

\noindent $(\implies)$ We proceed via contradiction. Suppose that $r_{\theta^*}$ is a global minimizer of $\mathcal{L}_{\mathrm{Soft-TAC}}(\theta; \mathcal{D}_{h})$ (i.e., $\mathcal{L}_{\mathrm{Soft-TAC}}(\theta^*; \mathcal{D}_h) = 0$), but $r_{\theta^*} \notin \mathcal{R}_{\mathrm{human}}$. Following Definition 2, there must exist at least one index $k \in \{1,\dots,N\}$ tracking a triplet $(\tau_k^{(1)}, \tau_k^{(2)}, y_k) \in \mathcal{D}_h$ such that \[y_k \cdot \Delta G_{r_{\theta^*}, \gamma}(\tau_k^{(1)}, \tau_k^{(2)}) < 0.\]

Examining the individual loss $\ell_k$ for this misaligned pair, we again use that $y_k \in \{-1,1\}$ and $\tanh$ is odd to fold the label inside:
\[
\ell_k = 1 - \lim_{\alpha \to \infty} \tanh\!\left(\alpha \, y_k \, \Delta G_{r_{\theta^*}, \gamma}\right).
\]
Since $y_k \, \Delta G_{r_{\theta^*}, \gamma} < 0$, the argument satisfies $\alpha \, y_k \, \Delta G_{r_{\theta^*}, \gamma} \to -\infty$ as $\alpha \to \infty$, giving
\[
\ell_k = 1 - (-1) = 2 \quad \text{\small (since $\lim_{z \to -\infty} \tanh(z) = -1$)}.
\]

As $\ell_k > 0$, and each remaining pointwise loss $\ell_m$ ($m \neq k$) is either $0$ or $2$ in the limit and hence satisfies $\ell_m \geq 0$, separating our misaligned index ($k$) from the empirical loss summation gives:
\begin{align*}
\mathcal{L}_{\mathrm{Soft-TAC}}(\theta^*; \mathcal{D}_h) 
&= \frac{1}{N} \left( \ell_k + \sum_{m \neq k} \ell_m \right) \quad \text{\small (by isolating the single misaligned index $k$)} \\
&\geq \frac{1}{N} (\ell_k + 0) \quad \text{\small (since $\ell_m \ge 0$ for all $m \ne k$)} \\
&> 0 \quad \text{\small (since $\ell_k > 0$ and $N > 0$)}.
\end{align*}
This strictly contradicts our original premise that $r_{\theta^*}$ is a global minimizer. Thus, $r_{\theta^*}$ must belong to the set $\mathcal{R}_{\mathrm{human}}$.
\end{proof}

\noisetolerance*

\begin{proof}
Ghosh et al. (2017) establish that a loss function is noise-tolerant under simple non-uniform label noise satisfying $\eta_{\mathbf{x}} < \frac{n-1}{n}$ across the entire domain $\mathcal{X}$ if it fulfills two conditions:
\begin{enumerate}
  \item \textbf{Symmetry Condition}: $\sum_{y \in \mathcal{Y}} L(f(\mathbf{x}), y) = C, \forall \mathbf{x} \in \mathcal{X}$, for some constant $C$. 
  \item \textbf{Zero Clean Risk}: The global minimizer of the expected risk achieves value of 0, i.e., $R_L(f^*) = 0$.
\end{enumerate}
We verify that Soft-TAC satisfies both conditions.

\paragraph{1. Verification of the Symmetry Condition:}
Let $\Delta G_{r_{\theta}, \gamma}(\mathbf{x})$ represent the predicted return difference output by our reward model for an arbitrary trajectory comparison pair $\mathbf{x}$. Summing the Soft-TAC loss definition across all potential labels in $\mathcal{Y} = \{-1, 1\}$ yields:
\begin{align*} 
\sum_{y \in \{-1, 1\}} L\left(\Delta G_{r_{\theta}, \gamma}(\mathbf{x}), y\right) 
&= \left[1 - (1)\tanh\left(\alpha \Delta G_{r_{\theta}, \gamma}(\mathbf{x})\right)\right] + \left[1 - (-1)\tanh\left(\alpha \Delta G_{r_{\theta}, \gamma}(\mathbf{x})\right)\right] \\ 
&= \left[1 - \tanh\left(\alpha \Delta G_{r_{\theta}, \gamma}(\mathbf{x})\right)\right] + \left[1 + \tanh\left(\alpha \Delta G_{r_{\theta}, \gamma}(\mathbf{x})\right)\right] \\
&= 1 + 1 = 2.
\end{align*}
Because the sum resolves directly to a constant value ($C=2$) completely independent of the predicted return difference, the symmetry requirement holds true across the domain $\mathcal{X}$.

\paragraph{2. Verification of Zero Clean Risk:}
Consider a parameter vector $\theta^*$ satisfying $r_{\theta^*} \in \mathcal{R}_{\mathrm{human}}^{\mathrm{pop}}$. By definition of $\mathcal{R}_{\mathrm{human}}^{\mathrm{pop}}$, any trajectory pair and preference label sampled from the clean population distribution $\mathcal{D}_h^{\mathrm{pop}}$ will always yield a predicted return difference with the correct sign, meaning $y \cdot \Delta G_{r_{\theta^*}, \gamma}(\mathbf{x}) > 0$ almost surely. 

To evaluate whether Soft-TAC satisfies the zero clean risk condition, we examine the limit of the expected risk as the temperature scaling parameter $\alpha$ approaches infinity. Because the $\tanh$ function is naturally bounded between $-1$ and $1$, the entire loss term inside the expectation is uniformly bounded. This allows us to apply the Dominated Convergence Theorem (DCT) to pull the limit inside the expectation operator:
\begin{align*}
\lim_{\alpha \to \infty} R_L(r_{\theta^*})
&= \lim_{\alpha \to \infty} \mathbb{E}_{(\mathbf{x}, y) \sim \mathcal{D}_h^{\mathrm{pop}}} \Big[ 1 - y \cdot \tanh\left(\alpha \Delta G_{r_{\theta^*}, \gamma}(\mathbf{x})\right) \Big] \\
&= \mathbb{E}_{(\mathbf{x}, y) \sim \mathcal{D}_h^{\mathrm{pop}}} \Big[ \lim_{\alpha \to \infty} \left( 1 - y \cdot \tanh\left(\alpha \Delta G_{r_{\theta^*}, \gamma}(\mathbf{x})\right) \right) \Big] \quad \text{\small (by the DCT)} \\
&= \mathbb{E}_{(\mathbf{x}, y) \sim \mathcal{D}_h^{\mathrm{pop}}} [ 1 - 1 ] \quad \text{\small (since $\lim_{z \to \infty} \tanh(z) = 1$)} \\
&= \mathbb{E}_{(\mathbf{x}, y) \sim \mathcal{D}_h^{\mathrm{pop}}} [ 0 ] = 0.
\end{align*}
Because the expected risk under clean labels reaches exactly zero (i.e., $R_L(r_{\theta^*}) = 0$), the second condition from Ghosh et al. (2017) is satisfied, proving that the Soft-TAC loss function is noise-tolerant as $\alpha \to \infty$.
\end{proof}

\section*{Empirical Appendix}
\section{User Study Details}
% In this section, we first describe the user interface of the study, then we provide additional results. 
\subsection{Environment: Lunar Lander}\label{sec:env_details_lunar_lander}
In the Lunar Lander environment, the agent begins near the top-center of the screen and must land on a fixed landing pad as quickly and smoothly as possible, avoiding excessive hovering or unstable movements. The agent selects actions from a 2-dimensional continuous space controlling the main engine thrust and the side engines for lateral movement. The state space is an 8-dimensional vector containing position, velocity, angle, angular velocity, and two leg-contact indicators. An episode ends when the lander crashes, goes off screen, or comes to rest—either due to a successful landing or because it has stopped moving.

\subsubsection{Reward Function Components}

The agent receives a reward defined as a linear combination of eight features. Below we describe each component used in the Lunar Lander reward function.

\begin{enumerate}
    \item \textbf{Progress-based rewards}
    \begin{itemize}
        \item \textbf{Progress toward the landing pad.}
        Measures the change in Euclidean distance between the lander and the landing pad across consecutive time steps. A negative weight rewards the agent for reducing this distance (moving closer).
        
        \item \textbf{Change in lander speed.}
        Measures the change in the lander’s velocity magnitude. A positive weight rewards acceleration, while a negative weight rewards deceleration.
        
        \item \textbf{Change in lander angle (uprightness).}
        Measures changes in the absolute tilt angle. A negative weight rewards the lander for becoming more upright.
        
        \item \textbf{Change in leg contact.}
        Measures differences in the number of legs in contact with the ground (0--2) between steps. A positive weight rewards gaining contact, while a negative weight rewards losing contact.
    \end{itemize}

    \item \textbf{Absolute rewards}
    \begin{itemize}
        \item \textbf{Side engine use.}
        A binary reward indicating whether the side engine is activated in the current step. A positive weight encourages more usage; a negative weight discourages it.
        
        \item \textbf{Main engine use.}
        A binary reward indicating whether the main engine is being used. Positive weights encourage usage; negative weights discourage it.
        
        \item \textbf{Crashing.}
        A binary reward indicating whether the lander crashes. Negative weights penalize crashes, whereas positive weights encourage them.
        
        \item \textbf{Safe landing.}
        A binary reward indicating when the lander completes a safe landing. Positive weights encourage successful landings; negative weights discourage them.
    \end{itemize}
\end{enumerate}
\subsection{User Study Interface}\label{sec:user_study_images}
\begin{figure*}[h!]
    \centering
    \begin{subfigure}[b]{0.48\textwidth}
        \centering
        \includegraphics[width=\textwidth]{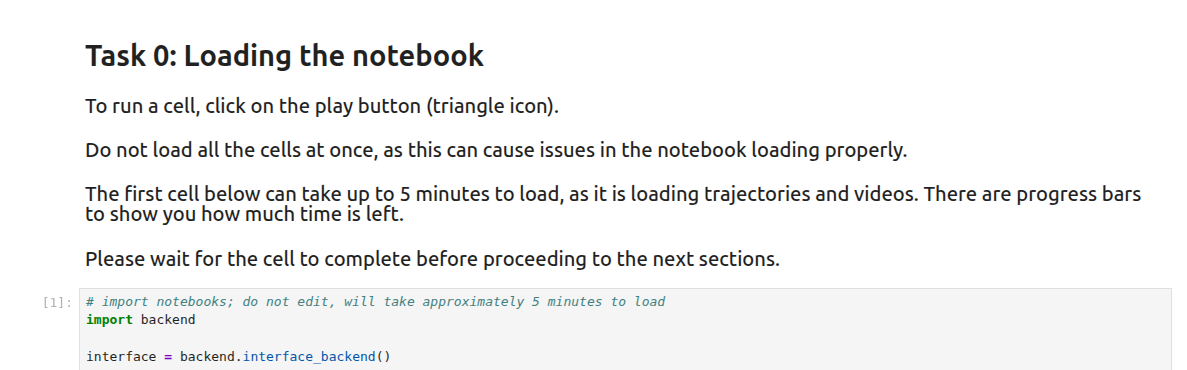}
        \caption{Task 0: Loading the notebook}
        \label{fig:user_study_task0}
    \end{subfigure}
    \hfill
    \begin{subfigure}[b]{0.48\textwidth}
        \centering
        \includegraphics[width=\textwidth, clip=true, trim=0pt 10cm 0cm 0cm]{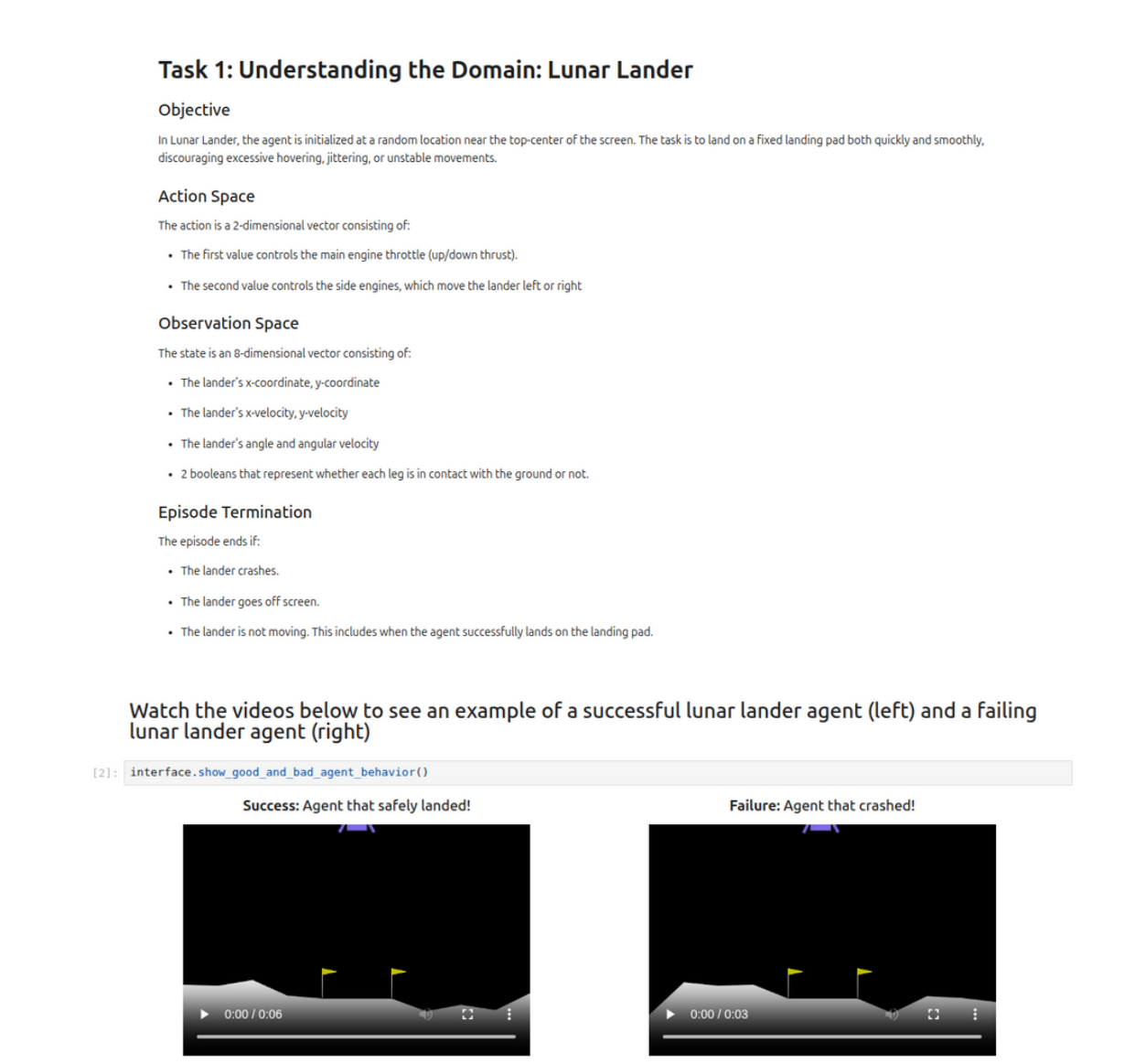}
        \caption{Task 1: Reviewing the domain}
        \label{fig:user_study_task1}
    \end{subfigure}
    
    \caption{User study interface walkthrough showing (a) Task 0 and (b) Task 1.}
    \label{fig:user_study_tasks_0_and_1}
\end{figure*}

\begin{figure*}[t]
    \centering
    \begin{subfigure}[b]{0.48\textwidth}
        \centering
        \includegraphics[width=\textwidth]{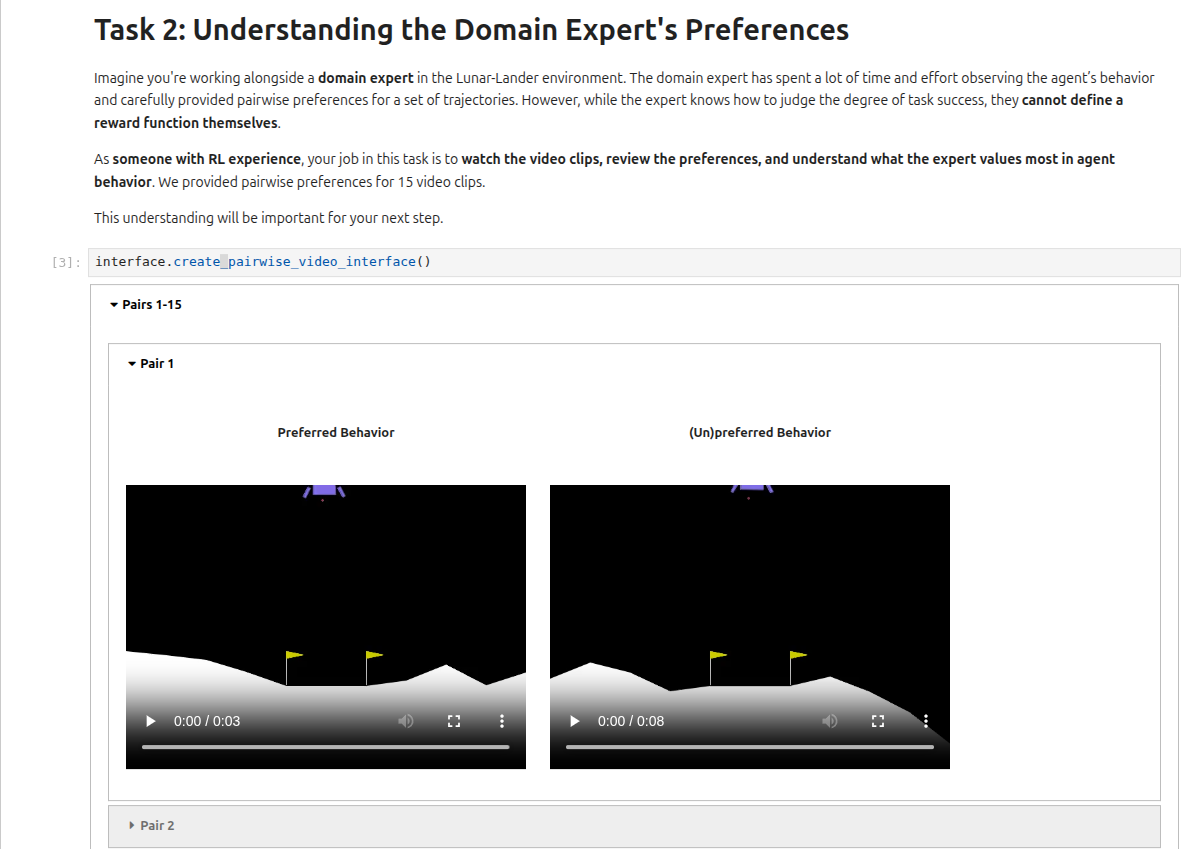}
        \caption{Task 2: Reviewing the domain expert's preferences}
        \label{fig:user_study_task2}
    \end{subfigure}
    \hfill
    \begin{subfigure}[b]{0.48\textwidth}
        \centering
        \includegraphics[width=\textwidth]{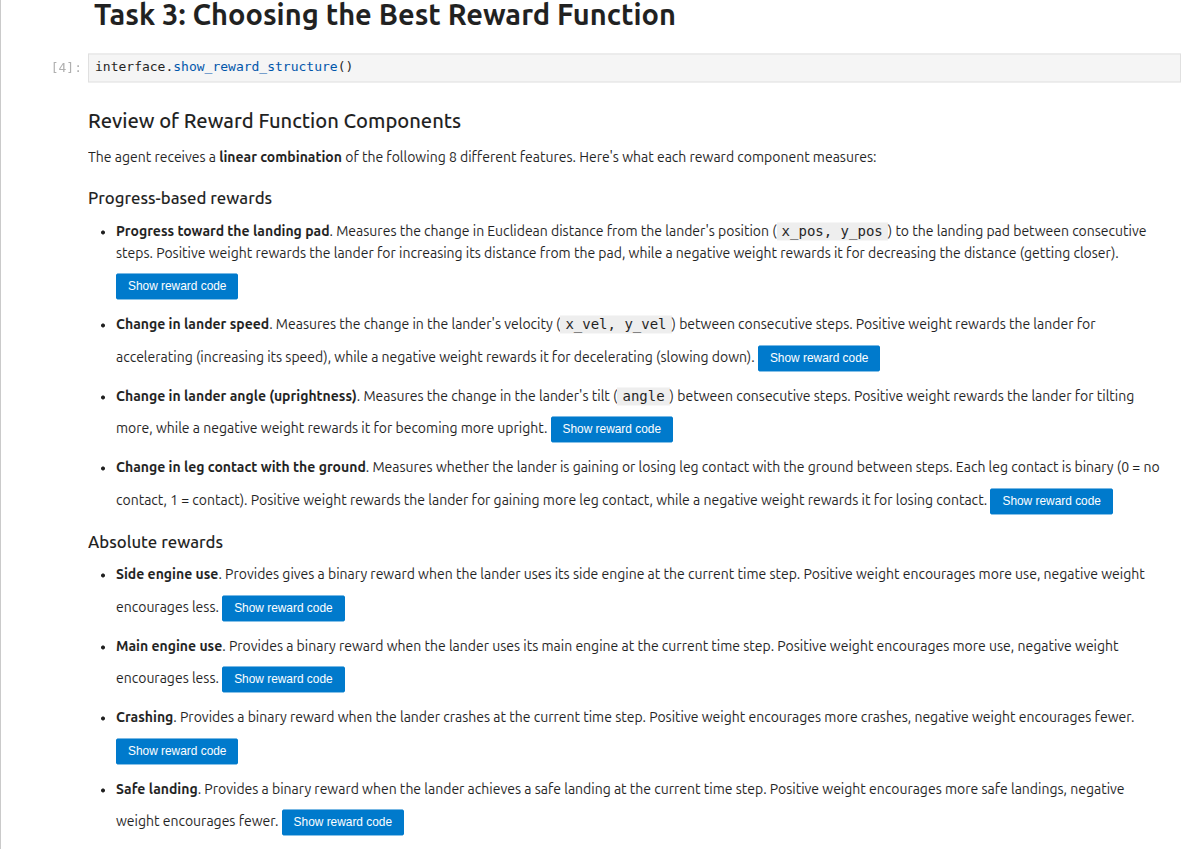}
        \caption{Task 3, Part A: Reviewing the Lunar Lander reward parts}
        \label{fig:user_study_task3_a}
    \end{subfigure}

    \begin{subfigure}[b]{0.48\textwidth}
        \centering
        \includegraphics[width=\textwidth]{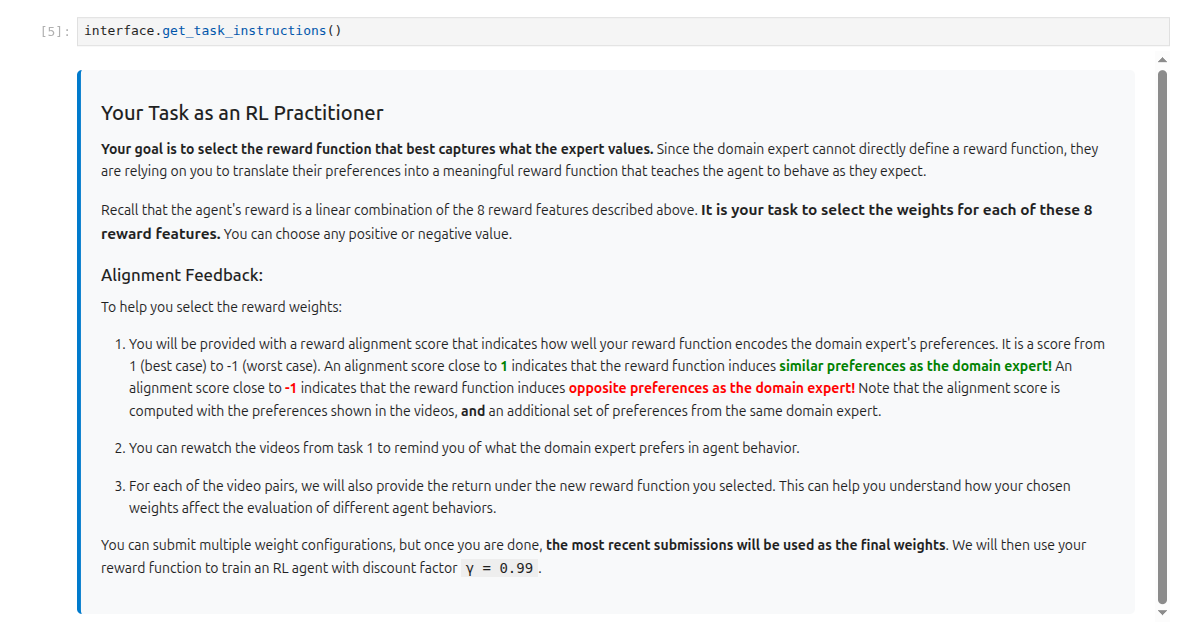}
        \caption{Task 3, Part B: Reviewing the reward tuning instructions}
        \label{fig:user_study_task3_b}
    \end{subfigure}
    \hfill
    \begin{subfigure}[b]{0.48\textwidth}
        \centering
        \includegraphics[width=\textwidth]{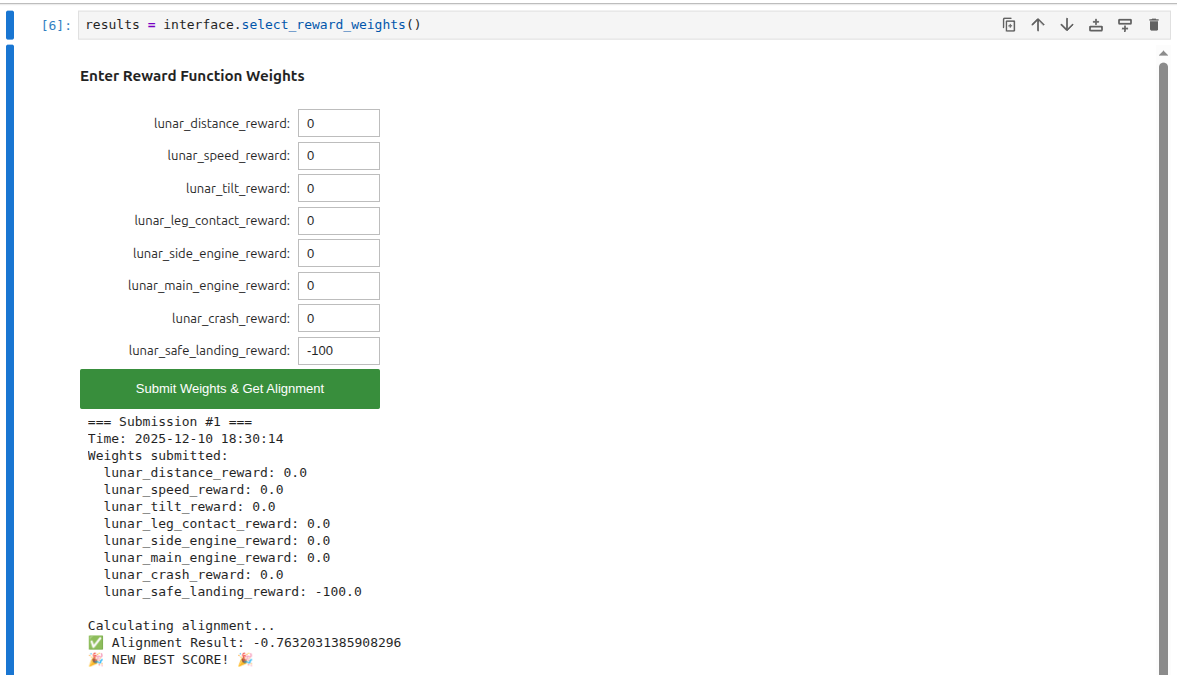}
        \caption{Task 3, Part C: Reward tuning UI}
        \label{fig:user_study_task3_c}
    \end{subfigure}
    
    \caption{User study interface walkthrough across Task 2 and Task 3.}
    \label{fig:user_study_tasks}
\end{figure*}

In Figures \ref{fig:user_study_task0}--\ref{fig:user_study_task3_c}, we present images of the Jupyter notebook interface used in the human-subject study.

% \begin{table*}[]
% \centering
% \caption{Participant-tuned Lunar Lander reward weights for each condition in the user study.}
% \label{tab:reward_parts_experience}
% \begin{tabular}{l c c c c c c c c}
% \toprule
% \textbf{Condition} & \textbf{Distance} & \textbf{Speed} & \textbf{Tilt} & \textbf{Leg Contact} & \textbf{Side Engine} & \textbf{Main Engine} & \textbf{Crash} & \textbf{Safe Landing} \\
% \midrule
% \multirow{5}{*}{\textsc{Control}}
%  & -1.0 & 0.0 & -1.0 & 1.0 & -0.1 & -0.1 & -10.0 & 10.0 \\
%  & -0.4 & -0.5 & -0.4 & 0.0 & -0.25 & -0.1 & -2.0 & 100.0 \\
%  & -55.5 & 0.0 & 0.0 & 10.0 & 5.0 & 1.0 & 20.0 & 100.0 \\
%  & 1.0 & 1.0 & 0.0 & 1.0 & 0.05 & 0.05 & -10.0 & 10.0 \\
%  & -2000.0 & -10.0 & -10.0 & 0.0 & -1.0 & -1.0 & -1000.0 & 1000.0 \\
% \midrule
% \multirow{6}{*}{\textsc{Alignment}}
%  & -2.0 & -1.0 & -0.1 & 1.0 & -0.01 & -0.05 & -5.0 & 1000.0 \\
%  & -1.0 & 0.1 & -0.1 & 1.0 & 0.0 & 0.0 & -10.0 & 10.0 \\
%  & 3.0 & 1.0 & -2.0 & 4.0 & 0.0 & 0.0 & -5.0 & 5.0 \\
%  & 0.0 & 0.0 & 0.0 & 0.0 & 0.0 & 0.0 & 0.0 & 1000.0 \\
%  & -10.0 & -2.5 & -1.5 & 1.5 & -0.1 & -0.05 & -25.0 & 78.0 \\
%  & -100.0 & -5.0 & -100.0 & 1.0 & 0.0 & 0.0 & -50.0 & 10000.0 \\
% \bottomrule
% \end{tabular}
% \end{table*}
\newpage

\begin{table*}[t]
\centering
\caption{Participant-tuned reward weights and resulting evaluation metrics (TAC, Average Landing-pad Success Rate, and number of reward tuning iterations).}
\label{tab:reward_parts_experience}
\small 
\begin{tabular}{l r r r r r r}
\toprule
\textbf{Metric} & \textbf{P1} & \textbf{P2} & \textbf{P3} & \textbf{P4} & \textbf{P5} & \textbf{P6} \\
\midrule
\multicolumn{7}{c}{\textsc{Control Condition}} \\
\midrule
\textit{Reward Weights} & & & & & & \\
Distance     & -1.0  & -0.4  & -55.5 & 1.0   & -2000.0 & -- \\
Speed        & 0.0   & -0.5  & 0.0   & 1.0   & -10.0   & -- \\
Tilt         & -1.0  & -0.4  & 0.0   & 0.0   & -10.0   & -- \\
Leg Contact  & 1.0   & 0.0   & 10.0  & 1.0   & 0.0     & -- \\
Side Engine  & -0.1  & -0.25 & 5.0   & 0.05  & -1.0    & -- \\
Main Engine  & -0.1  & -0.1  & 1.0   & 0.05  & -1.0    & -- \\
Crash        & -10.0 & -2.0  & 20.0  & -10.0 & -1000.0 & -- \\
Safe Landing & 10.0  & 100.0 & 100.0 & 10.0  & 1000.0  & -- \\
\addlinespace[0.5em]
\textit{Evaluation Metrics} & & & & & & \\
TAC          & 0.68 & 0.61 & 0.24 & 0.41 & 0.64   & -- \\
Landing-pad Success Rate  & 0.0 & 0.0 & 0.0 & 0.0 & 0.992   & -- \\
Tuning Iterations & 8 & 21 & 79 & 54 & 86   & -- \\
\midrule
\multicolumn{7}{c}{\textsc{Alignment Condition}} \\
\midrule
\textit{Reward Weights} & & & & & & \\
Distance     & -2.0  & -1.0  & 3.0   & 0.0   & -10.0   & -100.0 \\
Speed        & -1.0  & 0.1   & 1.0   & 0.0   & -2.5    & -5.0   \\
Tilt         & -0.1  & -0.1  & -2.0  & 0.0   & -1.5    & -100.0 \\
Leg Contact  & 1.0   & 1.0   & 4.0   & 0.0   & 1.5     & 1.0    \\
Side Engine  & -0.01 & 0.0   & 0.0   & 0.0   & -0.1    & 0.0    \\
Main Engine  & -0.05 & 0.0   & 0.0   & 0.0   & -0.05   & 0.0    \\
Crash        & -5.0  & -10.0 & -5.0  & 0.0   & -25.0   & -50.0  \\
Safe Landing & 1000.0& 10.0  & 5.0   & 1000.0& 78.0    & 10000.0\\
\addlinespace[0.5em]
\textit{Evaluation Metrics} & & & & & & \\
TAC          & 0.75 & 0.62 & 0.42 & 0.73 & 0.76  & 0.81  \\
Landing-pad Success Rate  & 0.6 & 1.0 & 0.0 & 0.168 & 1.0   & 0.992  \\
Tuning Iterations & 19 & 5 & 16 & 55 & 80  & 89  \\
\bottomrule
\end{tabular}
\end{table*}
\subsection{Domain Expert Preferences}
In the Lunar Lander user study, a domain expert (one of the authors) manually reviewed 148 pairs of video clips to provide preferences between them. 
The video clips compared a variety of behaviors. For example, some featured two successful trials (e.g., both landed on the pad), where experts preferred more efficient behaviors, such as faster landings. Conversely, some pairs consisted of two failures; in these cases, preference was given to ``better'' failures, such as crashing quickly to conserve fuel rather than hovering before an inevitable crash. 

To select preferences over the video clips, 
the domain expert was provided exclusively with video clips, with no access to numeric metadata such as rewards or metrics of any kind. Instructions were limited to information regarding the general state/action spaces of the environments as well as the high-level objectives: landing accuracy.

\subsection{Additional Results}\label{sec:ll_additional_results}
In addition to the NASA--TLX survey, we also asked participants to rate their reward design experience on a 1--7 scale. Participants in the Alignment group reported slightly higher experience (mean = 5.83, SE = 0.31) than those in the Control group (mean = 5.0, SE = 0.89). However, this difference was not statistically significant according to a Welch's \(t\)-test (\(t = 0.88\), \(p = 0.79\)). Thus, reward design experience is unlikely to be a confounding factor.

Next, Table \ref{tab:reward_parts_experience} summarizes the tuned reward functions provided by all participants. We find that 10 out of 11 participants opted for dense reward structures rather than sparse ones. The resulting reward weights also varied widely, especially for the distance, tilt, crash, and safe-landing components.

Table \ref{tab:user_study_stat_testing} details the statistical testing used to compare the success rates between the alignment and control conditions.

\begin{table*}[h]
\centering
\caption{User Study Descriptive statistics (Mean $\pm$ SE), Shapiro-Wilk normality tests ($p_{\text{SW}}$, where $p_{\text{SW}} > 0.05$ indicates normality), and Mann--Whitney $U$ test comparison between Alignment and Control groups.}
\label{tab:user_study_stat_testing}
\small
\begin{tabular}{l c c c c c c}
\toprule
 & \multicolumn{2}{c}{\textbf{Alignment}} & \multicolumn{2}{c}{\textbf{Control}} & \multicolumn{2}{c}{\textbf{Alignment vs.\ Control}} \\
\cmidrule(lr){2-3} \cmidrule(lr){4-5} \cmidrule(lr){6-7}
\textbf{Metric} & \textbf{Mean $\pm$ SE} & \textbf{$p_{\text{SW}}$} & \textbf{Mean $\pm$ SE} & \textbf{$p_{\text{SW}}$} & \textbf{$U$} & \textbf{$p$} \\
\midrule
Landing-pad success rate & $0.6267 \pm 0.1840$ & 0.0832 & $0.1984 \pm 0.1984$ & 0.0001 & $U = 24.5$ & 0.0416 \\
\bottomrule
\end{tabular}
\end{table*}

% --- Table 20: User Study Survey Results ---
\begin{table*}[h]
\centering
\caption{User Study Results for the NASA-TLX Survey and Years of Experience in Reward Design: Descriptive Statistics (Mean $\pm$ SE), Shapiro-Wilk Normality Tests ($p_{\text{SW}}$, where $p_{\text{SW}} > 0.05$ indicates normality), and Statistical Comparison between Alignment and Control groups. Parametric Student's $t$-test ($t$) or non-parametric Mann--Whitney $U$ test ($U$) statistics are reported based on normality.}
\label{tab:user_study_nasa_tlx_stat_testing}
\small
\begin{tabular}{l c c c c c c}
\toprule
 & \multicolumn{2}{c}{\textbf{Alignment}} & \multicolumn{2}{c}{\textbf{Control}} & \multicolumn{2}{c}{\textbf{Alignment vs.\ Control}} \\
\cmidrule(lr){2-3} \cmidrule(lr){4-5} \cmidrule(lr){6-7}
\textbf{Metric} & \textbf{Mean $\pm$ SE} & \textbf{$p_{\text{SW}}$} & \textbf{Mean $\pm$ SE} & \textbf{$p_{\text{SW}}$} & \textbf{Statistic} & \textbf{$p$} \\
\midrule
Experience       & $5.8333 \pm 0.3073$ & 0.2117 & $5.0000 \pm 0.8944$ & 0.4399 & $t = 0.8810$ & 0.7905 \\
Average Workload & $2.8056 \pm 0.2373$ & 0.8304 & $4.0667 \pm 0.4137$ & 0.6557 & $t = -2.6440$ & 0.0178 \\
\bottomrule
\end{tabular}
\end{table*}
\clearpage

\section{Soft TAC Details}
\subsection{Difference Between \citet{zheng2023diffkendall} and Soft-TAC}

While both methods share a $\tanh$ relaxation of Kendall's Tau for differentiability, they diverge in formulation and optimization role. \citet{zheng2023diffkendall} employs the correlation as a similarity measure for image classification, comparing two continuous feature vectors:$$\tilde{\tau}(\mathbf{x}, \mathbf{y}) = \mathbb{E}_{i > j} \big[ \tanh(\alpha(x_i - x_j)) \tanh(\alpha(y_i - y_j)) \big]$$Here, $\tilde{\tau}$ parameterizes a Softmax to produce class probabilities $P(y|x) \propto \exp(\tilde{\tau})$ for a standard cross-entropy loss. Conversely, Soft-TAC derives a direct reward learning objective by comparing continuous reward differences ($\Delta G_r$) against discrete preference labels ($y$):$$\mathcal{L} = 1 - \mathbb{E}_{(\tau^i, \tau^j, y) \sim \mathcal{D}_h} \big[ y \cdot \tanh(\alpha \Delta G_r) \big]$$Notably, while \citet{zheng2023diffkendall} uses the correlation as a feature-matching proxy for classification, Soft-TAC adopts it as the primary objective for aligning rewards with human feedback.

\subsection{Understanding the Gradient Behavior of Soft-TAC versus Cross-Entropy}\label{sec:grad_behavior_st}
In the main text, we discuss that the gradient of Soft-TAC goes to zero for data points where the prediction is confidently incorrect, whereas the gradient of Cross-Entropy does not. However, we note that this can be a positive attribute in cases where the dataset contains noisy or incorrect labels. We show that this phenomenon occurs in a toy supervised learning example. 

In our toy preference-learning example, we learn a linear reward function from pairwise comparisons. We consider five items with feature values $\phi_0 = 0, \phi_1 = 1, \phi_2 = 2, \phi_3 = 3$, and $\phi_4 = 4$, and aim to learn a single scalar weight $w$ defining a linear reward function $r(\phi_i) = w \phi_i$, such that the induced reward ordering is consistent with the observed preferences.

The training dataset consists of five pairwise comparisons: $(0,1,-1)$, $(1,2,-1)$, $(2,3,-1)$, $(3,4,-1)$, and $(2,4,+1)$. Each triplet $(\phi_i, \phi_j, y)$ denotes a comparison, where $y = -1$ indicates that the item with feature $\phi_i$ is preferred over the item with feature $\phi_j$, and $y = +1$ indicates the opposite. In all but one comparison, the item with the higher feature value is preferred. The comparison $(2,4,+1)$ therefore represents an incorrectly labeled data point, modeling the presence of noisy preference labels in the dataset.

We perform supervised learning using stochastic gradient descent with the Adam optimizer, a learning rate of 0.1, and 40 training epochs. We use SGD to explicitly examine how correctly and incorrectly labeled data points influence the gradient and, consequently, the weight updates. We compare two loss functions: Soft-TAC and the standard Cross-Entropy.

\begin{figure*}[h!]
    \centering
    \begin{subfigure}[b]{0.7\textwidth}
        \centering
        \includegraphics[width=\textwidth]{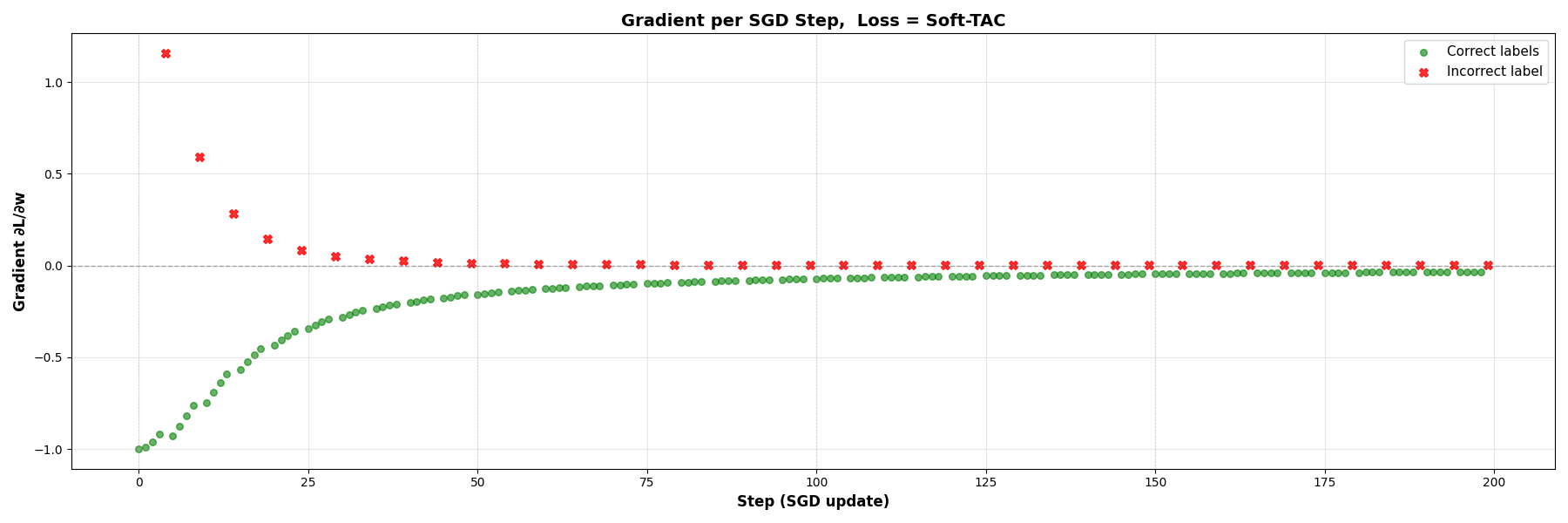}
        \caption{Gradient behavior per SGD step.}
        \label{fig:gradient_updates_soft_tac}
    \end{subfigure}

    \begin{subfigure}[b]{0.7\textwidth}
        \centering
        \includegraphics[width=\textwidth]{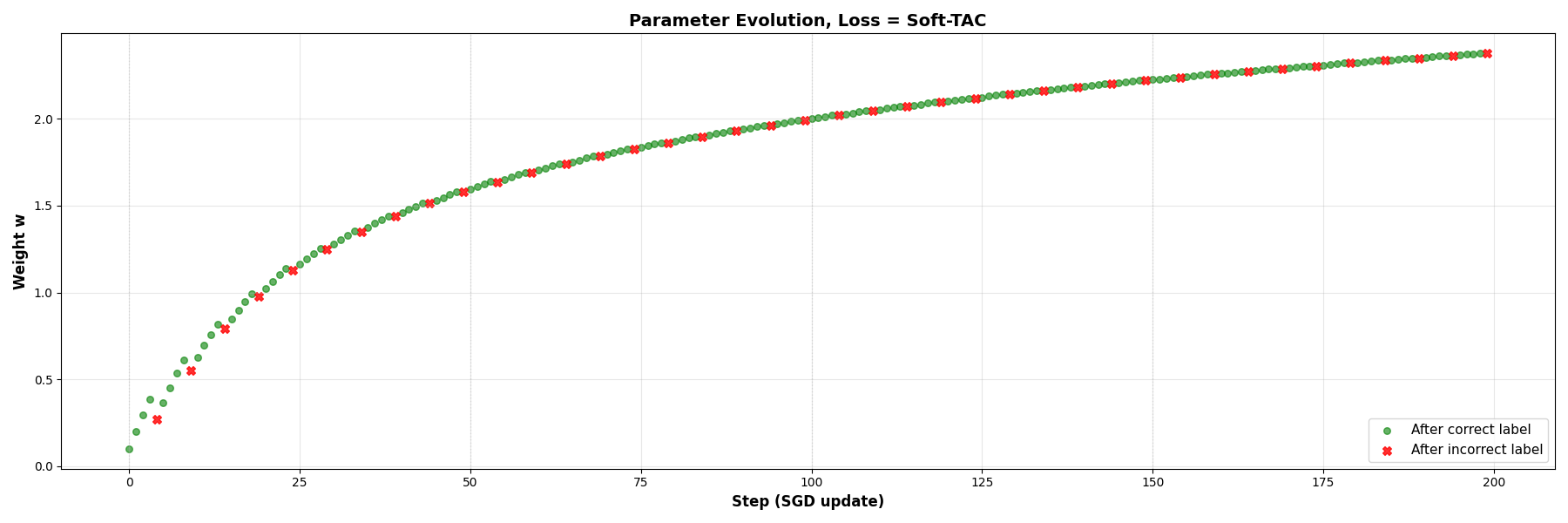}
        \caption{Parameter evolution showing impact of each label type.}
        \label{fig:weight_updates_soft_tac}
    \end{subfigure}
    \caption{Soft-TAC loss behavior in a simple toy preference learning example with \textbf{noisy} data.}
    \label{fig:toy_example_st}
\end{figure*}

As shown in Figure~\ref{fig:gradient_updates_soft_tac}, the gradient of the Soft-TAC objective for the incorrectly labeled data point decays toward zero, with a final magnitude of approximately $6\times10^{-4}$. In contrast, the gradients for the correctly labeled data points converge to values around $0.035$. This behavior results in a monotonic increase in the learned weight, which reaches a final value of approximately $2.3$, as shown in Figure~\ref{fig:weight_updates_soft_tac}.

\begin{figure*}[h!]
    \centering
    \begin{subfigure}[b]{0.7\textwidth}
        \centering
        \includegraphics[width=\textwidth]{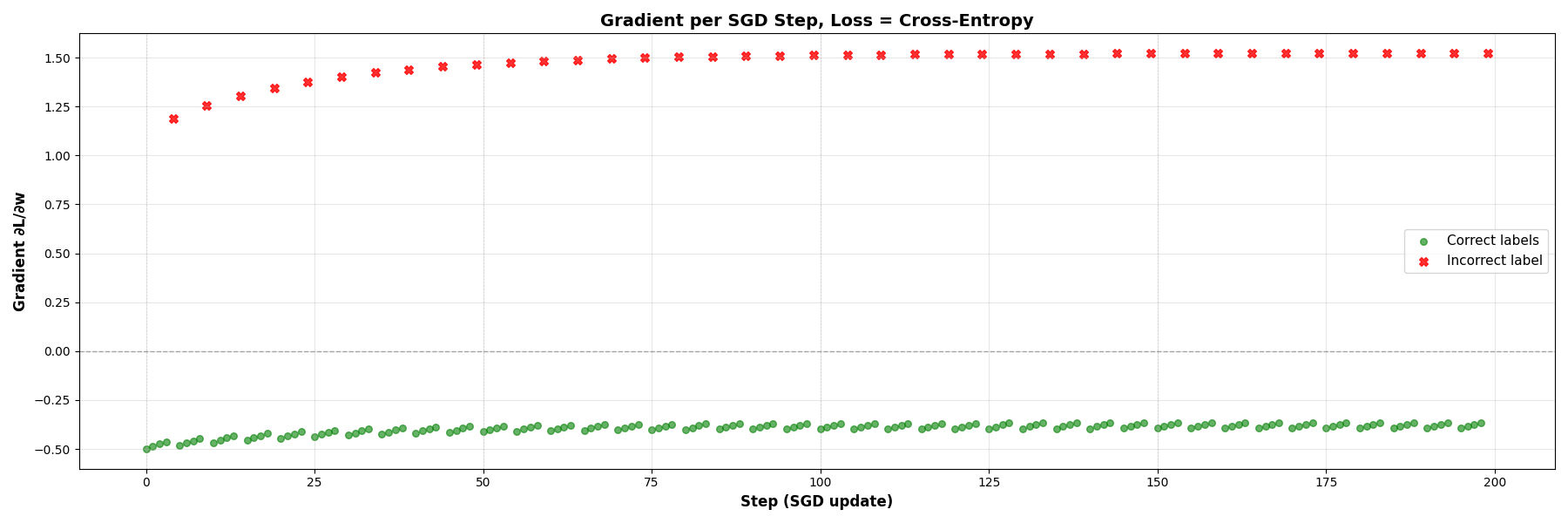}
        \caption{Gradient behavior per SGD step.}
        \label{fig:gradient_updates_cross_entropy}
    \end{subfigure}
        
    \begin{subfigure}[b]{0.7 \textwidth}
        \centering
        \includegraphics[width=\textwidth]{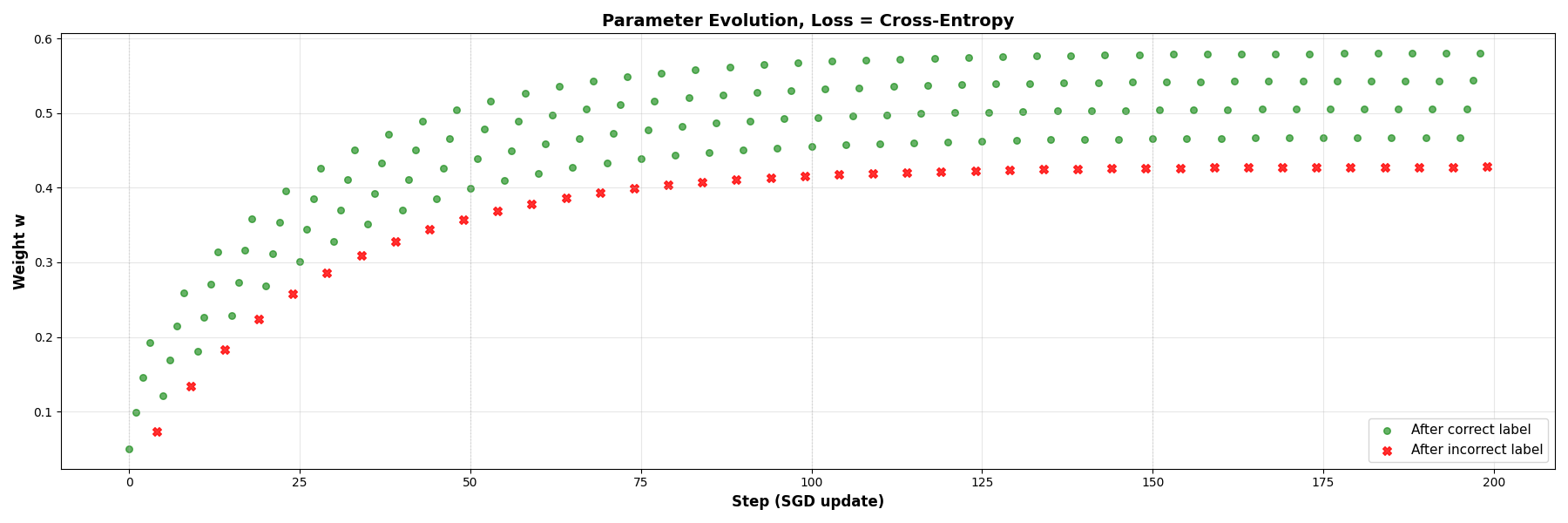}
        \caption{Parameter evolution showing impact of each label type.}
        \label{fig:weight_updates_cross_entropy}
    \end{subfigure}
    \caption{Cross-entropy loss behavior in a simple toy preference learning example with \textbf{noisy} data.}
    \label{fig:toy_example_ce}
\end{figure*}

By comparison, Figure~\ref{fig:gradient_updates_cross_entropy} shows that under the Cross-Entropy loss, the gradient for the incorrectly labeled data point remains large, persisting at approximately $1.5$ throughout training, while gradients for the correctly labeled data points remain in the range of $-0.3$ to $-0.4$. These conflicting gradients lead to unstable weight updates: although updates from correctly labeled data increase the weight, a single update from the incorrectly labeled data point substantially decreases it, undoing much of the accumulated progress. As a result, the weight increases only marginally over training, as shown in Figure~\ref{fig:weight_updates_cross_entropy}.

\begin{figure*}[h!]
    \centering
    \begin{subfigure}[b]{0.48\textwidth}
        \centering
        \includegraphics[width=\textwidth]{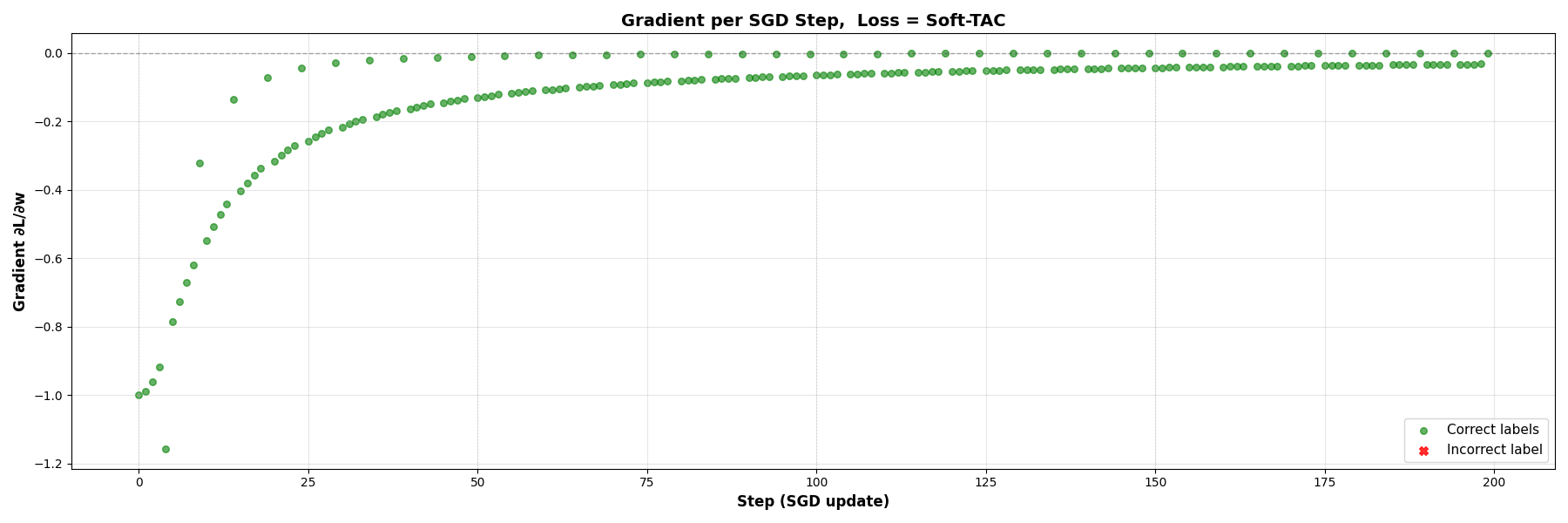}
        \caption{Gradient behavior per SGD step.}
        \label{fig:toy_example_gradients_st_all_correct}
    \end{subfigure}
    \hfill
    \begin{subfigure}[b]{0.48\textwidth}
        \centering
        \includegraphics[width=\textwidth]{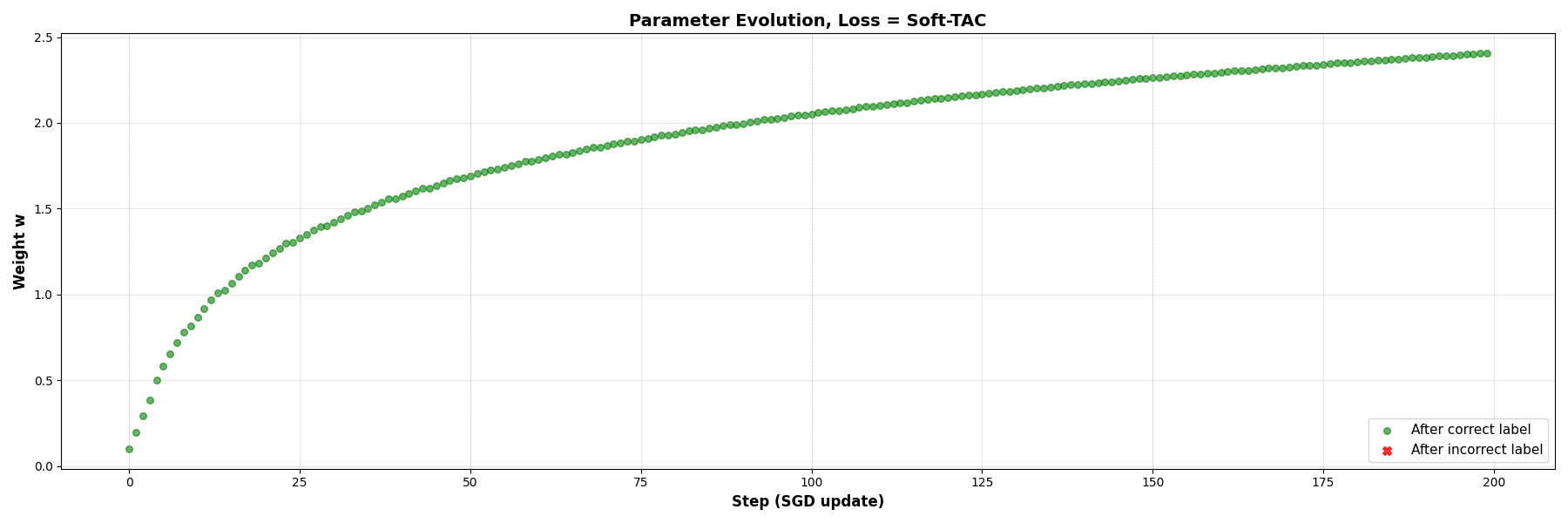}
        \caption{Parameter evolution showing impact of each label type.}
        \label{fig:toy_example_weights_st_all_correct}
    \end{subfigure}
    \caption{Soft-TAC loss behavior in a simple toy preference learning example with \textbf{noise-free} data.}
    \label{fig:toy_example_st_all_correct}
\end{figure*}

\begin{figure*}[h!]
    \centering
    \begin{subfigure}[b]{0.48\textwidth}
        \centering
        \includegraphics[width=\textwidth]{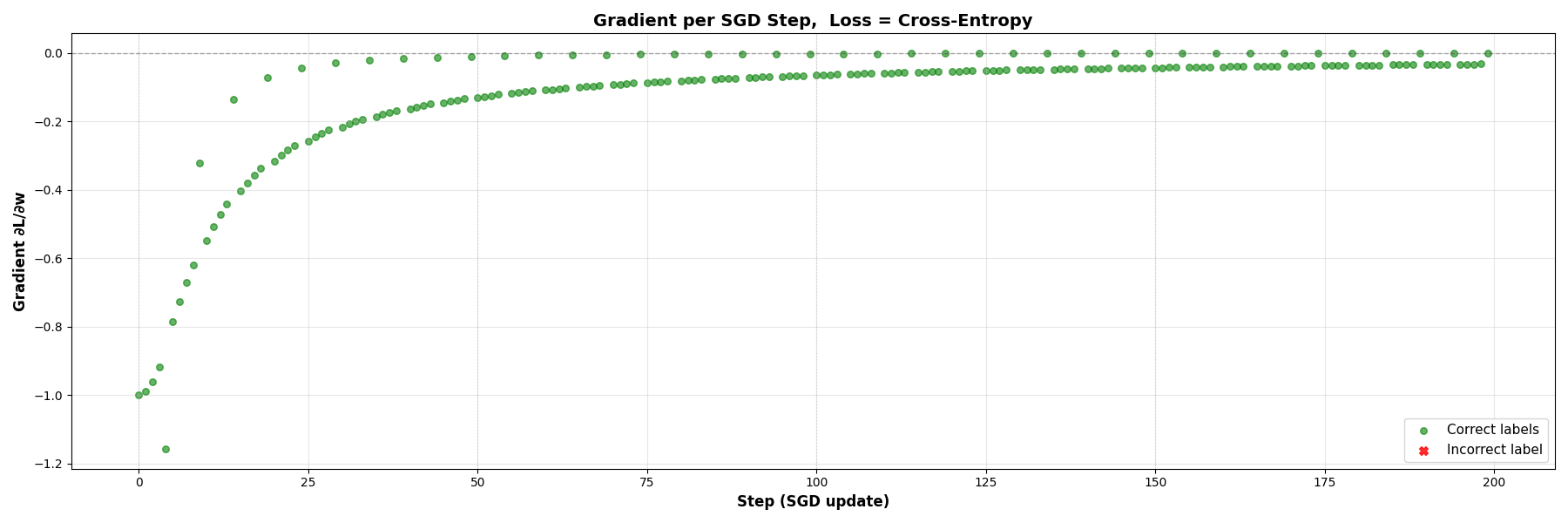}
        \caption{Gradient behavior per SGD step.}
        \label{fig:toy_example_gradients_ce_all_correct}
    \end{subfigure}
    \hfill
    \begin{subfigure}[b]{0.48\textwidth}
        \centering
        \includegraphics[width=\textwidth]{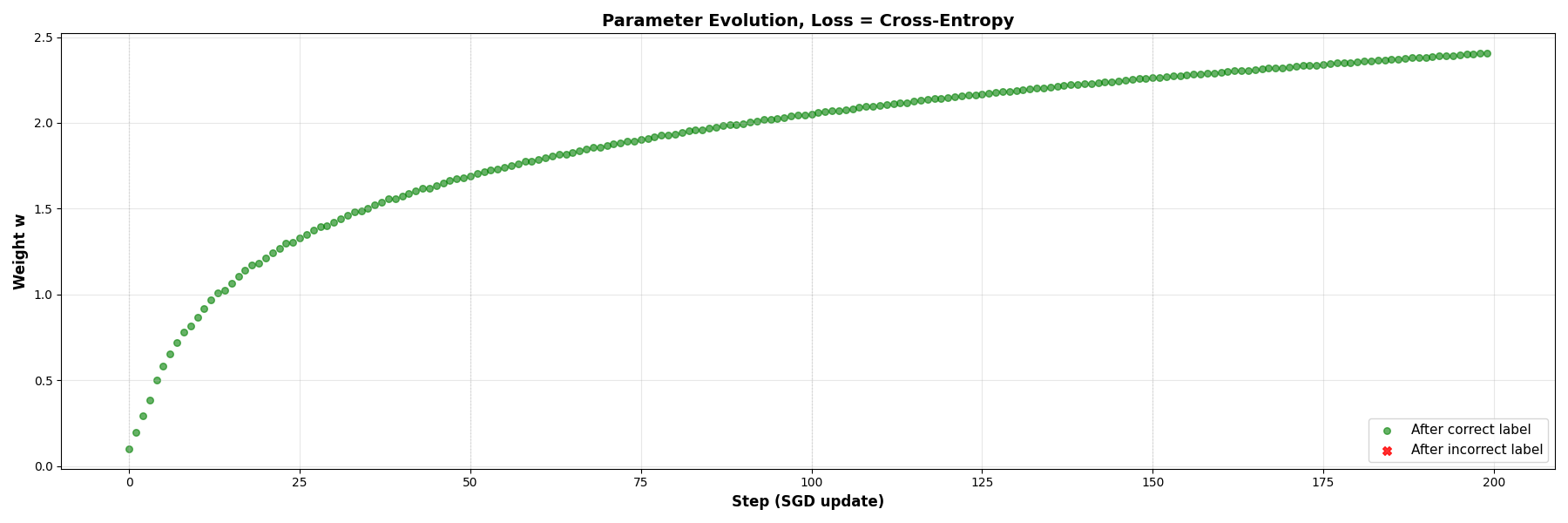}
        \caption{Parameter evolution showing impact of each label type.}
        \label{fig:toy_example_weights_ce_all_correct}
    \end{subfigure}
    \caption{Cross-Entropy loss behavior in a simple toy preference learning example with \textbf{noise-free} data.}
    \label{fig:toy_example_ce_all_correct}
\end{figure*}

Overall, this toy example illustrates how Soft-TAC naturally downweights inconsistent or noisy preference labels by driving their gradients toward zero, whereas Cross-Entropy continues to enforce large gradients on mislabeled comparisons, hindering stable reward learning. In the absence of label noise, both loss functions recover a reward that induces the correct preference ordering in this toy setting (see Figures \ref{fig:toy_example_st_all_correct} and \ref{fig:toy_example_ce_all_correct}).



\newpage
\section{GT7 Environment Details}\label{sec:env_details}

\subsection{Gran Turismo 7}\label{sec:env_details_GT}
GT7 is a high-fidelity racing simulator with multiple racing tracks, racing cars, and racing tasks. The agent's action space is 2-dimensional, with the agent controlling a car's velocity (e.g., accelerating or breaking) and steering input (e.g., turning left or right). The agent's state space consists of a variety of relevant features, such as an encoded static map of the track, its current position, velocity, and acceleration, as well as those of other opponents. 
For a complete description of the GT7 environment, see \cite{wurman2022outracing}.
The code for GT7 is proprietary.

\subsubsection{Reward Function Components}
The reward function to train a GT7 agent was defined as a linear combination of different features, and the features differed based on whether the task was time-trial or versus. We describe only the reward features we learned for each task below, and in Table \ref{tab:gt7_reward_weights} we outline the specific weights for each reward feature used in the default (i.e., not learned) reward function. Other reward components are based on those described \cite{wurman2022outracing}, along with a few additional parts that improved racing robustness. 
% The complete reward functions are confidential (i.e., not just the learned components) but contain additional parts such as those described in 

\begin{enumerate}
    \item \textbf{Time-trial}
    \begin{itemize}
        \item \textbf{Course Progress.}
        Measures the change in progress made along the track per step. More specifically, at each time step, we calculate the agent's distance from the closest point on the track's centerline. Then to calculate progress, we take the difference between the agent's current and previous distances. 
        
%         \item \textbf{Off-course Penalty.}
%     Measures the change in the total time (in seconds) the agent spends off track. For each tire, we maintain a cumulative counter of off-track time. At each timestep, we compute the per-tire increase in off-track time relative to the previous timestep. This value is then scaled by how fast the agent is driving (in kilometers). 
        
%         \item \textbf{Wall Collision.}
% Measures the change in the total amount of time (in seconds) the agent has been in contact with walls. We maintain a cumulative wall-contact time counter and, at each timestep, compute how much this value increased compared to the previous timestep. We then scale this delta value by how fast the agent is driving (in kilometers).
%         \item \textbf{Steering Penalty.}
% Measures the change in the steering input between consecutive time steps. This is used to discourage jittery steering.
%             \item \textbf{Steering History.}
%     Keeps a 3-step history of the steering input between consecutive time steps. This provides a penalty if the change in the steering input between consecutive time steps $i$ and $i+1$ is of a different sign than the change in the steering input between consecutive time steps $i+1$ and $i+2$. 
\end{itemize}

    \item \textbf{Versus Task}
    \begin{itemize}
%\item This contains the same five reward parts above as well as seven additional parts, outlined below.
        \item \textbf{Velocity-based Collision Penalty.}  This penalizes a car for colliding with another opponent and is based on the difference in velocity. More specifically, it is the squared difference in the car's velocity and the opponent's velocity at the time of the collision.
%         \item \textbf{Constant Car Collision Penalty.}
% Measures the change in the total amount of time (in seconds) the agent has been in contact with other cars. We keep a record of the cumulative time the agent has been in contact with any other car, and at each timestep we compute how much this value increased compared to the previous timestep. 

        \item \textbf{Overtaking.}
        This positively rewards the agent for overtaking opponents, but also penalizes the agent for being passed by an opponent. More specifically, we first find all opponents that are either 20 meters behind or 40 meters ahead of the agent. We refer to the set of these in-range opponents as $D_{opp-in-range}$. For each opponent $\in D_{opp-in-range}$, we calculate the change in the agent's relative position along the track for each opponent from the previous time step to the current time step. The reward is then a sum over all these deltas.

        % \item \textbf{Spin Out Penalty.}
        % Binary reward that is triggered if the agent's orientation relative to the track changes from mostly forward to mostly backward in a single timestep. No penalty is applied otherwise.
        % \item \textbf{Change in Centerline Distance Penalty.}
        % Measures the difference between consecutive time steps in the change in the centerline distance. The centerline distance is the distance from the center of the agent's car (in meters) to the nearest point on the track's centerline. This reward penalizes sudden changes in the car’s distance from the track centerline, encouraging smoother driving and discouraging excessive swerving.

        % \item \textbf{Tire Slip Penalty.}
        % Penalizes tires slipping in a direction different from their forward orientation. For each tire, we compute the slip angle scaled by the tire-slip ratio (capped at 1.0). The reward is then the sum of this scaled value across all tires.
        
        % \item \textbf{Fuel Consumption Penalty.}
        % Measures how much fuel was used by the agent between consecutive time steps.  

    \end{itemize}
\end{enumerate}

\begin{table}[h!]
\centering
\caption{GT7 Default Reward Weights}

\label{tab:gt7_reward_weights}
\begin{tabular}{l c c}
\toprule
\textbf{Reward Part} & \textbf{Versus} &  \textbf{Time-Trial} \\
\midrule
Course Progress & 1.0 & 1.0 \\
% Off-course & -10.0 & -10.0 \\
% Wall Collision & -50.0 & -10.0 \\
% Steering & -3.0 & -3.0 \\
% Steering History  & -5.0 & -5.0 \\
Velocity-based Collision & -0.5 & NA \\
% Constant Car Collision & -6.0 & NA \\
Overtaking Attempts & 0.5 & NA \\
% Spin Out & -3.0 & NA \\
% Change in Centerline Distance  & -3.0 & NA \\
% Tire Slip &  &  NA \\
% Fuel Consumption  &  & NA \\
\bottomrule
\end{tabular}
\end{table}

In the fast driving experiments for the time-trial task (RQ4-a), we learned the weight for the course progress reward part.  In the controlling aggressiveness/timidness experiments for the versus tasks (RQ4-b), we learned the velocity-based collision and overtaking reward weights. 

\subsubsection{Evaluation Metrics}
The performance of the fast driving experiments in the time-trial task was evaluated by the three metrics described  BIAI ratio, minimal lap time (in seconds), and the number of incomplete laps (i.e., lap time exceeds the maximum lap time of 1,800 seconds).

For the controlling aggressive/timid experiments, we describe the evaluation metrics in more detail.

% See also Figure \ref{fig:aggressive_timid_example} for examples of aggressive and timid driving behaviors, respectively.
\begin{enumerate}
        \item \textbf{Collisions.}
Evaluates the car's velocity toward an opponent when a collision is detected. When this value exceeds a set threshold, the counter for the number of collisions increases.
        \item \textbf{Off Track Instances.}
Tracks how many times the car has gone off the track by recording each instance when any of its four tires leave the track surface.  
        \item \textbf{Overtaking Attempts.}
We check for a sign change in the relative distance between consecutive time steps. A change from positive to negative indicates that the car has transitioned from being behind an opponent to being ahead in terms of forward progress along the track. If this sign change occurs while the car and its opponent are within 40 meters of each other, the event is classified as an overtaking attempt. 
        \item \textbf{Final Place.} Indicates the agent’s position rank at the end of the race out of 19 opponents. A final place of 0 refers to the first position, and 19 means the last.
 \end{enumerate}

\section{Preference Collection Details}\label{sec:preference_collection_details}

 \subsubsection{GT7 Cars used in Preference Dataset}\label{sec:car_pp}

\begin{figure}[h!]
    \centering
    % The \includegraphics command inserts your image file.
    \includegraphics[width=0.48\textwidth]{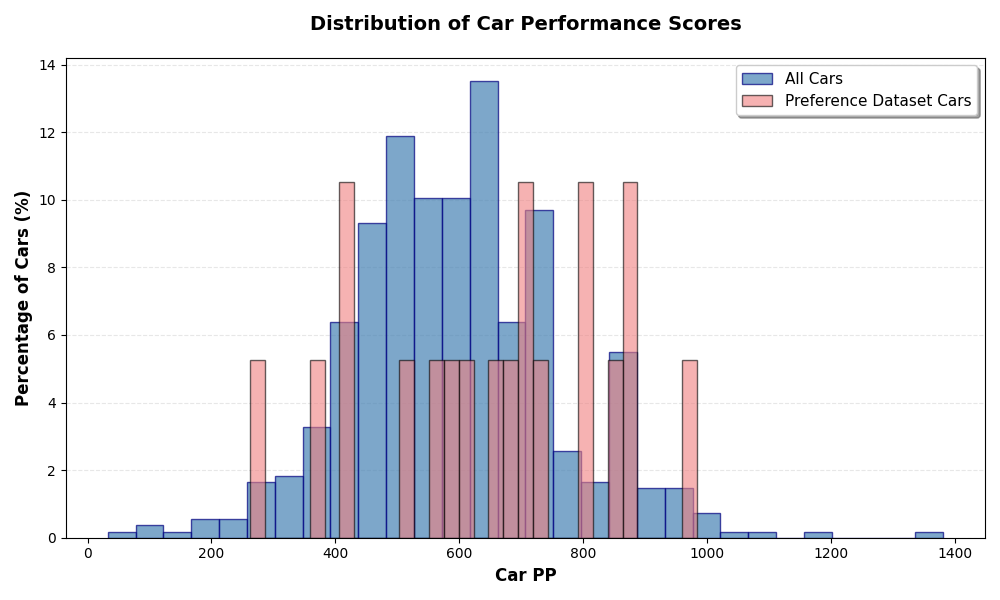}
    \caption{Distribution of Performance Points (PP) among all GT cars versus GT Cars in the preference dataset.}

    \label{fig:car_pp}
\end{figure}

In all GT7 experiments, we selected a subset of 19 cars to collect our preference dataset; it is not computationally feasible to collect data for all 552 cars available in GT7. These cars were chosen uniformly with respect to Performance Points (PP) to ensure good coverage across the full range of PP values. PP is an in-game metric measuring car performance, where higher PP indicates a more powerful and faster car. This selection ensures that the trajectories capture a diverse set of driving behaviors and performance levels. As shown in Figure~\ref{fig:car_pp}, the 19 selected cars provide broad PP coverage across the full set of cars.

\begin{figure}[h!]
    \centering
    % The \includegraphics command inserts your image file.
    \includegraphics[width=0.48\textwidth]{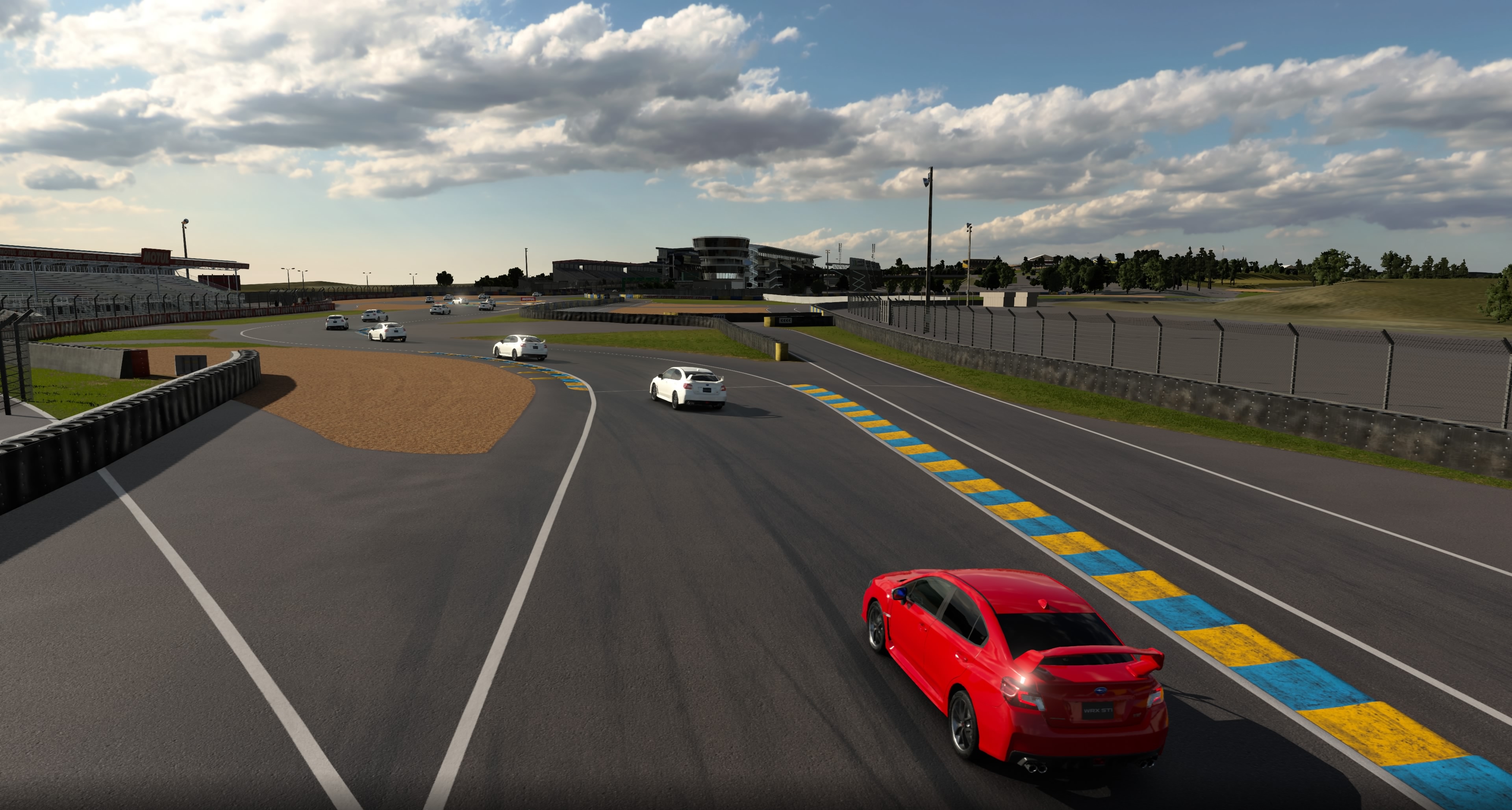}
    \caption{An illustration of the versus setup. The agent's car (in red) is placed in last place to chase 19 opponents (in white) cars in a three-lap race. Each opponent is evenly spaced ahead of the agent. Copyright \textcopyright\ 2025 Sony Interactive Entertainment Inc. Developed by Polyphony Digital Inc.}
    \label{fig:pursuit}
\end{figure}

\paragraph{GT7 Preference Dataset Construction}
To collect trajectories for the aggressiveness/timidness experiments, we used two types of versus tasks in which the agent started in 20th place and chased 19 opponents over a three-lap race. The tasks differed in the type of opponent the agent raced against: (1) BIAI opponents and (2) copies of the agent itself (see Figure \ref{fig:pursuit} for an illustration of the task setup).

Initially, we collected 45 trajectories and grouped them into nine sets of five, organized by car type and task type. Two trajectories were discarded from this initial set because the agent failed to complete the race, ensuring a high-quality dataset. Among the retained trajectories, four sets of nine were against copies of the agent itself, and the remaining five sets were against BIAI opponents.

We then collected an additional 36 trajectories, which were grouped into twelve sets of three using the same criteria. No trajectories were discarded from this second collection, all of which were races against BIAI opponents.

Of the final 79 trajectories, the agent’s car was drawn from a fixed set of 19 cars (as noted earlier). Across the two data collections, only two car types overlapped, resulting in eight trajectories for each of those car types. For the remaining car types, there were five trajectories per car type in the nine sets of five, except for the groupings that included discarded trajectories, which contained four. In the twelve sets of three, each car type had three trajectories.

This grouping facilitated easier labeling by domain experts, who ranked trajectories within each group from most aggressive to most timid. Note, the domain experts were not provided any metadata while giving preferences, they made their judgment based on the content of the trajectory videos alone. These rankings were subsequently converted into pairwise preferences.

For the fast driving time-trial experiments, we leveraged the available BIAI agents to generate 56 time-trial trajectories using the same set of 19 cars. We then followed a similar protocol, randomly sampling groups of five trajectories and automatically ranking them by lap time, from fastest to slowest. Because groups were sampled with replacement, the same trajectories could appear in multiple groups, resulting in repeated pairwise orderings. To avoid over-representing redundant information, we retained each unique pairwise preference only once in the final dataset, resulting in the 1429 trajectories.

\paragraph{Lunar Lander Preference Dataset Construction}
This is discussed earlier in the User Study section.
\paragraph{Reacher and Hopper Preference Dataset Construction}

For the Hopper and Reacher experiments, we leveraged the open-source user study data from \citet{rmse}. Their ethics-approved study collected discrete trajectory ratings from human participants (6 participants for Hopper and 7 participants for Reacher). In their setup, participants viewed videos of robot trajectories and assigned them a scalar rating where higher values correspond to higher trajectory quality (e.g., successful forward motion or target reaching) and lower values indicate poor behavior. To adapt these absolute ratings for our PbRL experiments, we constructed pairwise trajectory comparisons ($\tau_i, \tau_j$) from the dataset. A preference label was assigned such that $\tau_i \succ \tau_j$ whenever the assigned ratings satisfied $r(\tau_i) > r(\tau_j)$, while pairs with equal ratings ($r(\tau_i) = r(\tau_j)$) were  discarded. See Table \ref{tab:participant_preference_summary} for the number of ratings versus pairwise preferences constructed.

\begin{table}[h!]
\centering
\caption{Summary of human ratings and extracted strict binary preference pairs across participants for Reacher and Hopper environments.}
\label{tab:participant_preference_summary}
\begin{tabular}{c c c}
\toprule
\textbf{Participant} & \textbf{Original \# Ratings} & \textbf{\# Binary Preference Pairs} \\
\midrule
\multicolumn{3}{c}{\textbf{Reacher}} \\
\midrule
P1 & 159 & 9,195 \\
P2 & 159 & 10,103 \\
P3 & 155 & 8,880 \\
P4 & 159 & 9,781 \\
P5 & 159 & 9,482 \\
P6 & 159 & 10,795 \\
P7 & 159 & 11,181 \\
\midrule
\multicolumn{3}{c}{\textbf{Hopper}} \\
\midrule
P1 & 114 & 4,703 \\
P2 & 113 & 5,608 \\
P3 & 113 & 5,414 \\
P4 & 157 & 8,905 \\
P5 & 115 & 4,982 \\
P6 & 115 & 4,811 \\
\bottomrule
\end{tabular}
\end{table}

\section{Reward Learning Training Details }\label{sec:reward_learning_details}

We evaluated both linear and non-linear reward model architectures depending on the domain. All models were optimized using mini-batch gradient descent with Adam \citep{kingma2015adam}, tuning the learning rate and batch size via grid search per environment (and per participant for Reacher and Hopper). See Tables \ref{tab:gt7_reward_learning}, ~\ref{tab:lunar_lander_reward_learning}, and \ref{tab:nn_reward_learning} for the respective model architectures and grid search configurations, and Tables \ref{tab:reward_learning_hyperparams_gt_lunar_lander} and \ref{tab:reacher_hopper_hyperparams} for the final selected learning rates and batch sizes.
For the temperature hyperparameter ($\alpha$) used in both the Soft-TAC and Cross-Entropy loss functions, which scales differences in predicted returns, we fixed $\alpha = 1.0$. Because standard Cross-Entropy assumes $\alpha = 1.0$, using the same value for Soft-TAC ensures consistency across objectives.

To prevent overfitting, we employed early stopping during training. The early stopping procedure tracks the best model based on three metrics: TAC, accuracy, and loss. After each epoch, the model is considered to have improved if either of the following conditions holds:
\begin{enumerate}
    \item TAC increases (or remains constant) \emph{and} accuracy increases, or
    \item TAC increases (or remains constant) \emph{and} the loss decreases by more than $1 \times 10^{-4}$.
\end{enumerate}
Whenever an improvement is detected, the stored best model (including its metrics and weights) is updated and the counter of consecutive non-improving epochs is reset. If no improvement occurs, this counter is incremented. Training terminates once the model fails to improve for 50 consecutive epochs, at which point early stopping is triggered and the best checkpoint observed during training is returned.

\begin{table*}[h!]
\centering
\caption{Gran Turismo 7 (GT7) Linear Reward Model Configurations and Hyperparameter Search Grids.}
\label{tab:gt7_reward_learning}
\begin{tabular}{l c c}
\toprule
\textbf{Hyperparameter / Setting} & \textbf{GT7 (Aggressive--Timid)} & \textbf{GT7 (Fast Driving)} \\
\midrule
Model Type & Linear & Linear \\
Weight Initialization & $\mathcal{N}(0, 1)$ & $\mathcal{N}(0, 1)$ \\
Weight Clipping Bounds & $[0, 15]$ & $[0, 15]$ \\
Learning Rate Grid & $\{10^{-4}, 3\cdot10^{-4}, 5\cdot10^{-4}, 0.01, 0.03, 0.05\}$ & $\{10^{-4}, 3\cdot10^{-4}, 5\cdot10^{-4}, 0.01, 0.03, 0.05\}$ \\
Batch Size Grid & $\{4, 8, 12\}$ & $\{8, 16, 32\}$ \\
Max Epochs & 500 & 500 \\
Loss Temperature ($\alpha$) & 1.0 & 1.0 \\
\bottomrule
\end{tabular}
\end{table*}
\begin{table}[h!]
\centering
\caption{Lunar Lander Linear Reward Model Configuration and Hyperparameter Search Grid.}
\label{tab:lunar_lander_reward_learning}
\begin{tabular}{l c}
\toprule
\textbf{Hyperparameter / Setting} & \textbf{Lunar Lander} \\
\midrule
Model Type & Linear \\
Weight Initialization & $\mathcal{N}(0, 1)$ \\
Weight Clipping Bounds & $[-200, 200]$ \\
Learning Rate Grid & $\{10^{-4}, 3\cdot10^{-4}, 5\cdot10^{-4}, 0.01, 0.03, 0.05\}$ \\
Batch Size Grid & $\{8, 16, 24\}$ \\
Max Epochs & 5000 \\
Loss Temperature ($\alpha$) & 1.0 \\
\bottomrule
\end{tabular}
\end{table}

\begin{table*}[h!]
\centering
\caption{Reacher and Hopper Neural Network Ensemble Reward Model Configurations and Hyperparameter Search Grids.}
\label{tab:nn_reward_learning}
\begin{tabular}{l c c}
\toprule
\textbf{Hyperparameter / Setting} & \textbf{Reacher} & \textbf{Hopper} \\
\midrule
Model Type & Neural Network Ensemble & Neural Network Ensemble \\
Ensemble Size ($N$) & 3 & 3 \\
Hidden Layers & 3 & 3 \\
Hidden Units per Layer & 128 & 128 \\
Hidden Layer Activation & LeakyReLU & LeakyReLU \\
Output Layer Activation & Tanh & Tanh \\
Weight Initialization & Kaiming Uniform & Kaiming Uniform \\
Learning Rate Grid & $\{10^{-4}, 10^{-3}\}$ & $\{10^{-4}, 10^{-3}\}$ \\
Batch Size Grid & $\{128, 256, 512\}$ & $\{128, 256, 512\}$ \\
Max Epochs & 5000 & 5000 \\
Loss Temperature ($\alpha$) & 1.0 & 1.0 \\
\bottomrule
\end{tabular}
\end{table*}

\begin{table}[h!]
\centering
\caption{Hyperparameters used for Soft-TAC and Cross-Entropy in Lunar Lander and GT experiments.}
\label{tab:reward_learning_hyperparams_gt_lunar_lander}
\begin{tabular}{c c c}
\toprule
\textbf{Method} & \textbf{Learning Rate} & \textbf{Batch Size} \\
\midrule
\multicolumn{3}{c}{\textbf{Lunar Lander}} \\
\midrule
\textsc{Soft-TAC} & 0.05 & 16 \\
\textsc{Cross-Entropy} & 0.05 & 8 \\
\midrule
\multicolumn{3}{c}{\textbf{GT Fast Driving}} \\
\midrule
\textsc{Soft-TAC} & 0.05 & 8 \\
\textsc{Cross-Entropy} & 0.03 & 8 \\
\midrule
\multicolumn{3}{c}{\textbf{GT Aggressive}} \\
\midrule
\textsc{Soft-TAC} & 0.05 & 4 \\
\textsc{Cross-Entropy} & 0.05 & 8 \\
\midrule
\multicolumn{3}{c}{\textbf{GT Timid}} \\
\midrule
\textsc{Soft-TAC} & 0.05 & 4 \\
\textsc{Cross-Entropy} & 0.05 & 4 \\
\bottomrule
\end{tabular}
\end{table}

\begin{table}[htp!]
\centering
\caption{Hyperparameters used for Soft-TAC and Cross-Entropy across participants in the Reacher experiments.}
\label{tab:reacher_hopper_hyperparams}
\begin{tabular}{c c c c}
\toprule
\textbf{Method} & \textbf{Participant} & \textbf{Learning Rate} & \textbf{Batch Size} \\
\midrule
\multicolumn{4}{c}{\textbf{Cross-Entropy }} \\
\midrule
\textsc{Cross-Entropy} & P1 & 0.0001 & 512 \\
\textsc{Cross-Entropy} & P2 & 0.0010 & 256 \\
\textsc{Cross-Entropy} & P3 & 0.0010 & 128 \\
\textsc{Cross-Entropy} & P4 & 0.0001 & 512 \\
\textsc{Cross-Entropy} & P5 & 0.0010 & 128 \\
\textsc{Cross-Entropy} & P6 & 0.0010 & 256 \\
\textsc{Cross-Entropy} & P7 & 0.0010 & 128 \\
\midrule
\multicolumn{4}{c}{\textbf{Soft-TAC}} \\
\midrule
\textsc{Soft-TAC} & P1 & 0.0001 & 256 \\
\textsc{Soft-TAC} & P2 & 0.0010 & 256 \\
\textsc{Soft-TAC} & P3 & 0.0010 & 128 \\
\textsc{Soft-TAC} & P4 & 0.0001 & 512 \\
\textsc{Soft-TAC} & P5 & 0.0010 & 128 \\
\textsc{Soft-TAC} & P6 & 0.0010 & 256 \\
\textsc{Soft-TAC} & P7 & 0.0010 & 512 \\
\bottomrule
\end{tabular}
\end{table}

\begin{table}[]
\centering
\caption{Hyperparameters used for Soft-TAC and Cross-Entropy across participants in the Hopper experiments.}
\label{tab:hopper_hyperparams}
\begin{tabular}{c c c c}
\toprule
\textbf{Method} & \textbf{Participant} & \textbf{Learning Rate} & \textbf{Batch Size} \\
\midrule
\multicolumn{4}{c}{\textbf{Cross-Entropy }} \\
\midrule
\textsc{Cross-Entropy} & P1 & 0.0001 & 256 \\
\textsc{Cross-Entropy} & P2 & 0.0010 & 256 \\
\textsc{Cross-Entropy} & P3 & 0.0001 & 512 \\
\textsc{Cross-Entropy} & P4 & 0.0001 & 512 \\
\textsc{Cross-Entropy} & P5 & 0.0001 & 256 \\
\textsc{Cross-Entropy} & P6 & 0.0010 & 128 \\
\midrule
\multicolumn{4}{c}{\textbf{Soft-TAC}} \\
\midrule
\textsc{Soft-TAC} & P1 & 0.0010 & 256 \\
\textsc{Soft-TAC} & P2 & 0.0010 & 256 \\
\textsc{Soft-TAC} & P3 & 0.0010 & 256 \\
\textsc{Soft-TAC} & P4 & 0.0010 & 512 \\
\textsc{Soft-TAC} & P5 & 0.0010 & 512 \\
\textsc{Soft-TAC} & P6 & 0.0010 & 256 \\
\bottomrule
\end{tabular}
\end{table}

\clearpage
\section{Learned Linear Reward Functions for Lunar Lander and GT7}
In Table \ref{tab:learned_reward_parts_lunar_lander}, we present the learned Lunar Lander weights for all  all ten seeds for both Soft-TAC and Cross-Entropy. For reference, we also include the default environment reward weights.
Tables \ref{tab:reward_parts_aggressive_timid_gt} and \ref{tab:reward_parts_progress_gt} and report the corresponding learned reward parts for the GT experiments for all six seeds, covering both the aggressive–timid and the fast driving setting.

\begin{table*}[h!]
\centering
\caption{Learned Lunar Lander reward weights (Soft-TAC, Cross-Entropy, and Default) and resulting TAC scores.}
\label{tab:learned_reward_parts_lunar_lander}
\small
\begin{tabular}{l c c c c c c c c c}
\toprule
\textbf{} & \textbf{Distance} & \textbf{Speed} & \textbf{Tilt} & \textbf{Leg Contact} & \textbf{Side Engine} & \textbf{Main Engine} & \textbf{Crash} & \textbf{Safe Landing} & \textbf{TAC} \\
\midrule
\multicolumn{10}{c}{\textsc{Soft-TAC}} \\
\midrule
 & -6.34  & 0.16  & -18.02 & 4.81  & -0.26 & 0.18  & -32.92 & 99.29 & 0.88 \\
 & -4.58  & 0.84  & -9.00  & 4.19  & -0.18 & 0.14  & -25.38 & 37.63 & 0.86 \\
 & -6.03  & -1.81 & 0.08   & 3.95  & -0.07 & -0.08 & -15.42 & 22.79 & 0.85 \\
 & -10.79 & -3.74 & -3.95  & -0.72 & -0.29 & 0.07  & -27.93 & 60.92 & 0.84 \\
 & -4.98  & -1.35 & -3.95  & 4.74  & -0.08 & -0.03 & -15.98 & 24.70 & 0.84 \\
 & -11.22 & -3.52 & -3.18  & -0.63 & 0.02  & -0.26 & -26.82 & 43.06 & 0.84 \\
 & -4.26  & -1.56 & -3.26  & 1.72  & -0.05 & -0.06 & -9.77  & 8.79  & 0.84 \\
 & -3.60  & -0.39 & -5.69  & 1.70  & 0.04  & -0.08 & -11.56 & 11.29 & 0.82 \\
 & -3.64  & -1.92 & -3.36  & 0.93  & -0.05 & -0.06 & -8.09  & 8.09  & 0.82 \\
 & -5.56  & 0.20  & -2.31  & 0.34  & 0.05  & -0.11 & -12.19 & 12.43 & 0.81 \\
\midrule
\multicolumn{10}{c}{\textsc{Cross-Entropy}} \\
\midrule
 & -5.12  & -1.24 & -0.65 & 2.18 & -0.05 & -0.08 & -14.16 & 29.49 & 0.85 \\
 & -3.86  & -1.42 & -1.63 & 1.28 & -0.11 & 0.03  & -12.26 & 21.10 & 0.85 \\
 & -2.56  & -1.61 & -4.05 & 1.70 & -0.00 & -0.15 & -13.62 & 32.03 & 0.84 \\
 & -3.98  & -1.31 & -3.07 & 0.91 & -0.11 & -0.03 & -11.78 & 16.69 & 0.84 \\
 & -3.62  & -0.40 & -2.07 & 1.44 & -0.05 & 0.00  & -9.45  & 11.25 & 0.84 \\
 & -3.33  & -1.03 & -1.13 & 2.83 & -0.07 & -0.03 & -11.40 & 14.87 & 0.84 \\
 & -1.84  & -0.64 & -1.70 & 1.00 & -0.01 & -0.02 & -3.94  & 4.25  & 0.82 \\
 & -4.06  & -2.02 & -2.73 & 3.11 & -0.08 & -0.07 & -11.82 & 12.88 & 0.82 \\
 & -2.85  & -0.27 & -2.09 & 3.25 & -0.11 & 0.01  & -14.97 & 29.17 & 0.82 \\
 & -3.57  & -0.03 & -2.31 & 0.79 & -0.00 & -0.04 & -6.19  & 7.38  & 0.80 \\
\midrule
\multicolumn{10}{c}{\textsc{Default}} \\
\midrule
 & -100.0 & -100.0 & -100.0 & 10.0 & -0.03 & -0.3 & -100.0 & 100.0 & 0.62 \\
\bottomrule
\end{tabular}
\end{table*}

\begin{table*}[h!]
\centering
\caption{GT7 controlling aggressiveness and timidness: learned reward weights and resulting TAC scores.}
\label{tab:reward_parts_aggressive_timid_gt}
\small
\begin{tabular}{l l c c c}
\toprule
\textbf{Experiment} & \textbf{Method} & \textbf{Overtaking} & \textbf{Collision} & \textbf{TAC} \\
\midrule
\multirow{12}{*}{\textsc{Aggressive}}
 & \multirow{6}{*}{\textsc{Soft-TAC}} 
 & 2.73 & 0.05 & 0.77 \\
 &  & 2.80 & 0.09 & 0.77 \\
 &  & 2.75 & 0.03 & 0.77 \\
 &  & 2.76 & 0.00 & 0.77 \\
 &  & 2.82 & 0.10 & 0.77 \\
 &  & 2.81 & 0.09 & 0.77 \\
\cmidrule{2-5}
 & \multirow{6}{*}{\shortstack{\textsc{Cross-}\\\textsc{Entropy}}} 
 & 1.73 & 0.0 & 0.67 \\
 &  & 1.36 & 0.0 & 0.69 \\
 &  & 1.18 & 0.0 & 0.56 \\
 &  & 1.33 & 0.0 & 0.60 \\
 &  & 1.32 & 0.0 & 0.60 \\
 &  & 1.18 & 0.0 & 0.56 \\
\midrule
\multirow{12}{*}{\textsc{Timid}}
 & \multirow{6}{*}{\textsc{Soft-TAC}} 
 & 0.0 & 2.47 & -0.27 \\
 &  & 0.0 & 2.49 & -0.27 \\
 &  & 0.0 & 2.20 & -0.27 \\
 &  & 0.0 & 4.87 & -0.24 \\
 &  & 0.0 & 2.25 & -0.27 \\
 &  & 0.0 & 2.23 & -0.27 \\
\cmidrule{2-5}
 & \multirow{6}{*}{\shortstack{\textsc{Cross-}\\\textsc{Entropy}}} 
 & 0.0 & 0.82 & -0.32 \\
 &  & 0.0 & 0.82 & -0.32 \\
 &  & 0.0 & 0.83 & -0.32 \\
 &  & 0.0 & 1.73 & -0.27 \\
 &  & 0.0 & 0.82 & -0.32 \\
 &  & 0.0 & 0.83 & -0.32 \\
\bottomrule
\end{tabular}
\end{table*}

\begin{table}[h!]
\centering
\caption{GT7 fast driving experiments: learned progress weights and resulting TAC scores.}
\label{tab:reward_parts_progress_gt}
\begin{tabular}{l c c}
\toprule
\textbf{Method} & \textbf{Progress} & \textbf{TAC} \\
\midrule
\multirow{6}{*}{\textsc{Soft-TAC}} 
 & 7.98 & 0.97 \\
 & 7.23 & 0.97 \\
 & 7.62 & 0.97 \\
 & 8.18 & 0.97 \\
 & 3.94 & 0.97 \\
 & 8.20 & 0.97 \\
\midrule
\multirow{6}{*}{\textsc{Cross-Entropy}} 
 & 0.47 & 0.965 \\
 & 0.50 & 0.965 \\
 & 0.51 & 0.965 \\
 & 0.47 & 0.965 \\
 & 0.47 & 0.965 \\
 & 0.48 & 0.965 \\
\bottomrule
\end{tabular}
\end{table}

\clearpage
\section{RL Training Infrastructure and Compute 
Details}\label{sec:RL_training_eval}
For GT7 and Lunar Lander experiments, we trained asynchronously using a centralized learner with distributed rollout workers. 
In the Lunar Lander experiments, the learner was allocated a single GPU with 8\,GB of memory, and experience was collected by two rollout workers, each allocated one CPU and 2,048\,MiB of memory. 
In the GT7 experiments, rollouts were conducted across 15 PlayStation consoles (a mix of PS4 and PS5 units, the only supported platforms). The centralized GT7 learner utilized a single GPU and seven CPUs with a total memory allocation of 65\,GB. Further details on the GT7 infrastructure setup can be found in \citet{wurman2022outracing}.

For Reacher and Hopper experiments, training was conducted synchronously without distributed rollout workers. Each experiment was executed on a single GPU and one CPU, with 3,000\,MiB of allocated memory. 

See Tables~\ref{tab:rl_hyperparams} and \ref{tab:reacher_hopper_rl_hyperparams} for the complete RL hyperparameter configurations across all environments.

\begin{table*}[h!]
\centering
\caption{RL Hyperparameters used in both GT7 and Lunar Lander experiments.}
\label{tab:rl_hyperparams}
\begin{tabular}{l c c}
\toprule
\textbf{Hyperparameter} & \textbf{GT7} & \textbf{Lunar Lander} \\
\midrule
RL Algorithm & QR-SAC \citep{wurman2022outracing}& SAC \citep{sac} \\
Optimizer & Adam \citep{kingma2015adam} & Adam \\
Number of Quantiles & 32 & NA \\
Number of Critic Networks & 2 & 2 \\
Number of Target Critic Networks & 2 & 2 \\
Number of Policy Networks & 1 & 1 \\
Learnable Entropy & False & True \\
Fixed Entropy Regularization Coefficient & 0.01 & N/A \\
Entropy Learning Rate & N/A & $1 \times 10^{-4}$ \\
Target Entropy & N/A & -2 \\
Critic Learning Rate & $2.5 \times 10^{-5}$ & $5 \times 10^{-4}$ \\
Actor Learning Rate & $1.25 \times 10^{-5}$ & $5 \times 10^{-4}$ \\
Target Critics Learning Rate & 0.005 & 0.005\\
Discount Factor & 0.9896 & 0.99 \\
Batch Size & 512 & 32 \\
Replay Buffer Size & 10000000 & 20000 \\
Global Norm of Critics Gradient Clipping & 10.0 & 10.0 \\
Number of Training Epochs & $\sim$2200 & 10000 \\
Number of Gradient Steps per Epoch & 6000 & 200 \\
\bottomrule
\end{tabular}
\end{table*}

\begin{table*}[h!]
\centering
\caption{RL Hyperparameters used in Reacher and Hopper experiments.}
\label{tab:reacher_hopper_rl_hyperparams}
\begin{tabular}{l c c}
\toprule
\textbf{Hyperparameter} & \textbf{Reacher} & \textbf{Hopper} \\
\midrule
RL Algorithm & SAC \citep{sac} & SAC \citep{sac} \\
Optimizer & Adam \citep{kingma2015adam} & Adam \citep{kingma2015adam} \\
Optimizer Betas ($\beta_1, \beta_2$) & (0.9, 0.999) & (0.9, 0.999) \\
Number of Critic Networks & 2 & 2 \\
Number of Target Critic Networks & 2 & 2 \\
Number of Policy Networks & 1 & 1 \\
Network Hidden Layers & 2 & 2 \\
Network Hidden Dimension & 1024 & 1024 \\
Learnable Entropy & True & True \\
Initial Entropy Regularization Coefficient ($\alpha$) & 0.1 & 0.1 \\
Entropy Learning Rate & $1 \times 10^{-4}$ & $1 \times 10^{-4}$ \\
Actor Learning Rate & $1 \times 10^{-4}$ & $1 \times 10^{-4}$ \\
Critic Learning Rate & $1 \times 10^{-4}$ & $1 \times 10^{-4}$ \\
Actor Update Frequency & 1 & 1 \\
Target Critic Update Frequency & 2 & 2 \\
Target Critics Update Rate & 0.005 & 0.005 \\
Global Norm of Critics Gradient Clipping & N/A & N/A \\
Discount Factor ($\gamma$) & 0.99 & 0.99 \\
Batch Size & 1024 & 1024 \\
Replay Buffer Capacity & 1,000,000 & 1,000,000 \\
Number of Seed Steps & 1,000 & 1,000 \\
Number of Gradient Steps per Epoch & \multicolumn{2}{c}{Equal to env steps per epoch (i.e., 1 gradient step per env step)} \\
Number of Training Epochs & 2,000 & 2,000 \\
\bottomrule
\end{tabular}
\end{table*}

\clearpage
\section{Policy Selection and Evaluation Details}\label{sec:policy_selection}
After training the RL policies, we then selected policies for evaluation based on certain criteria. 

In GT7 specifically, we followed a specific policy selection protocol, as previous work has shown that even after the training has converged, the policies of different training checkpoints can exhibit subtle but meaningful differences in driving behavior \citep{wurman2022outracing}.  More specifically, during training, we partially evaluated and saved the current policy (i.e., with exploration disabled) every five epochs. By partial evaluation, we mean that policies were evaluated on a small subset of cars rather than the full set of 552, as evaluating all cars at this frequency would be prohibitively expensive.

To select policies for evaluation, we employed a two-step selection protocol. First, we filtered candidate policies based on whether their evaluation metrics exceeded predefined thresholds, considering only metrics associated with the driving style of interest. For example, as time-trial tasks focused on fast driving, we required the BIAI ratio to be lower than 0.979. In versus tasks targeting aggressiveness or timidness, we considered metrics for overtaking attempts, lap time, and final place (as described in Appendix \ref{sec:env_details_GT}). For aggressive behaviors, we filtered out the slowest 25\% of policies based on final place and minimum lap time, as well as any policies with fewer than 7 overtaking attempts. For timid behaviors, we selected policies that had the lowest 15\% of overtaking attempts. We found that using these distinct criteria produced the most aggressive and most timid policies, rather than simply taking opposite extremes of the same metrics.

\begin{figure*}[h!]
    \centering
    \begin{subfigure}[b]{0.45\textwidth}
        \centering
        \includegraphics[width=\textwidth]{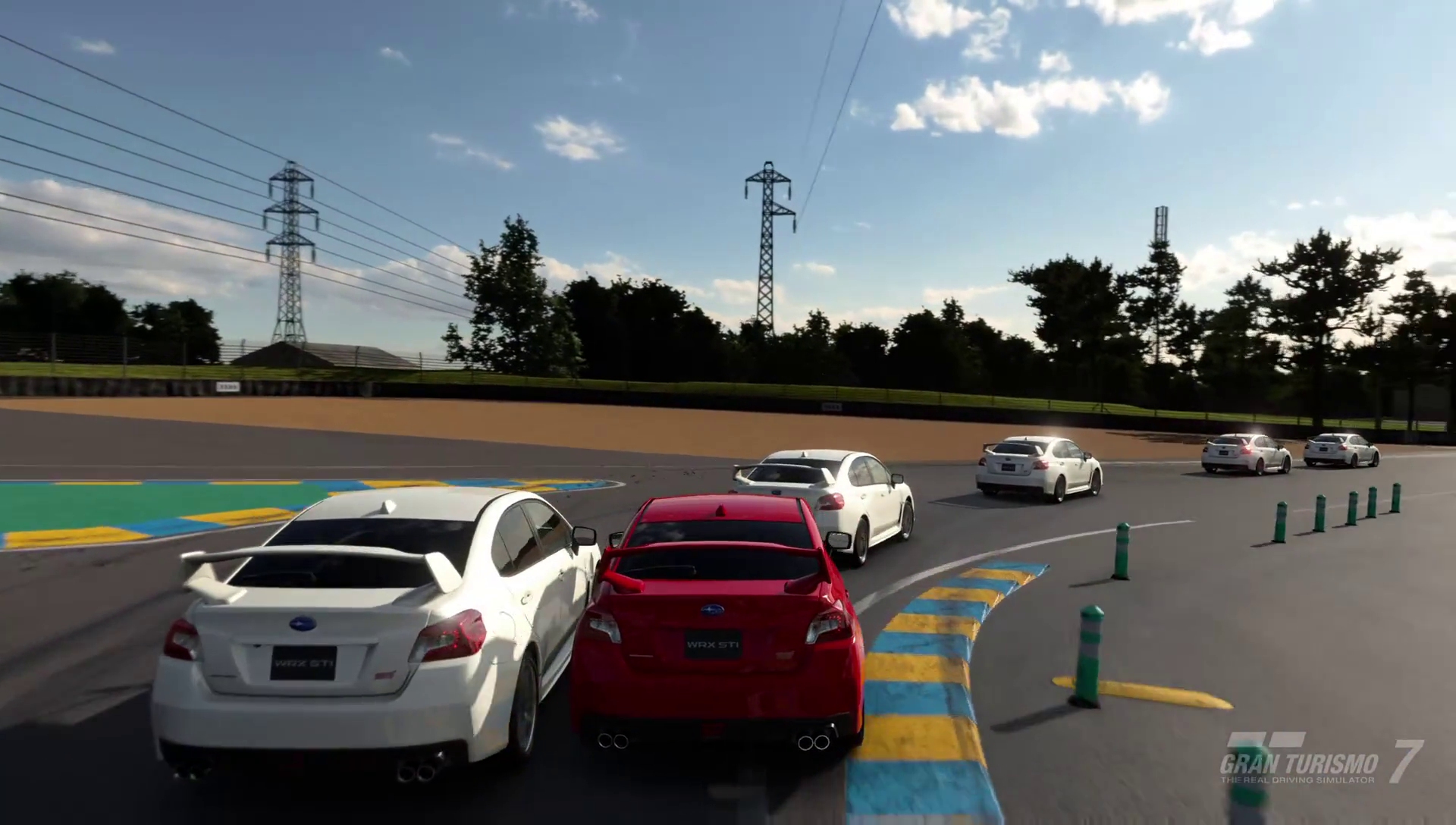}
        \caption{}
        \label{fig:aggressive_example}
    \end{subfigure}
    \hfill
    \begin{subfigure}[b]{0.45\textwidth}
        \centering
        \includegraphics[width=\textwidth]{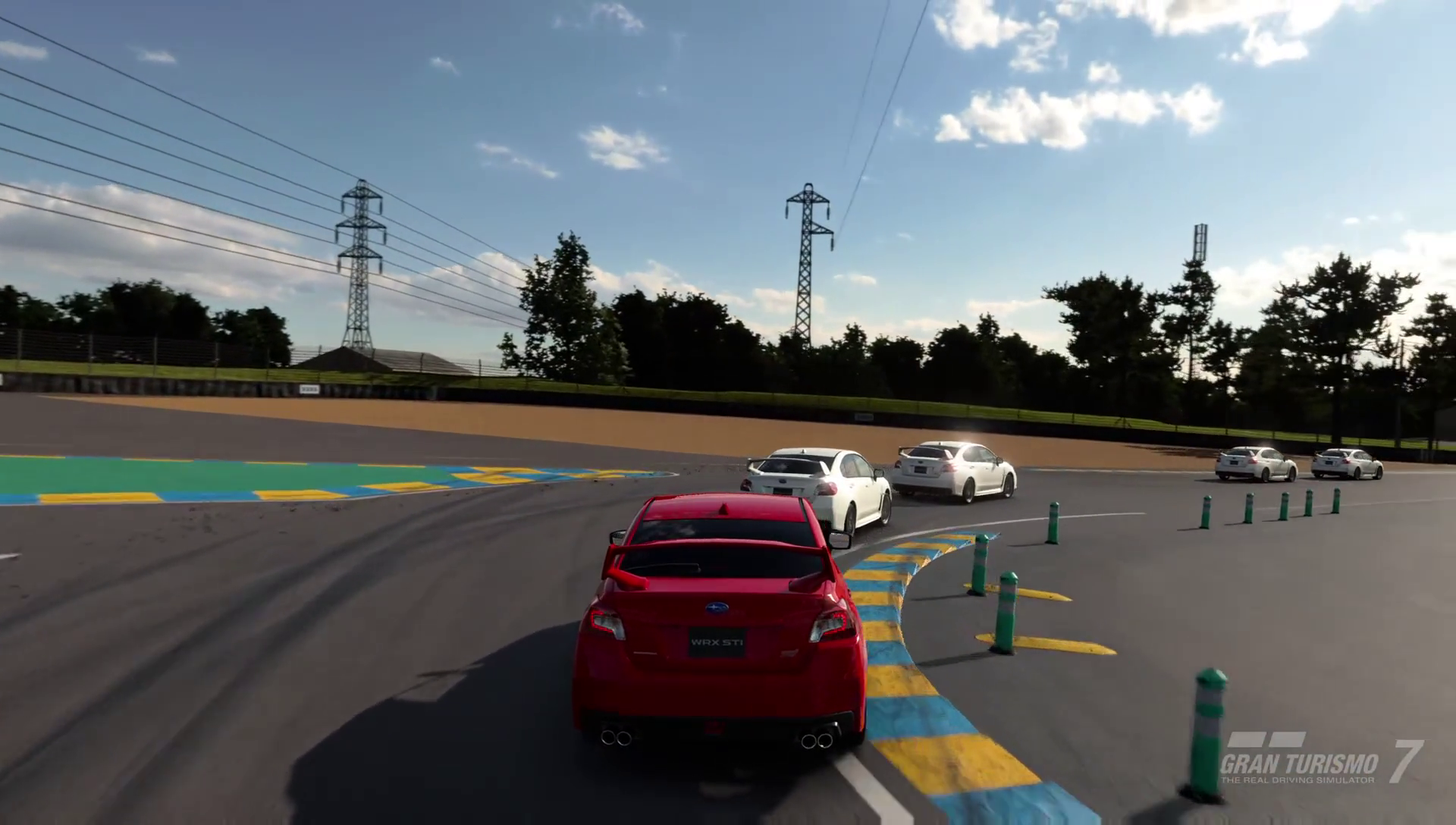} % 
        \caption{}
        \label{fig:timid_example}
    \end{subfigure}

    \caption{Example Soft-TAC learned driving style for aggressive (a) and timid (b). An aggressive agent will engage with opponents and attempt to overtake, resulting in more car interactions such as collisions. Whereas a timid agent will keep a distance from opponents, with minimal car contacts.}
    \label{fig:aggressive_timid_example}
\end{figure*}

Second, among the filtered policies, we performed a full evaluation on all 552 cars and selected the best policy to report. By best, we mean the policy that achieved the most favorable metrics for the task. For the time-trial task, we selected the policy with the fastest lap time; for the aggressive-timid experiments, we selected policies exhibiting the most aggressive or timid behaviors. Figure \ref{fig:aggressive_timid_example} shows examples of aggressive and timid driving learned with the Soft-TAC objective.

For Lunar Lander, Reacher and Hopper we just selected a few policies at the end of training for evaluation. 
See Table~\ref{tab:evaluation_settings} for details on our evaluation protocol hyperparameters.

\begin{table}[h!]
\centering
\caption{Evaluation Protocol Settings across Environments.}
\label{tab:evaluation_settings}
\begin{tabular}{l c c c}
\toprule
\textbf{Environment} & \textbf{Evaluation Seeds} & \textbf{Eval Episodes / Policy} & \textbf{Eval Policies (Episode Checkpoints)} \\
\midrule
GT7 (Timid Driving)        & 6  & 1  & N/A (special policy selection) \\
GT7 (Fast Driving)         & 6  & 1  & N/A (special policy selection) \\
GT7 (Aggressive Driving)   & 9  & 1  & N/A (special policy selection) \\
Lunar Lander               & 10 & 5  & 5 (Epochs 8000, 8500, 9000, 9500, 10000) \\
Lunar Lander (User Study)  & 5  & 5  & 5 (Epochs 8000, 8500, 9000, 9500, 10000) \\
Reacher                    & 10 & 10 & 10 (Last 10 Epochs: 4910--5000, step 10) \\
Hopper                     & 10 & 10 & 10 (Last 10 Epochs: 1910--2000, step 10) \\
\bottomrule
\end{tabular}
\end{table}
\clearpage
\section{Descriptive Statistics and Statistical Testing for Main Experiments}

\begin{table*}[h]
\centering
\caption{Lunar Lander Reward Learning Experiments: Shapiro-Wilk normality tests ($p_{\text{SW}}$) and statistical comparisons of Soft-TAC (ST) against Cross-Entropy (CE). Because metrics violated normality ($p_{\text{SW}} \le 0.05$), Mann–Whitney $U$ test statistics ($U$) are reported. Unadjusted $p$-values ($p$) and Benjamini–Hochberg adjusted $p$-values ($p_{\text{BH}}$) control the False Discovery Rate across $m = 2$ comparisons.}
\label{tab:lunar_lander_st_vs_ce}
\small
\setlength{\tabcolsep}{4pt}
\begin{tabular}{l c c c c c c c}
\toprule
 & \multicolumn{2}{c}{\textbf{ST}} & \multicolumn{2}{c}{\textbf{CE}} & \multicolumn{3}{c}{\textbf{ST vs.\ CE}} \\
\cmidrule(lr){2-3} \cmidrule(lr){4-5} \cmidrule(lr){6-8}
\textbf{Metric} & \textbf{Mean $\pm$ SE} & \textbf{$p_{\text{SW}}$} & \textbf{Mean $\pm$ SE} & \textbf{$p_{\text{SW}}$} & \textbf{Stat} & \textbf{Raw $p$} & \textbf{$p_{\text{BH}}$} \\
\midrule
Landing-pad success rate     & $0.6960 \pm 0.1238$ & 0.006 & $0.5000 \pm 0.1667$ & 0.000 & $U = 57.5$ & 0.287 & 0.287 \\
Outside landing success rate & $0.1360 \pm 0.0704$ & 0.000 & $0.0000 \pm 0.0000$ & 1.000 & $U = 65.0$ & 0.039 & 0.078 \\
\bottomrule
\end{tabular}
\end{table*}
\begin{table*}[h]
\centering
\caption{Comparing learned reward weights (ST) against participant-tuned reward weights (Alignment and Control conditions) on Lunar Lander landing-pad success rate. Descriptive statistics (Mean $\pm$ SE), Shapiro-Wilk normality tests ($p_{\text{SW}}$), and Mann–Whitney $U$ test statistics ($U$) with unadjusted $p$-values ($p$) and Benjamini–Hochberg adjusted $p$-values ($p_{\text{BH}}$) are reported ($m = 2$ comparisons).}
\label{tab:lunar_lander_st_vs_tuned}
\small
\setlength{\tabcolsep}{4pt}
\begin{tabular}{l c c c c c c c}
\toprule
 & \multicolumn{2}{c}{\textbf{ST}} & \multicolumn{2}{c}{\textbf{Tuned Condition}} & \multicolumn{3}{c}{\textbf{ST vs.\ Tuned}} \\
\cmidrule(lr){2-3} \cmidrule(lr){4-5} \cmidrule(lr){6-8}
\textbf{Comparison} & \textbf{Mean $\pm$ SE} & \textbf{$p_{\text{SW}}$} & \textbf{Mean $\pm$ SE} & \textbf{$p_{\text{SW}}$} & \textbf{Stat} & \textbf{Raw $p$} & \textbf{$p_{\text{BH}}$} \\
\midrule
ST vs.\ Alignment & $0.6960 \pm 0.1238$ & 0.006 & $0.6267 \pm 0.1840$ & 0.083 & $U = 32.5$ & 0.412 & 0.412 \\
ST vs.\ Control   & $0.6960 \pm 0.1238$ & 0.006 & $0.1984 \pm 0.1984$ & 0.000 & $U = 42.0$ & 0.019 & 0.038 \\
\bottomrule
\end{tabular}
\end{table*}
% --- Table 18: Reacher & Hopper Returns ---
\begin{table*}[h]
\centering
\caption{Reacher and Hopper Experiments: Summary of final environment returns per participant ($n=10$ runs per group) evaluated using learned reward models from Soft-TAC (ST) and Cross-Entropy (CE). Results are reported as Mean $\pm$ SE, where higher values indicate better performance.}
\label{tab:reacher_hopper_descriptive_stats}
\small
\setlength{\tabcolsep}{8pt}
\begin{tabular}{l c c}
\toprule
 & \multicolumn{2}{c}{\textbf{Final Return (Mean $\pm$ SE)}} \\
\cmidrule(lr){2-3}
\textbf{Participant} & \textbf{Cross-Entropy (CE)} & \textbf{Soft-TAC (ST)} \\
\midrule
\multicolumn{3}{l}{\textbf{Reacher}} \\
P1 & $-14.39 \pm 0.45$ & $-13.29 \pm 0.93$ \\
P2 & $-53.17 \pm 3.76$ & $-21.00 \pm 5.27$ \\
P3 & $-17.59 \pm 1.32$ & $-14.16 \pm 1.61$ \\
P4 & $-31.35 \pm 5.38$ & $-22.68 \pm 0.35$ \\
P5 & $-51.77 \pm 6.96$ & $-48.44 \pm 4.16$ \\
P6 & $-15.50 \pm 0.55$ & $-20.25 \pm 1.49$ \\
P7 & $-30.27 \pm 1.72$ & $-23.00 \pm 5.34$ \\
\midrule
\multicolumn{3}{l}{\textbf{Hopper}} \\
P1 & $973.03 \pm 142.58$  & $932.40 \pm 32.35$  \\
P2 & $1018.69 \pm 49.48$  & $955.06 \pm 101.46$ \\
P3 & $759.16 \pm 126.86$  & $1127.42 \pm 70.52$ \\
P4 & $937.62 \pm 87.40$   & $951.82 \pm 110.23$ \\
P5 & $1120.57 \pm 159.28$ & $955.84 \pm 67.86$  \\
P6 & $1138.11 \pm 66.65$  & $1142.97 \pm 104.07$ \\
\bottomrule
\end{tabular}
\end{table*}
\begin{table*}[h]
\centering
\caption{Paired statistical evaluation of final environment returns across all participants for Reacher ($n=7$ pairs) and Hopper ($n=6$ pairs). Shapiro-Wilk normality test $p$-values on paired differences ($p_{\text{SW}}$), test statistics ($t$ or $W$), and $p$-values are reported. A paired $t$-test is reported where $p_{\text{SW}} > 0.05$; otherwise, a Wilcoxon signed-rank test is reported.}
\label{tab:reacher_hopper_paired_tests_conditional}
\small
\setlength{\tabcolsep}{10pt}
\begin{tabular}{l c c c}
\toprule
\textbf{Environment} & \textbf{Normality ($p_{\text{SW}}$)} & \textbf{Test Statistic} & \textbf{$p$-value} \\
\midrule
Reacher ($n=7$) & 0.042 & $W = 24.0$ & 0.055 \\
Hopper ($n=6$)  & 0.083 & $t = 0.26$  & 0.401 \\
\bottomrule
\end{tabular}
\end{table*}

% --- Table: GT Experiments, Fast Driving Style Statistical Comparisons ---
\begin{table*}[h]
\centering
\caption{GT Experiments, Fast Driving Style: Shapiro-Wilk normality tests ($p_{\text{SW}}$) and statistical comparisons of Soft-TAC (ST) against Cross-Entropy (CE) and the Default reward. Because ST violated normality across all metrics ($p_{\text{SW}} \le 0.031$), Mann–Whitney $U$ test statistics ($U$) are reported for all comparisons. Unadjusted $p$-values ($p$) and Benjamini–Hochberg adjusted $p$-values ($p_{\text{BH}}$) control the False Discovery Rate across $m = 6$ comparisons.}
\label{tab:gt_fast_driving_hypothesis_tests}
\small
\setlength{\tabcolsep}{4pt}
\begin{tabular}{l c c c c c c}
\toprule
 & \multicolumn{3}{c}{\textbf{Descriptive Statistics (Mean $\pm$ SE)}} & \multicolumn{3}{c}{\textbf{Normality Tests ($p_{\text{SW}}$)}} \\
\cmidrule(lr){2-4} \cmidrule(lr){5-7}
\textbf{Metric} & \textbf{CE} & \textbf{ST} & \textbf{Default} & \textbf{CE} & \textbf{ST} & \textbf{Default} \\
\midrule
Lap Time (s)    & $277.71 \pm 0.31$   & $270.72 \pm 0.73$   & $276.03 \pm 0.21$   & 0.820 & 0.001 & 0.047 \\
BIAI Ratio      & $0.9961 \pm 0.0008$ & $0.9711 \pm 0.0026$ & $0.9908 \pm 0.0008$ & 0.562 & 0.001 & 0.117 \\
Incomplete Laps & $11.00 \pm 1.95$    & $1.17 \pm 0.40$     & $5.00 \pm 1.44$     & 0.596 & 0.031 & 0.489 \\
\midrule
 & \multicolumn{3}{c}{\textbf{ST vs.\ CE}} & \multicolumn{3}{c}{\textbf{ST vs.\ Default}} \\
\cmidrule(lr){2-4} \cmidrule(lr){5-7}
\textbf{Metric} & \textbf{Stat} & \textbf{Raw $p$} & \textbf{$p_{\text{BH}}$} & \textbf{Stat} & \textbf{Raw $p$} & \textbf{$p_{\text{BH}}$} \\
\midrule
Lap Time (s)    & $U = 0.0$ & 0.001 & 0.003 & $U = 0.0$ & 0.001 & 0.003 \\
BIAI Ratio      & $U = 0.0$ & 0.002 & 0.003 & $U = 0.0$ & 0.002 & 0.003 \\
Incomplete Laps & $U = 0.0$ & 0.002 & 0.003 & $U = 4.5$ & 0.015 & 0.015 \\
\bottomrule
\end{tabular}
\end{table*}
% --- Table: GT Experiments, Aggressive Driving Style Statistical Comparisons ---
\begin{table*}[h]
\centering
\caption{GT7 Experiments, Aggressive Driving Style: Shapiro-Wilk normality tests ($p_{\text{SW}}$) and statistical comparisons of Soft-TAC (ST) against Cross-Entropy (CE) and the Default reward. For normally distributed metrics ($p_{\text{SW}} > 0.05$), Welch's $t$-test statistics ($t$) are reported; otherwise, Mann–Whitney $U$ test statistics ($U$) are reported. Unadjusted $p$-values ($p$) and Benjamini–Hochberg adjusted $p$-values ($p_{\text{BH}}$) control the False Discovery Rate across $m = 8$ comparisons.}
\label{tab:gt_aggressive_driving_hypothesis_tests}
\small
\setlength{\tabcolsep}{4pt}
\begin{tabular}{l c c c c c c}
\toprule
 & \multicolumn{3}{c}{\textbf{Descriptive Statistics (Mean $\pm$ SE)}} & \multicolumn{3}{c}{\textbf{Normality Tests ($p_{\text{SW}}$)}} \\
\cmidrule(lr){2-4} \cmidrule(lr){5-7}
\textbf{Metric} & \textbf{CE} & \textbf{ST} & \textbf{Default} & \textbf{CE} & \textbf{ST} & \textbf{Default} \\
\midrule
Overtaking Attempts & $3.60 \pm 0.77$   & $4.03 \pm 0.55$   & $4.02 \pm 0.57$   & 0.693 & 0.240 & 0.184 \\
Collisions          & $18.98 \pm 4.32$  & $24.05 \pm 3.11$  & $9.28 \pm 1.46$   & 0.724 & 0.869 & 0.232 \\
Off-Track Instances & $3.94 \pm 0.58$   & $7.90 \pm 0.49$   & $3.65 \pm 1.03$   & 0.188 & 0.328 & 0.048 \\
Final Place         & $16.62 \pm 0.65$  & $16.63 \pm 0.39$  & $16.01 \pm 0.53$  & 0.850 & 0.904 & 0.406 \\
\midrule
 & \multicolumn{3}{c}{\textbf{ST vs.\ CE}} & \multicolumn{3}{c}{\textbf{ST vs.\ Default}} \\
\cmidrule(lr){2-4} \cmidrule(lr){5-7}
\textbf{Metric} & \textbf{Stat} & \textbf{Raw $p$} & \textbf{$p_{\text{BH}}$} & \textbf{Stat} & \textbf{Raw $p$} & \textbf{$p_{\text{BH}}$} \\
\midrule
Overtaking Attempts & $t = 0.455$ & 0.330 & 0.528 & $t = 0.009$ & 0.497 & 0.578 \\
Collisions          & $t = 0.954$ & 0.182 & 0.365 & $t = 4.302$ & 0.002 & 0.007 \\
Off-Track Instances & $t = 5.196$ & $< 0.001$ & 0.002 & $U = 33.0$ & 0.008 & 0.020 \\
Final Place         & $t = 0.015$ & 0.506 & 0.578 & $t = 0.945$ & 0.816 & 0.816 \\
\bottomrule
\end{tabular}
\end{table*}
% --- Table 20: GT7 Timid ---
% --- Table: GT Experiments, Timid Driving Style Statistical Comparisons ---
\begin{table*}[h]
\centering
\caption{GT7 Experiments, Timid Driving Style: Shapiro-Wilk normality tests ($p_{\text{SW}}$) and statistical comparisons of Soft-TAC (ST) against Cross-Entropy (CE) and the Default reward. For normally distributed metrics ($p_{\text{SW}} > 0.05$), Welch's $t$-test statistics ($t$) are reported; otherwise, Mann–Whitney $U$ test statistics ($U$) are reported. Unadjusted $p$-values ($p$) and Benjamini–Hochberg adjusted $p$-values ($p_{\text{BH}}$) control the False Discovery Rate across $m = 8$ comparisons.}
\label{tab:gt_timid_driving_hypothesis_tests}
\small
\setlength{\tabcolsep}{4pt}
\begin{tabular}{l c c c c c c}
\toprule
 & \multicolumn{3}{c}{\textbf{Descriptive Statistics (Mean $\pm$ SE)}} & \multicolumn{3}{c}{\textbf{Normality Tests ($p_{\text{SW}}$)}} \\
\cmidrule(lr){2-4} \cmidrule(lr){5-7}
\textbf{Metric} & \textbf{CE} & \textbf{ST} & \textbf{Default} & \textbf{CE} & \textbf{ST} & \textbf{Default} \\
\midrule
Overtaking Attempts & $2.06 \pm 0.46$   & $1.30 \pm 0.19$   & $2.72 \pm 0.43$   & 0.979 & 0.330 & 0.635 \\
Collisions          & $2.47 \pm 0.76$   & $1.00 \pm 0.20$   & $4.40 \pm 1.06$   & 0.369 & 0.705 & 0.725 \\
Off-Track Instances & $2.75 \pm 1.31$   & $0.61 \pm 0.18$   & $2.48 \pm 0.42$   & 0.020 & 0.136 & 0.463 \\
Final Place         & $17.47 \pm 0.39$  & $18.19 \pm 0.16$  & $16.98 \pm 0.31$  & 0.949 & 0.573 & 0.775 \\
\midrule
 & \multicolumn{3}{c}{\textbf{ST vs.\ CE}} & \multicolumn{3}{c}{\textbf{ST vs.\ Default}} \\
\cmidrule(lr){2-4} \cmidrule(lr){5-7}
\textbf{Metric} & \textbf{Stat} & \textbf{Raw $p$} & \textbf{$p_{\text{BH}}$} & \textbf{Stat} & \textbf{Raw $p$} & \textbf{$p_{\text{BH}}$} \\
\midrule
Overtaking Attempts & $t = -1.530$ & 0.086 & 0.086 & $t = -3.044$ & 0.010 & 0.023 \\
Collisions          & $t = -1.869$ & 0.057 & 0.086 & $t = -3.159$ & 0.011 & 0.023 \\
Off-Track Instances & $U = 6.0$    & 0.032 & 0.052 & $t = -4.061$ & 0.003 & 0.018 \\
Final Place         & $t = 1.704$  & 0.067 & 0.086 & $t = 3.472$  & 0.005 & 0.018 \\
\bottomrule
\end{tabular}
\end{table*}
\begin{table*}[h]
\centering
\caption{GT Experiments, Cross-Condition Comparison (Aggressive vs.\ Timid): Shapiro-Wilk normality tests ($p_{\text{SW}}$) and statistical comparisons of Soft-TAC (ST) under Aggressive versus Timid driving styles. For normally distributed metrics ($p_{\text{SW}} > 0.05$), Welch's $t$-test statistics ($t$) are reported. Unadjusted $p$-values ($p$) and Benjamini–Hochberg adjusted $p$-values ($p_{\text{BH}}$) control the False Discovery Rate across $m = 4$ comparisons.}
\label{tab:gt_st_aggressive_vs_timid_hypothesis_tests}
\small
\setlength{\tabcolsep}{4pt}
\begin{tabular}{l c c c c c c c}
\toprule
 & \multicolumn{2}{c}{\textbf{Descriptive Statistics (Mean $\pm$ SE)}} & \multicolumn{2}{c}{\textbf{Normality Tests ($p_{\text{SW}}$)}} & \multicolumn{3}{c}{\textbf{ST Aggressive vs.\ ST Timid}} \\
\cmidrule(lr){2-3} \cmidrule(lr){4-5} \cmidrule(lr){6-8}
\textbf{Metric} & \textbf{ST Aggressive} & \textbf{ST Timid} & \textbf{ST Aggressive} & \textbf{ST Timid} & \textbf{Stat} & \textbf{Raw $p$} & \textbf{$p_{\text{BH}}$} \\
\midrule
Overtaking Attempts & $4.03 \pm 0.55$   & $1.30 \pm 0.19$ & 0.240 & 0.330 & $t = 4.667$   & 0.002 & 0.002 \\
Collisions          & $24.05 \pm 3.11$  & $1.00 \pm 0.20$ & 0.869 & 0.705 & $t = 7.401$   & $< 0.001$ & $< 0.001$ \\
Off-Track Instances & $7.90 \pm 0.49$   & $0.61 \pm 0.18$ & 0.328 & 0.136 & $t = 13.865$  & $< 0.001$ & $< 0.001$ \\
Final Place         & $16.63 \pm 0.39$  & $18.19 \pm 0.16$ & 0.904 & 0.573 & $t = -3.668$  & 0.004 & 0.004 \\
\bottomrule
\end{tabular}
\end{table*}
% --- Table 21: Reacher & Hopper Hypothesis Tests Side-by-Side ---

\clearpage
\section{Generalization of Learned Reward Models in GT7}
In GT7, we were also interested in whether reward functions learned from preferences on a single track can generalize to produce effective policies on an unseen track—that is, whether the learned reward functions are specific to the track in the preference dataset. We tested this in the fast driving time-trial experiment. In this experiment, we reused the reward function that was learned only using the preference dataset collected on the Sarthe track to train a QR-SAC agent on the Autodrome Lago Maggiore (Maggiore) track. We chose Maggiore because of its technical layout, including complex cornering sequences and elevation management. These track characteristics contrast with the high-speed, longitudinal nature of Sarthe. This setup allows us to assess whether reward functions can generalize to tracks that differ from those used during preference collection. The three metrics (BIAI ratio, lap time, and number of incomplete laps) reported in Table \ref{tab:GT_control_lap_time_maggiore} show that the reward model trained on Sarthe track preferences can still produce effective policies on the Maggiore track. In particular, reward models trained with Soft-TAC reduced the average lap time by 1.4 seconds compared to Cross-Entropy and outperformed the default reward by 0.4 seconds. Reported metrics exclude incomplete laps that exceed the 1800-second threshold; this never occurred with Soft-TAC, but did with the other baselines.

\begin{table}[h]
\centering
\caption{BIAI ratio ($\pm$ SE), minimal lap time ($\pm$ SE), and number of incomplete laps after RL training with each reward (lower is better) on the Maggiore track for the time-trial task in GT7.}
\label{tab:GT_control_lap_time_maggiore}
\begin{tabular}{l ccc}
\toprule
\textbf{Method} & \textbf{BIAI} & \textbf{Lap} & \textbf{Lap Time} \\
 & \textbf{Ratio} & \textbf{Time} & \textbf{$>$ 1800} \\
\midrule
\textsc{Soft-} & 0.980 $\pm$ 0.00 &  135.85 $\pm$ 0.20 & 0 \\
\textsc{TAC} & & & \\
\textsc{Cross-} & 0.990 $\pm$ 0.00 & 137.39 $\pm$ 0.31 & 15 \\
\textsc{Entropy} & & & \\
\textsc{Default} & 0.983 $\pm$ 0.00 & 136.33 $\pm$ 0.11 & 1 \\
\bottomrule
\end{tabular}
\end{table}

\section{Empirical Robustness of TAC}
\begin{figure*}[h!]
    \centering
    % The \includegraphics command inserts your image file.
    \includegraphics[width=0.95\textwidth]{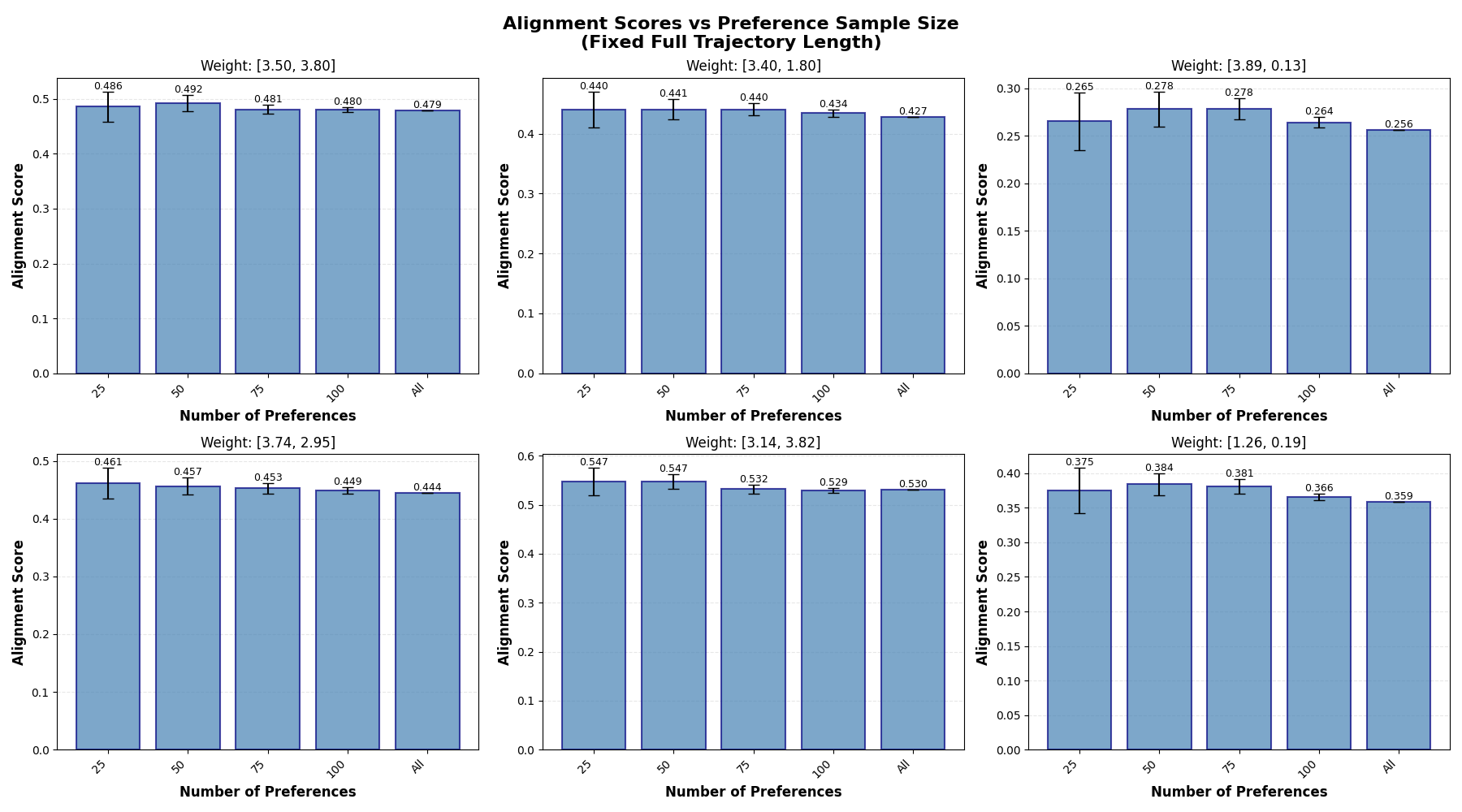}
    \caption{Average TAC ($\pm$SE) versus Number of Human Preferences in GT7}

    \label{fig:TAC_versus_pref}
\end{figure*}

We performed two ablations to further evaluate the robustness of TAC. First, we examined how TAC is affected by the number of human preferences used in its computation. Using the human preference dataset from the aggressive–timid reward learning experiment, we randomly sampled 30 sets of preferences with sample sizes of 25, 50, 75, and 100 from the full set of 118. Separately, we randomly sampled reward weights for the overtaking and collision features to generate multiple reward functions (each shown in an individual plot in Figure~\ref{fig:TAC_versus_pref}). TAC scores were then computed for each reward function using the different preference sample sizes. We found that the average TAC score remained largely consistent across all tested sample sizes, while the standard error decreased as the number of preferences increased. This result indicates that TAC is robust even when the preference dataset was small, making it suitable for domains where collecting preference data is challenging. 

\begin{figure*}[h!]
    \centering
    % The \includegraphics command inserts your image file.
    \includegraphics[width=0.95\textwidth]{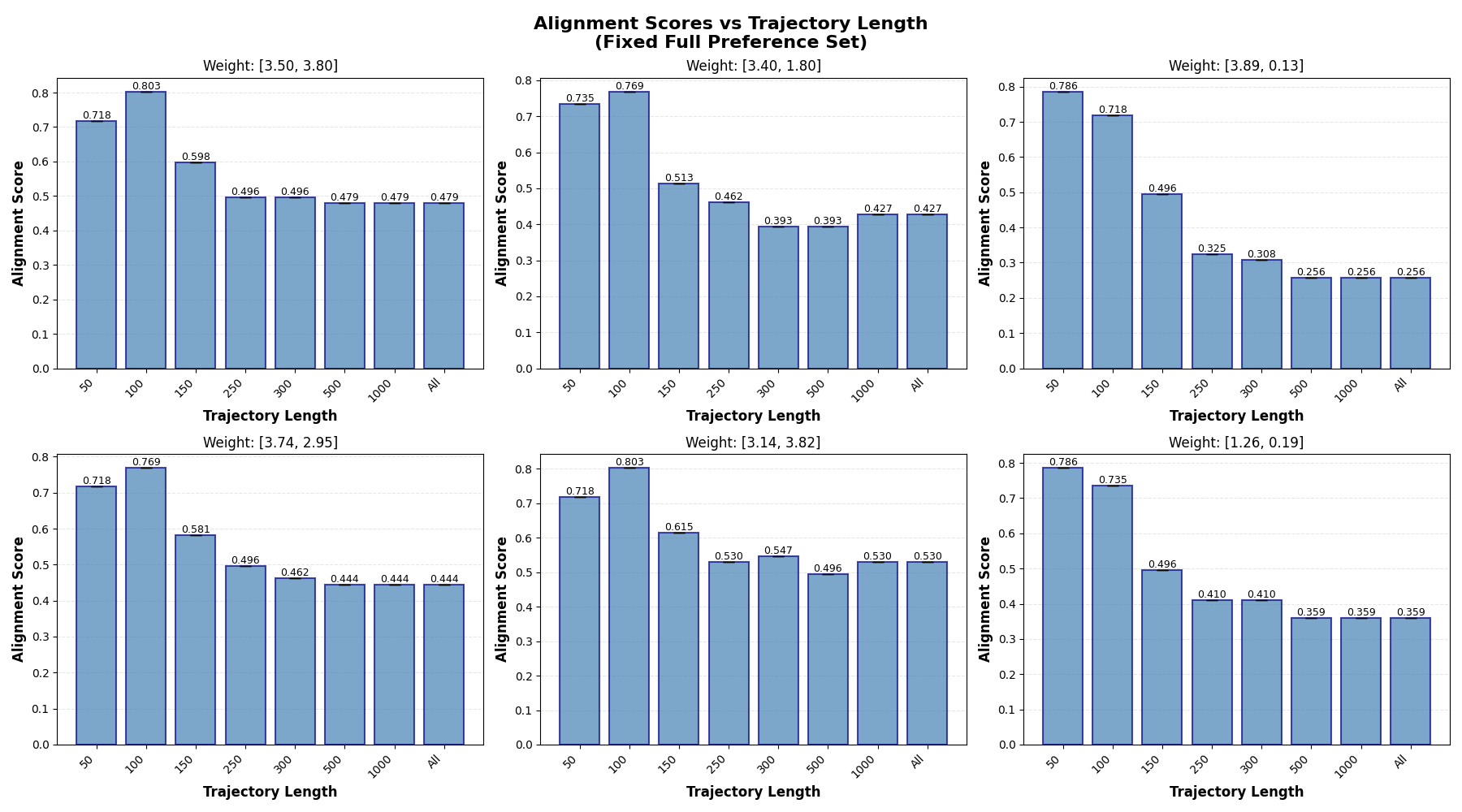}
    \caption{TAC versus Trajectory Length in GT7}

    \label{fig:TAC_versus_traj_length}
\end{figure*}

For the second ablation, we examined the effect of trajectory length on TAC by reusing the same aggressive–timid preferences collected over full trajectories, while varying only the length of the trajectory segments used to compute TAC. In our main experiments, preferences were elicited over entire trajectories, whereas prior PbRL work often considers preferences over shorter trajectory segments. Accordingly, in this ablation, we fix the preference dataset and vary only the trajectory length. GT7 trajectories are relatively long, ranging from 2,040 to 12,407 time steps; here, we computed TAC scores using trajectory segments of length 50, 100, 150, 250, 300, 500, and 1,000 time steps, extracted from the original trajectories. As shown in Figure~\ref{fig:TAC_versus_traj_length}, TAC was also robust to the trajectory length: using segment lengths of at least 500 time steps produced results comparable to full trajectories, which are much shorter segments compared to full GT7 trajectories. Segments shorter than 500 yielded lower similarity, which is expected because the human preferences were expressed over entire trajectories, and shorter segments may not capture sufficient behavioral information to reflect those preferences.

\end{document}